\documentclass[HARVARD,Times1COL]{WileyNJDv5} %STIX1COL,STIX2COL,STIXSMALL

\articletype{Article Type}%

\usepackage{tabularx}
\usepackage[table]{xcolor}
\usepackage{array}
\usepackage{enumitem}
\usepackage{tabularx}
\usepackage[table]{xcolor}
\usepackage{array}
\usepackage{enumitem}
\usepackage{threeparttable}
\newcolumntype{Y}{>{\raggedright\arraybackslash}X}

\definecolor{expblue}{HTML}{4F8DD3}
\definecolor{expgreen}{HTML}{6DBB63}
\definecolor{exppurple}{HTML}{8B6BC8}

\definecolor{lightblue}{HTML}{F3F8FF}
\definecolor{lightgreen}{HTML}{F6FBF3}
\definecolor{lightpurple}{HTML}{F8F5FC}

\newcommand{\cellitemsfirst}[1]{%
\vspace{-0.1em}
\begin{itemize}[
  leftmargin=1.1em,
  itemsep=1pt,
  topsep=5pt,
  parsep=0pt,
  partopsep=0pt,
  label={\tiny$\diamond$}
]
#1
\end{itemize}
}

\newcommand{\cellitems}[1]{%
\begin{itemize}[
  leftmargin=1.1em,
  itemsep=1pt,
  topsep=5pt,
  parsep=0pt,
  partopsep=0pt,
  label={\tiny$\diamond$}
]
#1
\end{itemize}
}

\usepackage{graphicx} 
\usepackage{subcaption}
\usepackage[font=normalsize]{caption} 
\received{Date Month Year}
\revised{Date Month Year}
\accepted{Date Month Year}
\journal{Journal}
\volume{00}
\copyyear{2023}
\startpage{1}

\begin{document}

\title{Factors Influencing the Emergence of Dependency Length Minimization in Neural Agent Simulations}
% EVOLANG: \title{Simulating Dependency Length Minimization \\ using neural-network based learning and communication}
% CogSci: Neural-agent Language Learning and Communication: Emergence of Dependency Length Minimization
% COLING: Endowing Neural Language Learners with Human-like Biases: \\ A Case Study on Dependency Length Minimization
% Factors influencing the Emergence of Dependency Length Minimization in neural agent simulations

\author[1]{Yuqing Zhang}

\author[2]{Tessa Verhoef*}

\author[1]{Gertjan van Noord}

\author[1]{Arianna Bisazza*}

\authormark{Zhang \textsc{et al.}}

\titlemark{Factors Influencing the Emergence of Dependency Length Minimization in Neural Agent Simulations}

\address[1]{
\orgdiv{Center for Language and Cognition},
\orgname{University of Groningen},
\orgaddress{
\city{Groningen},
\country{The Netherlands}
}
}

\address[2]{
\orgdiv{Leiden Institute of Advanced Computer Science (LIACS)},
\orgname{Leiden University},
\orgaddress{
\city{Leiden},
\country{The Netherlands}
}
}

\address[]{*Shared senior authorship.}

\corres{
Corresponding author: Yuqing Zhang
\email{yuqing.zhang@rug.nl}
}

\abstract[Abstract]{Given various grammatical options, language users prefer the word order choice that reduces the overall length of syntactic dependencies, a principle known as dependency length minimization (DLM). The origins of this preference remain an open question, particularly whether it originates from constraints on efficient information processing. Computational simulations provide a powerful approach to identifying the factors influencing the emergence of linguistic phenomena. 
However, previous simulations of DLM have not examined realistic interaction contexts and have produced mixed results.
The present study investigates the emergence of DLM in artificial languages using a recently proposed language learning and communication framework based on recurrent neural networks (RNNs). In this framework, agents are trained to speak and interpret artificial languages and then use these languages to communicate. 
Using this framework, we study the impact of several factors related to processing limitations in a communicative setting, such as noise during listening, limited speaker capacity, and incremental sentence processing.
Our results reveal a complex interplay among these factors in shaping word order preferences in neural agents. 
Specifically, in the full meaning space, agents regularize toward a single dominant word order, while in the half meaning space they show a short-before-long preference that only aligns with DLM in verb-initial languages. A consistent DLM preference emerges only when agents are subject to incremental processing pressure.
These results suggest that limitations in human cognitive processing may indeed play a role in shaping DLM.
Our findings provide insights into the conditions under which neural models replicate human-like preferences for minimizing syntactic dependency distances and highlight the challenges of designing emergent communication models that capture human cognitive biases in language processing.}

\keywords{language emergence, language universals, dependency length minimization, neural-network-based simulations, incremental processing}

% \jnlcitation{\cname{%
% \author{},
% \author{},
% \author{}, and
% \author{}}.
% \ctitle{} \cjournal{\it J Cog Sci.} 
% \cvol{2025;00(00):0--0}.}

\jnlcitation{\cname{%
\author{Yuqing Zhang},
\author{Tessa Verhoef},
\author{Gertjan van Noord}, and
\author{Arianna Bisazza}}.
\ctitle{Factors Influencing the Emergence of Dependency Length Minimization in Neural Agent Simulations}
\cjournal{\it J Cog Sci.}
\cvol{2026;00(00):0--0}.}

\maketitle

\renewcommand\thefootnote{}

\renewcommand\thefootnote{\fnsymbol{footnote}}
\setcounter{footnote}{1}

\section{Introduction}\label{sec1}

When multiple word order choices are available to express a message, human language speakers prefer the order that reduces the overall dependency length between syntactic heads and their dependents \citep{arnold2000heaviness, futrell2020dependency}.
The origins of such a tendency towards dependency length minimization (DLM) remain an open question \citep{culbertson2014language,fedzechkina2018human}: does it emerge from tracking statistical patterns in linguistic input, reflect domain-general cognitive biases in information processing, or stem from pressures to communicate efficiently?
Crosslinguistic experiments have established that sentences with longer dependencies increase various processing demands, such as parsing time, cognitive load, and memory retrieval challenges \citep{gibson1998linguistic, grodner2005consequences, lewis2005activation}. % sentence processing, linguistic/sentential complexity
A fruitful approach to studying the influence of human cognitive biases and socio-cultural processes in shaping linguistic structure is to simulate them computationally \citep{steels1997synthetic, Hurford_2000, de2006computer}. 
Recent advances in machine learning and computational linguistics have yielded powerful % neural-network-based
statistical learners that can deal surprisingly well with the complexity of human languages and can be used to set up increasingly realistic simulations. 
Artificial neural networks rely on statistical learning for acquiring representations and assume minimal inductive biases specialized for language. 
By comparing the behavior of models with that of human learners, we can gain insights into which types of linguistic knowledge 
can be learned from the data by a weakly biased, domain-general learner, %can be learned just on the basis of the data, 
and what aspects arise from additional cognitive constraints or communication pressures from an empirical standpoint. These comparisons may also reveal that neural network models lack important human-like biases, which in turn can explain the limited abilities of current language models to generalize in linguistically sound ways \citep{hahn2020universals, chaabouni2021communicating,portelance2021emergence,ren2020compositional,warstadt2022artificial, portelance2023roles}.

In this line of work, researchers have investigated the biases emerging from sequence-processing neural networks through controlled artificial language communication setups. In particular, \citet{chaabouni2019word} studied whether sequence-to-sequence agents based on the Long Short-Term Memory (LSTM) architecture trained \textit{in iterated learning settings} exhibit preferences for natural word-order constraints, including a tendency to minimize long-distance dependencies. Although LSTM-based agents lacked explicit hierarchical syntactic representations, results showed that LSTM-based agents displayed a strong bias against long-distance dependencies, and the bias became stronger after generational transmissions, suggesting that locality pressures can arise from general sequence-processing and learning dynamics even without hard-coded syntactic structure. However, an important challenge in this line of work is that such artificial learners still often behave differently from human learners \citep{chaabouni2019anti,galke2022emergent, munoz2024contrasting}. Specifically in the context of DLM, \citet{munoz2024contrasting} showed that humans show a higher degree of optimization according to DLM than large language models.
Some simulation studies have reported results that contradict the findings of \citet{chaabouni2019word}. In production tasks involving miniature artificial languages, % LSTMs were found to prefer short dependencies while 
pre-trained Transformers such as Bidirectional and Auto-Regressive Transformer (BART) and T5 did not show a preference for short dependencies \citep{zhao2022probe}. These studies, however, are difficult to compare directly due to differences in architectures, \mbox{(pre-)training} regimes, and artificial language design.
Moreover, most studies adopted linearized meaning representations, which could be implicitly biased towards specific word orders, potentially influencing the observed preferences for short dependencies.

% The current work is situated within this broader line of research that
% Our experimental design extends this research direction by isolating processing and communication constraints while keeping sequence length fixed. 

In this study, we address these issues by adopting the recently proposed neural-agent language learning and communication (NeLLCom) framework \citep{lian2023communication} to simulate the emergence of DLM preferences in artificial languages used by paired speaking and listening neural agents. NeLLCom agents first learn to speak and interpret predefined artificial languages, and then use them to communicate, providing a platform for examining the evolution of specific linguistic properties.
Secondly, we follow an artificial language learning paradigm and avoid any pre-training of the networks to rule out the possibility that the observed preferences are inherited from statistical properties of real-language training corpora.
Thirdly, we use miniature languages that are directly inspired by an artificial language learning experiment with human subjects \citep{fedzechkina2018human}. We model dependency relations through non-uniform word distributions and the selectional preferences of verbs \citep{katz-fodor-63, mcrae1998modeling}, i.e., the strength of association between an action and its likely agent or patient. 
Finally, we represent meaning in a way that is not biased toward any specific linear order, following recent work on the simulation of language universals \citep{lian2023communication, lian-etal-2024-nellcom}. 

%%% explain the factors and our expectations
Building upon recent studies, we investigate the impact of several factors on developing a preference for shorter dependencies. These factors are related to processing limitations and environmental pressures in a communicative setting, namely: noise during listening \citep{gibson2013rational, brochhagen2017effects, futrell-levy-2017-noisy}, % through a word dropout technique \citep{gal2016theoretically} whereby randomly chosen input tokens are masked to the listener 
limited speaker capacity \citep{lewis2005activation, vasishth2019computational}, and incremental sentence processing \citep{kamide2003time, rita-etal-2020-lazimpa}. % or the extent to which an utterance's meaning can be guessed before hearing it entirely.
% in humans there is no specific 'intrinsic' bias for DLM, but that the preference emerges from domain-general processing constraints.
We therefore frame the study around the following research questions:
(i) When agents are trained on a neutral mixed language, do they drift toward shorter dependencies during communication? 
(ii) Do noise during listening and reduced speaker capacity affect DLM preferences? Because these manipulations operationalize broader communicative and architectural pressures, their effects on dependency length must be tested empirically.
(iii) Does explicitly rewarding incremental interpretation in the listener elicit a DLM preference? Since such models incorporate more human-like memory constraints through incremental prediction, they are more cognitively plausible \citep{rita-etal-2020-lazimpa}.
This formulation allows us to distinguish between processing-based predictions of DLM and broader operational manipulations that may affect word order through indirect pressures on communication and production.
It also allows us to test whether DLM emerges as a default production preference in speaker agents, or whether it requires listener-side incremental comprehension pressure to become communicatively advantageous.
% We formulate the following hypotheses:
% (i) When noise is present, speakers trained on utterances mixed with short and long dependency length (DL) will develop an increased preference to produce more short DL alternatives during communication, as overall shorter DL increases robustness to transmission errors under noisy-channel communication \citep{gibson2013rational, brochhagen2017effects, futrell-levy-2017-noisy}.}
% (ii) When processing capacity is limited, speakers increasingly favor utterances with shorter DLs because reduced DL lowers memory and retrieval costs during production and comprehension \citep{lewis2005activation, vasishth2019computational}.}
% (iii) When communication success is evaluated by listeners with an incremental processing mechanism, speakers will show a stronger preference for shorter DLs, since such models incorporate more human-like memory constraints through incremental prediction and are therefore more cognitively plausible \citep{rita-etal-2020-lazimpa}.}
% by the extent to which an utterance's meaning can be guessed before hearing it entirely. 
% a higher learning speed and final learning performance for the language that has overall shorter dependencies, in both listening and speaking agents, 
To test potential architecture-dependent effects, we include a partial comparison between agents based on vanilla Recurrent Neural Networks \citep{elman1990finding} and Gated Recurrent Units (GRUs) \citep{cho-etal-2014-learning}. The detailed architectural setup is specified in \S\ref{sec:adaptation_nellcom}.
% But the gating mechanism makes it potentially ``too good" at 
% remembering the history \citep{can2020gating}. While vanilla RNN can be considered as a baseline for RNNs and is more prone to forgetting.

Our results reveal a complex interplay between these factors in shaping word order preferences in neural agents. 
% The result we found made us realize something important about those languages, which had not emerged from the human experiments (i.e. a possible competing pressure to regularize word order), and we rightfully commented on that , and later fixed it through subsequent experiments. 
The observed ordering patterns were not fully anticipated by our original DLM-based predictions, as agents showed competing ordering pressures beyond DLM.
Across several experimental conditions, agents developed a robust preference to produce utterances in which shorter constituents precede longer ones. 
This short-before-long preference \citep{behaghel1909beziehungen} offers a more comprehensive explanation of the observed ordering behavior than DLM alone.
% This bias alone accounts for much of the observed ordering behavior in both verb-initial and verb-final artificial languages.
However, when communication success was evaluated by listeners with an incremental processing mechanism, agents developed a preference consistent with DLM that cannot be explained by short-before-long ordering alone. In this setting, incremental comprehension during communication increased the preference for the word order that minimized the linear distance between heads and dependents over its alternative.
% Under our experimental conditions, neural agents show a preference for reducing DL in production, particularly when processing limitations are strengthened in learning verb-initial languages.
% Specifically, for verb-initial artificial languages, noise, limited speaker capacity, and incremental processing contribute to an increased preference for the word order that reduces DLs over its alternative. % in the production of the verb-initial language
% By contrast, for verb-final languages, only incremental processing promotes DLM. 
Although some effects were architecturally dependent, these findings suggest that limitations in human cognitive processing can shape DLM preferences. 
% The weaker preference towards DLM in the verb-final language is consistent with patterns found in natural languages \citep{futrell2015large, jing2022dependency, ferrer-i-cancho_optimality_2022}.

This paper builds upon prior work \citep{yuqing-2024-dlm} and extends it significantly by modifying the experimental design to eliminate competing pressures from word order regularization, which turns out to have a key effect on the observed results.
Specifically, we conduct three linked experiments: Experiment 1 (\S\ref{sec:exp1}) adopts a "full-meaning space" setup that closely mirrors the original human experimental design by \citet{fedzechkina2018human}, showing that neural agents tend to reduce overall production entropy through word order regularization, which overrides DLM. To disentangle DLM effects from word order regularization, experiments 2 and 3 use a simpler half-meaning space. Experiment 2 (\S\ref{sec:exp2}) demonstrates that short-before-long preferences override DLM, a tendency that is further facilitated by noise and limited speaker capacity. Finally, to address the role of comprehension, Experiment 3 (\S\ref{sec:exp3}) explicitly models incremental prediction in the listener, successfully showing that incremental processing elicits DLM.

Overall, our findings offer insights into the essential elements that contribute to the development of DLM preferences in purely statistical, neural-network-based learners. More generally, we demonstrate the importance of making simulations more realistic and cognitively plausible by considering factors, such as noise and incremental processing, that are typically overlooked in recent neural-agent simulations of language emergence.

\section{Background}
\subsection{Dependency Length Minimization}

% % When grammatical alternatives exist for the construction, there is a preference for constituents of shorter length to occur closer to their syntactic heads, yielding a shorter overall dependency length in the sentence.
Although the world's languages exhibit a wide range of variation along many linguistic dimensions \citep{comrie1989language, croft2002typology, evans2009myth}, work in linguistic typology shows that certain linguistic patterns (co-)occur more frequently than expected by chance, known as statistical language universals \citep{greenberg1963some, dryer1992greenbergian}.
A popular hypothesis posits that observed linguistic properties arise from the trade-off between the need to reliably exchange information and the cost of language production and processing effort \citep{zipf1949human, jaeger2011language, kemp2012kinship, kirby2015compression, gibson2019efficiency}. This hypothesized principle has been proposed to operate on one of the most fundamental grammatical properties -- word order \citep{futrell2020dependency}.
DLM, operationalized as minimizing the overall linear distance between words linked in the grammatical head-dependent relationships, is one such well-known word order universal \citep{ferrer2004euclidean}.

The generally accepted motivation for DLM is that minimizing dependency length enables efficient language comprehension and production. 
From the perspective of communication, dependency length is a component of the effort involved in sending and receiving messages, beyond message length. 
% general cognitive constraints or imperfect memory of individual language speakers \citep{futrell-levy-2017-noisy}. 
From the perspective of language processing, shorter dependency length reduces the difficulty associated with information retrieval. % incremental production and comprehension.
For instance, \citet{hawkins1994performance} proposes that reducing the linear order between related constituents minimizes the search time required for an incremental language parser to determine the correct head of a phrase.  \citet{gibson1998linguistic} proposes that long dependencies exert extra pressure on working memory in both language parsing and generation by requiring language users to keep word representations in the working memory for a longer time. Whether due to a search-time or memory constraint, DLM is regarded as inherent to the underlying language processing mechanism, which is commonly assumed to be universal rather than dependent on a given speaker's language.
% corpus studies: Ramon Ferrer-i-Cancho. 2017. The placement of the head that maximizes predictability: An information theoretic approach. -- Ramon Ferrer-i-Cancho did one of those early corpus studies on DLM)

The universality of DLM has been established by both corpus-based analyses \citep{ferrer2004euclidean, liu2008dependency, futrell2015large, ferrer39placement} and psycholinguistic studies \citep{gibson1998linguistic, grodner2005consequences}.
Specifically, statistical corpus tests have shown that the observed word orders of human languages have a shorter dependency length than random baselines \citep{futrell2015large}, suggesting that DLM is a cross-linguistic phenomenon and providing correlative evidence of the existence of the DLM bias. Furthermore, psycholinguistic experiments using an artificial language learning paradigm among different groups of human participants \citep{fedzechkina2018human, fedzechkinadependency, zhao2022probe} have found that subjects exposed to novel miniature languages systematically restructured the input toward shorter dependency lengths, regardless of the order preferences found in their native language. These findings connect the structures found in the corpus statistics to the general information processing biases in individual language speakers. 
The artificial languages used in our work (Figure~\ref{fig:lang}) are directly inspired by those used in the human setup of \citet{fedzechkina2018human}.

The abstract DLM principle takes on different surface orders in different languages but serves the same goal of reducing the dependency length. In other words, speakers of different language types have different word order preferences. For example, users of verb-initial languages (i.e., languages that place the verb before its dependents) tend to place short postverbal constituents before long ones, i.e., a short-before-long preference in postverbal positions \citep{arnold2000heaviness, wasow2002postverbal}. In contrast, users of verb-final languages (i.e., languages that place the verb after its dependents) typically prefer long preverbal constituents before short ones, i.e., a long-before-short ordering in preverbal positions \citep{yamashita2001long}.
%  the strength of the preference is modulated by L1 statistics. 
%
While extensive corpus studies have confirmed a universal preference for DLM, they also show considerable variation in the level of DLM optimization across languages \citep{ferrer-i-cancho_optimality_2022}. For example, \citet{gildea2010grammars} found that the observed dependency length of English is closer to its theoretical optimal baseline for DLM than verb-final German. \citet{liu2008dependency} found that among the 20 languages examined, Mandarin Chinese has the longest average dependency length. When closely examining double prepositional phrase orderings in English and Mandarin, \citet{liu2019comparative} found that Mandarin Chinese shows a weaker preference for shorter dependencies.
\citet{futrell2015large} observed that head-final languages such as Japanese and Turkish have longer dependencies than head-initial languages like Italian and Indonesian. 
In the analysis of transitive and ditransitive constructions in Basque, Polish, and Spanish, \citet{ros2018minimizing} found that Basque, an Object-Verb (OV) order language characterized by both verb-medial and verb-final orders, shows weaker DLM effects than the other two verb-initial languages.
In addition, the degree of word order shifting for DLM was found to be lower in Basque \citep{ros2018minimizing} than in Japanese \citep{yamashita2001long} and Korean \citep{dennison2007universal}. % --- verb-final

% DLM for preverbal constituent ordering preferences in SOV languages, including Basque, Hindi, Japanese, Korean, Latin, Persian, and Turkish
% The basic word order is subject–verb–object (SVO), as in English. Otherwise, Chinese is chiefly a head-final language, meaning that modifiers precede the words that they modify. 
% Online experiments from Yamashita and Chang (2001) showed that speakers tend to place long arguments ahead of short ones before the head verb in both transitive and ditransitive constructions in Japanese.

The high cross-linguistic variation in dependency length suggests that word order is not always optimized solely for DLM, because languages must balance dependency length with other word order principles.
Consistent with this view, \citet{hahn2022crosslinguistic} show that word order patterns reflect a trade-off between dependency locality and information locality.
Related information-theoretic accounts make similar predictions from different perspectives.
% One important class of such pressures comes from information-theoretic principles of word order. 
For example, \textit{surprisal minimization} or \textit{predictability maximization} \citep{ferrer39placement, ferrer-i-cancho_predictability_2024, ferrer-i-cancho_optimal_2025} may favor different placements of heads and dependents in a sentence, as suggested by studies on noun phrases \citep{ferrer-i-cancho_optimal_2025} and star structure sequences \citep{ferrer-i-cancho_anti_2021}.
Another account is \textit{Uniform Information Density} (UID), which predicts that speakers tend to distribute information relatively evenly across an utterance. UID has been used to explain local production choices such as syntactic reduction \citep{jaeger2006speakers}, and recent work has extended UID-based explanations to cross-linguistic word order patterns \citep{clark2023cross}.
Word order is also shaped by domain-general cognitive biases that can counteract DLM. For instance, an \textit{animacy-first} bias often drives speakers to place animate entities earlier, regardless of their syntactic role \citep{chang2009learning}. Similarly, \textit{lexical grouping} strategies may lead learners to cluster semantically related items linearly rather than computing abstract hierarchical dependencies.
% projectivity (no crossing dependency edges and/or projection lines in dependency parse trees) 
% and head consistency, which do not always lead to a minimized DL \citep{futrell2020dependency}. 
In addition, DLM has also been found to correlate with other grammatical features. Specifically, languages that are head-final or have rich morphological cases display weaker DLM tendencies \citep{futrell2020dependency, futrell2015large, gildea2010grammars, liu2020mixed}. Among possible explanations, rich morphological systems are argued to add redundancy and provide cognitive ease that alleviates the need for DLM. The DLM principle is weaker in verb-final subject-object-verb (SOV) languages such as Korean, German and Turkish \citep{ferrer-i-cancho_optimality_2022, liu2020mixed, futrell2020dependency, dyer2023revisiting}. % Japanese and Turkish 
Based on these observations, \citet{jing2022dependency} have proposed a slightly more refined formulation of dependency length optimization, by which DLM is inversely correlated with the overall presence of head-final dependencies, and the accumulation of parallel dependents and disharmonic hierarchical structures in verb-final languages can lead to anti-DLM patterns.
% In addition, in the large-scale corpus study, \citet{futrell2015large} also reported that real verb-final languages tend to show less DLM compared to verb-initial languages. The differences align with \citet{jing2022dependency}'s refined formulation of dependency length optimization,
% which shows that DLM effects are weaker or even reversed in head-final languages. Their study demonstrates that the accumulation of parallel dependents and disharmonic hierarchical structures in verb-final languages can lead to anti-DLM patterns.} 
% ---- moved from section 7.1

% case marking adds redundancy to the speech signal and increases the probability of successful communication (MacWhinney et al., 1984; Sasaki and MacWhinney, 2006; Tily, 2010)

\subsection{Neural Network-based Simulations}
Neural-network-based communication models provide a useful platform for testing existing psycholinguistic theories on the origin of linguistic universals, because they allow researchers to manipulate the communicative and architectural constraints under which linguistic systems emerge \citep{chaabouni2019anti, chaabouni2019word, yedetore-etal-2023-poor, mueller2023plant}. Previous emergent communication studies have shown that such models can give rise to linguistic patterns such as compositional structure \citep{chaabouni2020compositionality}, the shape bias \citep{portelance2021emergence}, efficient color-naming systems \citep{chaabouni2021communicating, carlsson2024cultural, zhang-etal-2025-nellcom}, and the word-order/case-marking trade-off \citep{lian2023communication,lian-etal-2024-nellcom, lian2025simulating}. However, such patterns do not necessarily emerge from generic communication pressure alone; specific constraints such as least-effort and impatience biases \citep{rita-etal-2020-lazimpa} or interfering noise \citep{ueda-washio-2021-relationship} may be needed for neural agents to develop an efficient code consistent with Zipf's law of abbreviation. This motivates our use of neural-agent communication models to test whether constraints motivated by noisy-channel communication, limited processing capacity, and incremental comprehension are sufficient to produce DLM-compatible word order.

Some of the previous simulation studies were specifically focused on DLM:  
\citet{chaabouni2019word} trained sequence-to-sequence LSTM agents to communicate about paths (e.g., UP DOWN LEFT) in a simple grid world by using miniature artificial languages as utterances (e.g., M1 UP 1 M2 DOWN 1 M3 LEFT 1). %(M1 is the first marker to indicate first step of the trajectory)
Their experiments showed higher learning speed for short-dependency languages, which resulted in increased production of short-dependency utterances across generations of a simulated iterated learning procedure. Since their meaning representation was itself a sequence, it cannot be ruled out that the RNN sequence-to-sequence agents were actually biased towards languages having less reordering with respect to the input meaning sequence.
In a different study, \citet{futrell2019rnns} found that RNNs trained on English texts prefer ordering short constituents before long ones regarding several DLM-related linguistic phenomena (heavy NP shift, particle shift, dative alternation, and genitive alternation). 
\citet{zhao2022probe} conducted similar tests using pre-trained transformers (i.e. BART \citep{lewis-etal-2020-bart} and Generative Pre-trained Transformer (GPT-2) \citep{radford2019language})
and found these models acquired human-like word order preferences that are consistent with the DLM principle. 
% zhao used original English sentences as meaning and artificial language sequence as utterance. 
% transformers can connect every word with others with the attention mechanism (i.e., no explicit notion of sequence order). 
However, the same models did not show a similar preference in production tasks involving an artificial semi-English language, which \citet{zhao2022probe} attributed largely to the lack of explicit memory constraints in transformer architectures due to self-attention.
In summary, previous neural simulations of DLM leave open several methodological questions: linearized meaning representations create simple input-reordering biases \citep{chaabouni2019word}, pre-trained Transformers may reflect English-specific training biases and lack human-like memory limits, and most paradigms ignore dynamic communicative pressures. Our framework addresses these gaps by isolating the factors influencing DLM using unordered meaning representations, agents trained from scratch, and explicit communicative constraints such as noise and incremental processing.

\section{Method}
Our experimental setup is motivated by several factors.
Firstly, we used miniature languages that were directly inspired by an artificial language learning experiment with human subjects \citep{fedzechkina2018human}.
We tested whether and under which conditions neural learners exposed to an artificial language containing mixed patterns would prefer typologically common patterns rather than rare ones.
Secondly, we combined the artificial language learning setup with a communication game between agents, following the neural-agent language learning and communication (NeLLCom) framework recently proposed to simulate the evolution of language universals \citep{lian2023communication}. Using this framework, we represented meaning in a way that was not biased towards any specific linear order. This choice also aligned with the common architectural setting in emergent communication studies \citep{chaabouni2019anti, chaabouni2020compositionality}. 
Thirdly, we examined the impact of several factors related to processing limitations, including noise during listening \citep{gibson2013rational, brochhagen2017effects, futrell-levy-2017-noisy}, limited speaker capacity, and incremental sentence processing \citep{kamide2003time}.

Guided by the general motivation, the rest of this section introduces the chosen framework, the miniature artificial languages, and the proposed factors that enable more realistic simulations and mimic processing limitations. 

\subsection{Adaptation of the NeLLCom Framework}
\label{sec:adaptation_nellcom}

While artificial neural learners often behave differently from human learners, recent advancements have been made to bridge this gap \citep{galke2022emergent, lian2023communication, lian-etal-2024-nellcom, lian2025simulating}. \citet{lian2023communication} introduced the NeLLCom framework for simulating language learning and change with artificial languages and neural-network learners, which addresses some of the challenges posed by the differences between artificial and human learning and provides promising solutions to make the simulations more realistic.

\begin{figure*}[htbp]
    \centering
    \begin{subfigure}{0.49\textwidth} 
        \centering
        \includegraphics[width=\linewidth]{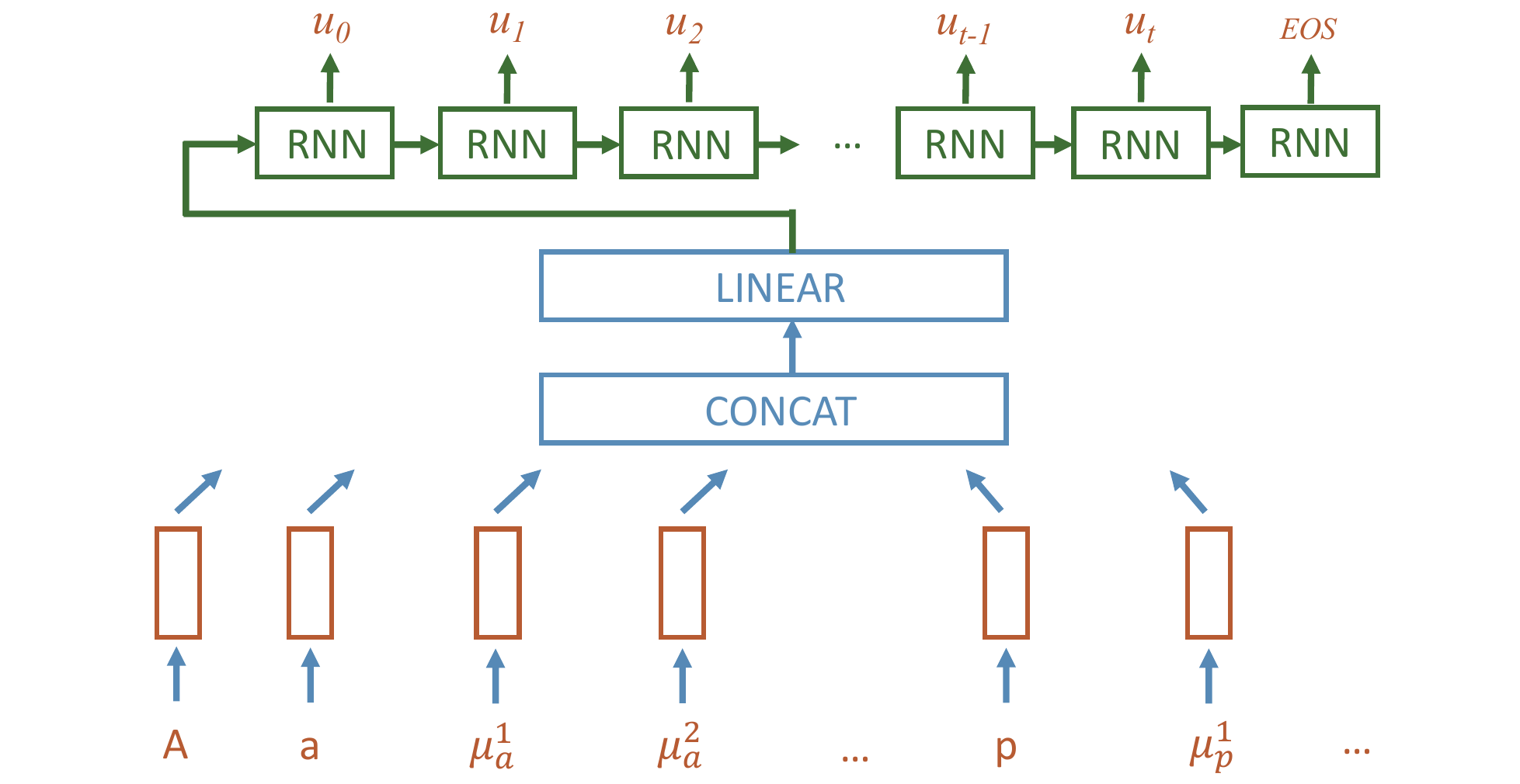}
        \caption{RNN-based speaker architecture}
        \label{fig:spk}
    \end{subfigure}
    \hfill 
    \begin{subfigure}{0.49\textwidth}
        \centering
        \includegraphics[width=\linewidth]{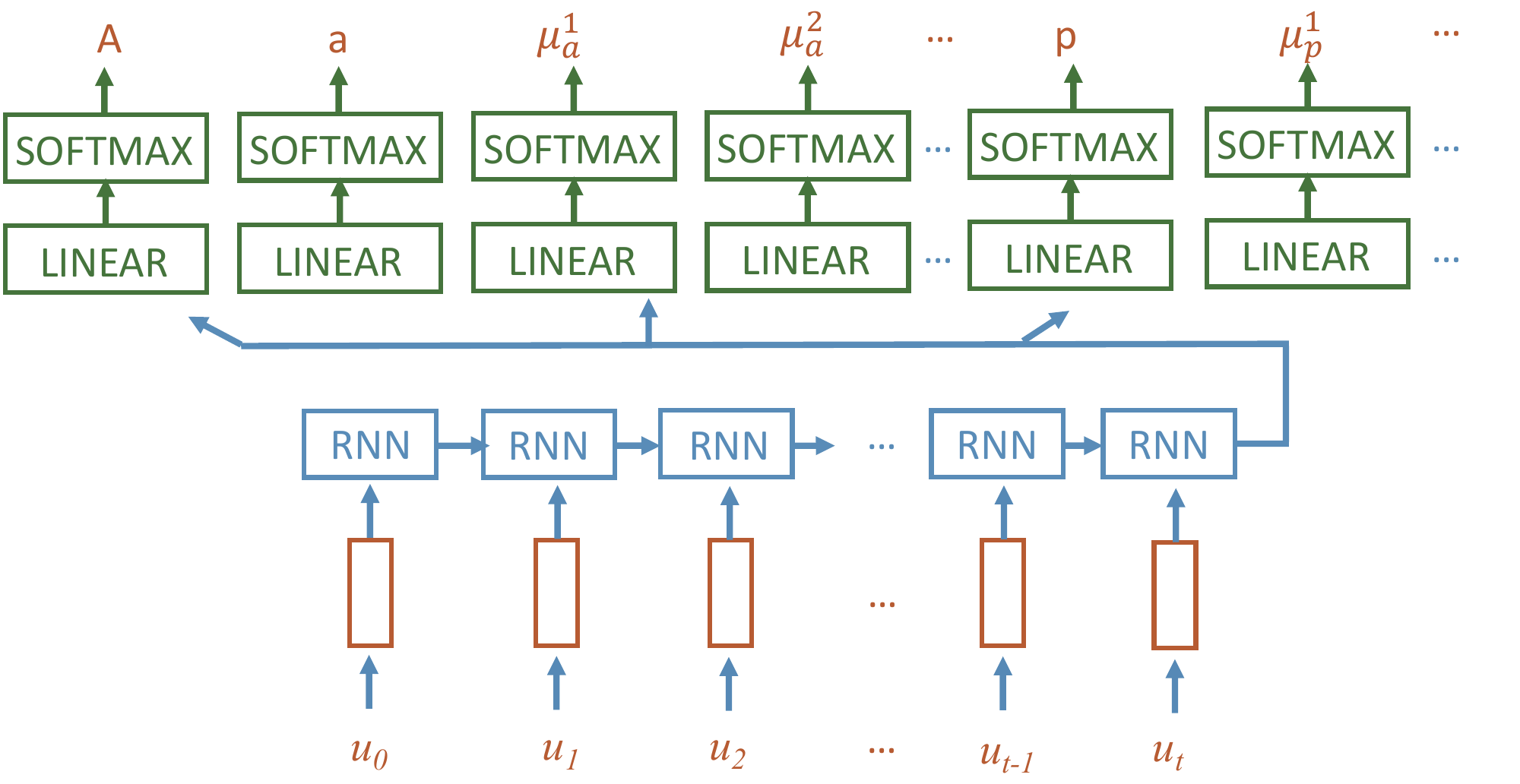}
        \caption{RNN-based listener architecture}
        \label{fig:lst}
    \end{subfigure}
    \caption{Neural agent architectures.}
\end{figure*}

In the original framework, neural network-based agents are trained to exchange messages in a simplified world of agent-patient-action triplets (e.g., \textit{Tom-Jerry-chase}) using predefined miniature languages. 
Listening is defined as the process of converting a sequence of symbols (utterance: $\mathbf{u}$) into an unordered set of items (meaning: $\mathbf{m}$), whereas speaking is the reverse process of converting a meaning into an utterance.
After a listening/speaking training phase based on supervised learning (SL), pairs of agents communicate with each other using the learned language while maximizing communication success via reinforcement learning (RL).
Specifically, both agents' goal is to maximize the listener's ability to reconstruct the intended meaning $\mathbf{m}$, given a speaker-generated utterance $\hat{\mathbf{u}}$.
Testing is performed on meanings not observed during any training phase.
To enable experimenting with different dependency lengths, we expanded \citet{lian2023communication}'s original meaning space $m$ composed of action-agent-patient triplets ($\{A,a,p\}$, $n=3$) by adding optional modifier phrases to agent and patient: $\mu_a$ (agent modifier phrase) and $\mu_p$ (patient modifier phrase), respectively.
Each modifier phrase consists of three items corresponding, respectively, to adposition, adjective, and inanimate noun as in \citet{fedzechkina2018human}'s experiment (e.g., \textit{by frozen river}).

In terms of \emph{architecture},
the listening agent has a \textit{sequence-to-linear} structure (see Figure~\ref{fig:lst}). The input utterance is passed to the listening agent's encoder
% \footnote{Following \citet{lian2023communication}, we use Gated Recurrent Units, or GRU \citep{cho-etal-2014-learning} in all experiments. In the preliminary phase of our experiments, we observed similar results obtained from LSTM and GRU models.} 
and consumed word by word until the end-of-sentence (EOS) token. The listener then passes the last hidden state to $n$ parallel linear layers, one for each aspect of the meaning. 
Finally, each of the $n$ elements is generated through a softmax layer. 
The speaking agent has a mirrored \textit{linear-to-sequence} architecture (see Figure~\ref{fig:spk}). 
% sequences are encoded/decoded by RNNs while meanings are represented by unordered sets of items.

We tested agents based on two recurrent architectures: vanilla recurrent neural networks (RNNs) \citep{elman1990finding} and gated recurrent units (GRUs) \citep{cho-etal-2014-learning}\footnote{The GRU condition was included to test whether the results depend on the agents' memory architecture. GRUs have been commonly used in emergent communication research \citep{dessi2019focus, chaabouni2020compositionality}. Their gating mechanisms, namely update and reset gates, control how much past information is retained or forgotten, making GRUs better at retaining history than vanilla RNNs \citep{can2020gating}.}. However, the GRU architecture was not evaluated in Experiment 3 because agents under the incremental-listener setup did not reach sufficiently reliable communication performance even after extensive hyperparameter tuning.

\subsubsection{Supervised Language Learning} During the initial SL phase, agents learn the predefined artificial language from a dataset of meaning-utterance pairs $\langle\mathbf{m}, \mathbf{u}\rangle$. 
The speaker network ($S$), parameterized by $\theta_S$, is trained to generate the gold-standard utterance $\mathbf{u} = [w_1, ..., w_I]$ by minimizing the cross-entropy loss:
$$Loss_{sup}(S)=-\sum_{i=1}^{I}\log p_{\theta_S}(w_i|\mathbf{w_{<i}},\mathbf{m})$$
Concurrently, the listener network ($L$), parameterized by $\theta_L$, is trained to reconstruct the unordered elements $e$ of the intended meaning $\mathbf{m}$ given the gold utterance $\mathbf{u}$:
$$Loss_{sup}(L)=-\sum_{e\in\mathbf{m}}\log p_{\theta_L}(e|\mathbf{u})$$
where $e \in \{A, a, p, \mu_a^1,\mu_a^2,\mu_a^3, \mu_p^1,\mu_p^2,\mu_p^3\}$ represents a specific semantic role present in the meaning $\mathbf{m}$.

\subsubsection{Optimizing Communication Success by RL} After SL convergence, the agents enter the communication phase where they interact without gold-standard utterances. The speaker generates an utterance $\hat{\mathbf{u}}$ for a meaning $\mathbf{m}$, and the listener attempts to reconstruct $\mathbf{m}$. To simulate communication pressure, the speaker's policy is updated using the REINFORCE algorithm \citep{williams1992simple} to maximize a communicative reward $r_L$. This reward is defined as the negative cross-entropy of the listener's prediction:
$$r_L(\mathbf{m},\hat{\mathbf{u}})=\sum_{e\in\mathbf{m}}\log p_{\theta_L}(e|\hat{\mathbf{u}})$$
The speaker is then updated by minimizing the expected communication loss:
$$Loss_{comm}(S,L)=-r_L(\mathbf{m},\hat{\mathbf{u}})\sum_{i=1}^{I}\log p_{\theta_S}(\hat{w}_i|\mathbf{\hat{w}_{<i}},\mathbf{m})$$
This RL setup allows the language to drift from its original distribution if %alternative word orders (e.g., those minimizing dependency length) 
that yields higher communicative success under processing constraints. % t the language can change in whatever way happens to maximise comm accuracy
\citet{lian2023communication,lian-etal-2024-nellcom} used this framework to simulate the emergence of the word-order/case-marking trade-off and found that a human-like trade-off appears during communication without hard-coding specific biases in the agents.

\subsection{Miniature Languages} 
\label{sec:miniaturelang}

\begin{figure*}[!t]
\centering
  \includegraphics[width=\linewidth]{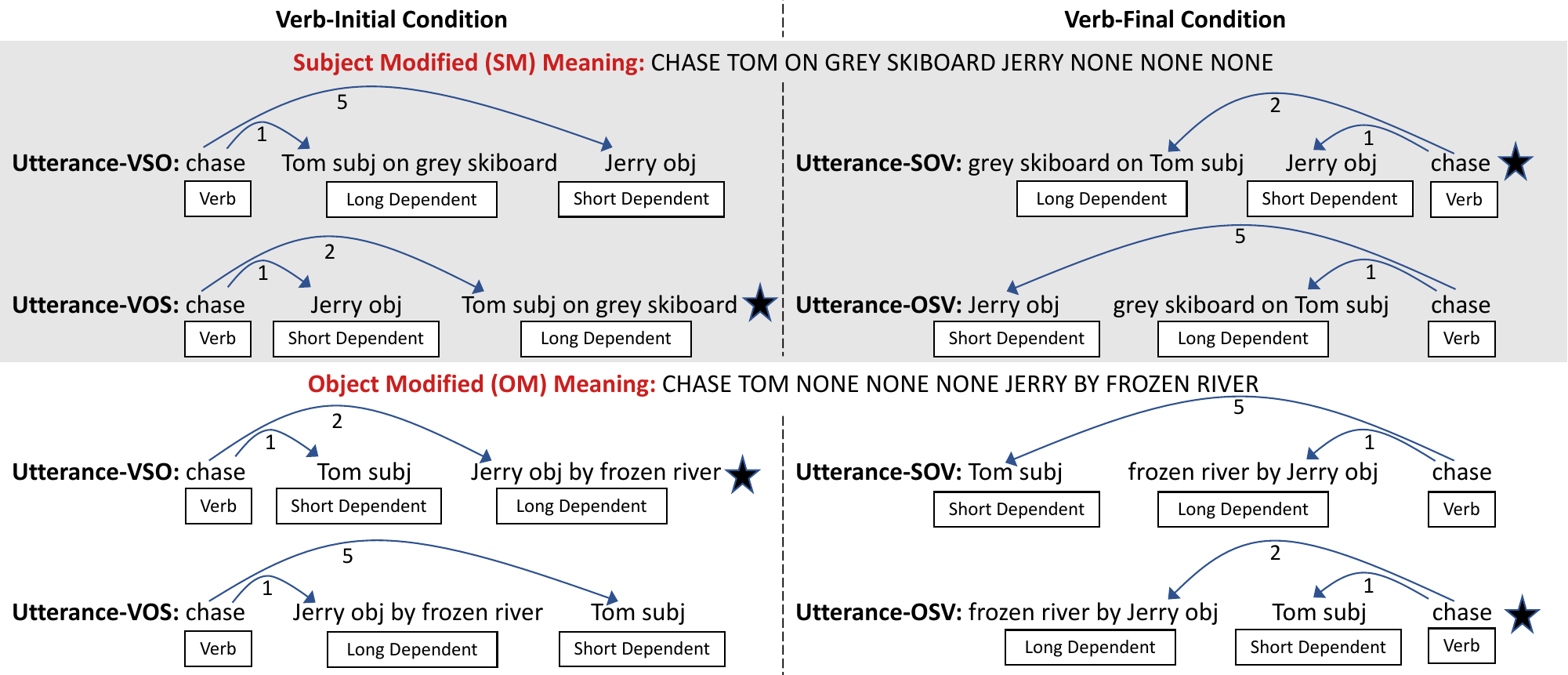}
  \caption{The miniature languages used in our simulations, inspired by \citet{fedzechkina2018human}. Case marking `obj' occurs after all objects, and `subj' occurs after all subjects. 
  Shaded parts contain utterances expressing the same meaning. Curved arrows represent grammatical dependencies between the verb and its two dependents. Numbers represent dependency lengths, measured in words. Since the relative positions of other constituents stay the same for each condition, we operationalized DLM as minimizing the linear distance between the verb and the head of the subject and the head of the direct object. Total dependency length is minimized (starred) by placing short-before-long dependents in the verb-initial language (left), and vice versa by placing long-before-short dependents in the verb-final language (right).
  For simplicity, case markers were not counted when calculating dependency length.
  }
  \label{fig:lang}
\end{figure*}

To study DLM, we exposed the agents to various flexible-order case-marking miniature languages where long- and short-dependency utterances occurred with different distributions (illustrated in Figure~\ref{fig:lang}). 
Directly inspired by the human learning experiments of \citet{fedzechkina2018human}, we designed: 
\begin{enumerate}
\item a \textbf{verb-initial mixed} language that has flexible order (50\% verb-subject-object (VSO) and 50\% verb-object-subject (VOS)), and nouns that can be modified with \textit{post-}nominal prepositional phrases (e.g., \textit{Jerry by frozen river} for the meaning describing the scene \textsc{Jerry by frozen river}; 
\item a \textbf{verb-final mixed} language that also has flexible order (50\% subject-object-verb (SOV) and 50\% object-subject-verb (OSV)), and nouns that can be modified with \textit{pre-}nominal prepositional phrases (e.g., \textit{frozen river by Jerry} for the meaning \textsc{Jerry by frozen river}).
\end{enumerate}

The ordering of the adposition (e.g., \textit{by}) relative to its dependent (e.g., \textit{frozen river}) and head (e.g., \textit{Jerry}) followed the typologically frequent distribution (e.g., \textit{frozen river by Jerry} for \textsc{Jerry by frozen river} in the verb-final language) \citep{fedzechkina2018human, fedzechkinadependency, zhao2022probe}. In addition, all utterances had a fixed length (nine tokens). Although the overall utterance length was fixed, the relative distance between dependents and the verb varied depending on whether the long or short constituent appeared closer to the verb, allowing us to manipulate dependency length in a controlled manner. Fixing utterance length in this way also avoided potential confounds arising from sentence length, which has been shown to influence the visibility of DLM effects in natural language data \citep{ferrer39placement, ferrer-i-cancho_anti_2021, ferrer-i-cancho_optimality_2022, ferrer-i-cancho_predictability_2024, ferrer-i-cancho_optimal_2025}. 

The dependency structure of the artificial languages used in our simulations was relatively simple: the verb functioned as the head, while the subject and object were its direct dependents. Modifier phrases attached within noun phrases, so within each comparison, both the dependency topology and the sequence length remained constant, and the manipulation affected only the linear distance between the verb and the heads of its subject/object dependents.
Previous work has shown that, in dependency structures where one central head is directly connected to all other words in a short sequence (i.e., simple star-like structures), predictability maximization may favor orders that diverge from DLM
\citep{ferrer39placement, ferrer-i-cancho_predictability_2024}. Our two-level hierarchical design reduced this potential competition and allowed us to examine ordering preferences under controlled structural configurations and processing pressures.
In addition, our simulations abstracted away from factors such as \textit{animacy first} and \textit{lexical grouping} tendencies. The neural agents did not encode animate-inanimate distinctions, and the experimental design did not favor clustering of similar lexical items, which allowed us to isolate the contribution of processing constraints.
% ensuring that any observed ordering preferences cannot be attributed to these alternative explanations.

Different from the setup in \citet{fedzechkina2018human} and \citet{yuqing-2024-dlm}, both subject and object constituents were marked for their grammatical roles. 
For readability, Figure~\ref{fig:lang} displays these labels as post-nominal markers, with \textit{`subj'} following subjects and \textit{`obj'} following objects. 
However, these labels were not separate tokens in the agents' input or output sequences. 
Instead, each animate noun had two role-specific lexical forms, one used as a subject and one used as an object; thus, a displayed form such as \textit{Tom subj} corresponded to a single atomic symbol in the model vocabulary.
This modification to the setup of \citet{yuqing-2024-dlm} (where only object markers were used) served to test the consistency of prior findings. 
An additional reason for this adjustment arose from observations in our preliminary experiments: when only one marker was present, neural agents preferred to place the constituent with the marker at the very beginning, potentially confounding any underlying preference for DLM. The presence of both case markers ensured the two constituents were equally unambiguous, informative, and of equal weight. 
Given the architectural design, meanings could be regarded as a combination of \textit{unordered} slots in a semantic tree, thus avoiding a trivial tendency of the agents to minimize reordering with respect to the input meaning representation. 
%main experiment 
In all experiments, the meanings described scenes containing only one long constituent (i.e., either the agent or the patient has adpositional-phrase modification). For each meaning, there were two possible utterance orderings (SO or OS). For all \textit{short} languages, 
% i.e., \textit{skewed short} and \textit{uniform short} --- but these have not been introduced yet
utterances were generated by placing \textit{short} dependents closer to the head verb (see Figure~\ref{fig:lang}, starred utterances). Correspondingly, for all \textit{long} languages, utterances were generated by placing \textit{long} dependents closer to the head verb (see Figure~\ref{fig:lang}, non-starred utterances). For languages \textit{mixed} with both short and long dependency utterances, we randomly sampled half of the meanings and generated short dependency utterances, and then generated long dependency utterances for the other half of the randomly sampled meanings. The subject and object were never the same within a single meaning. 
%%%%%%%TODO UPDATE MANUSCRIPT

\subsection{Conditional Word Distributions}
\label{con_distr}

% co-occurrence probability (or mutual information)
Artificial language learning experiments with human participants or neural agents typically assume meaning spaces where all items are uniformly distributed \citep{fedzechkina2016miniature, lian2023communication}. In reality, however, human languages exhibit highly skewed word frequency distributions resembling Zipfian patterns \citep{zipf1949human}. %, i.e., the frequency of any word is inversely proportional to its rank: $p(w) \propto \frac{1}{{rank(w)}}$. 
Moreover, certain word combinations tend to co-occur much more frequently than others, an especially well-known tendency for verbs and their arguments, called selectional preference \citep{katz-fodor-63}. More generally, \citet{futrell2019syntactic} demonstrated that syntactic heads and their dependents are characterized by word pairs with especially high mutual information. Building upon these insights regarding statistical co-occurrence patterns in syntactic dependency structures, we incorporate non-uniform word distributions and selectional preferences of verbs, i.e., the conditional probability of the dependents given the presence of the head verb, into our miniature artificial language design.

Incorporating more realistic input distributions in artificial language learning simulations may be crucial to ensure the generalizability of findings to real languages \citep{hupkes2019compositionality}. % (Sect 6.1.1 Naturalisation of structural properties)
In our context particularly, endowing head-dependent pairs with realistic statistical properties may be crucial to simulate the emergence of DLM. 
In fact, while human subjects have a pre-existing notion of head-dependent relations between concepts and are likely to recognize such relations in the novel artificial language provided in the lab, our neural learners have no prior experience of language or the world and simply perceive sequences of random symbols at the beginning of training. 
By introducing verb-subject and verb-object selectional preferences in our language design, we tested whether shorter dependencies could strengthen the listener's ability to recover missing bits of a sentence by relying on the presence of frequently co-occurring words in the nearby context.

Concretely, we constructed a \textbf{skewed} and a \textbf{uniform} version of the verb-initial and verb-final languages.
In the skewed version, we accounted for the tendency of certain animate nouns to co-occur more frequently with specific verbs, as well as their varying probabilities of taking agent or patient roles given each transitive verb. Thus, animate nouns had different probabilities of being agent or patient \textit{given} a particular verb. For simplicity, all meaning items other than agent and patient occurred with uniform probability. 
To generate the skewed version of each language, we used discrete distributions that roughly corresponded to equally spaced sample points from the Zipf distribution \citep{zipf1949human}: 

\begin{equation}
f(k; s, N)=\frac{1}{H_{N, s}} \frac{1}{k^s}
\label{zipf_eq}
\end{equation}

\noindent
where $N$ stands for the number of elements; $k$ is the rank counting from 1; $s$ is the exponent parameter; $H_{N, s}$ is a generalized harmonic number. In the experiments, we used $s=0$ for the uniform condition and $s=1.5$ for the skewed condition. While large adult vocabularies typically exhibited an exponent closer to 1.0, heavily restricted communication systems (e.g., early child language) often displayed steeper Zipfian distributions \citep{baixeries2013evolution}. Applying $s=1.5$ to our 32-word vocabulary ensured that frequency gradients and selectional preferences remained highly salient, providing untrained neural agents with a statistically robust signal to bootstrap syntactic dependencies.
To have a sufficient meaning space to train neural agents, we expanded the lexicon to ten transitive verbs, ten animate nouns, four adpositions, three adjectives, and three inanimate nouns, leading to a vocabulary size $|V_{2}|$ of $10+10+4+3+3+2$ (markers) $=32$. 

\subsection{Noisy Communication}

Real-life communication between speakers and listeners is often subject to various sources of noise and errors that occur during transmission, for instance, due to external factors or limited listener attention \citep{gibson2013rational, brochhagen2017effects}. % partially attentive listeners
Noisy communication makes it challenging for listeners to accurately perceive every word spoken by the speaker. Moreover, listeners may not always maintain full attention, resulting in further inaccuracies in communication. 

Noise in the agents’ environment has been proposed as an important factor in emergent communication by previous simulation studies \citep{ueda-washio-2021-relationship, chaabouni2022emergent, kucinski2021catalytic, tucker2021emergent}.
In particular, \citet{ueda-washio-2021-relationship} implemented noise on the speaker side, listener side, and channel side. They found that noise imposing internal difficulty on a speaker agent may cause Zipf’s law of abbreviation. \citet{kucinski2021catalytic} highlight the catalytic
role of noise in the emergence of compositionality.
Notably, \citet{futrell-levy-2017-noisy} found that a model of sentence processing in which the context was noisy predicted that sentences would be easier to process when words in syntactic dependencies were close. Based on these positive results in previous agent-based simulations, we hypothesized that noise could play an important role in the emergence of word order preference.
To simulate noise, we adopted a \textbf{word dropout} technique \citep{gal2016theoretically}. Formally, we applied an independent Bernoulli random variable $\delta_i$ to each token in the sequence:
% with dropout rate $\delta$ by which randomly chosen parts of the input tokens are masked to the listener:

\begin{equation}
    \delta_i \sim \text{Bernoulli}(p), \quad i \in \{1, \dots, L\}
\end{equation}
% \begin{equation}
%     \delta \sim \text{Bernoulli}(p), \quad p \in \{0, 0.1\} 
% \end{equation}
where $p \in \{0, 0.1\}$ represented the constant dropout probability. If $\delta_i = 1$, the embedding of the $i$-th token was replaced by that of a dummy \textsc{mask} token, which was not updated during training \citep{gal2016theoretically, sennrich2016edinburgh}.
Crucially, all agent-generated utterances were fixed at $L=9$ tokens. 
% Following \citet{lian2023communication}, if the model generates a complete grammatical structure before reaching this limit, the sequence is padded to length 9. 
Applying Bernoulli word dropout uniformly across the sequence yielded an expected 0.9 masked tokens per utterance. 
% Because the Bernoulli word dropout is applied uniformly across this fixed-length sequence, the expected number of masked tokens per utterance is mathematically constant across all communication rounds. %: $\mathbb{E}[\sum_{i=1}^L \delta_i] = 9 \times 0.1 = 0.9$. 
% This guarantees that, on average, approximately one token is masked per sequence. 
Preliminary experiments confirmed that this 10\% masking probability ($p=0.1$) provided a sufficient communicative bottleneck pressure to influence processing strategies without causing the catastrophic signal degradation that prevented RL convergence.
% Specifically, words that are sampled by a Bernoulli distribution (with probability $p$ set to either 0 or 0.1 in our setup) get their word embedding replaced by that of a dummy \textsc{mask} token, which is not updated during training \cite{gal2016theoretically, sennrich2016edinburgh}.
% or zero embedding 
%a technique that introduces perturbations to word embeddings using a dropout mask sampled from a Bernoulli distribution $Ber(1 - p)$, in which 'p' represents the dropout probability (i.e., specific elements of the input are substituted with a [MASK] token, whose embeddings remain fixed throughout the entire process. The substitution is determined by a Bernoulli distribution with probability values ranging from 0 to 0.2 in our setting).
While probabilistic dropout was commonly used to avoid overfitting when training neural networks, we applied it here to both training \textit{and} testing to simulate what listeners would encounter during their learning and communication. 
This approach allowed us to more accurately replicate real-world communication scenarios, in which listeners must rely on various cues and context to comprehend the speaker's message under imperfect transmission conditions. % recover the intended meaning

% \AB{TODO find a place to clarify our expectations of interplay between noise and DLM}
% whether RNN  do incrementally process info 
% (during training, they are aware of this co- earlier) skewed: faster rise in the acc
% not within the utterance

% add general expectation!: Noise - condition prob leads to advantages to guess missing words --rnn, short dl, less intermediate materials due to forgetting
In general, it was expected that neural networks, especially RNNs, benefited from shorter dependency lengths due to reduced intermediate contents and memory effects, such as forgetting. The presence of noise in training conditions, by introducing uncertainty in word prediction, may also encourage these networks to develop preferences for shorter dependencies to facilitate more efficient processing.
However, in certain language typologies, other factors may counterbalance this effect. For instance, in verb-final languages, placing a constituent at the beginning of an utterance may improve information retrieval or utterance predictability, even if it results in longer dependencies.
% For instance, in verb-initial languages, shorter dependencies are anticipated to support better information retrieval, whereas in verb-final languages, longer dependencies may optimize utterance understanding by placing a constituent at the beginning, reducing the need for intermediate processing steps.

% (Hinton et al., 2012; Srivastava et al., 2014) (Sennrich et al., 
% cite: Learning to Perturb Word Embeddings for Out-of-distribution QA

\subsection{Limited Speaker Capacity}
Neural network architectural features, such as depth, width, and the number of parameters, have been shown to affect the model's capacity to process and represent linguistic phenomena efficiently \citep{mueller2023plant}, possibly influencing the emergence of linguistic preferences like DLM.
Recent work on Transformer-based models of sentence processing further suggests that larger models are not necessarily more cognitively plausible, and that introducing recency-based limitations can improve alignment with human reading behavior \citep{oh-schuler-2023-surprisal, clark-etal-2025-linear}. These findings motivated our use of model capacity as a simplified operational analogue of processing limitations.
In this study, we constrained the speaker’s capacity by limiting the size of hidden layers in the speaker network. 
Specifically, in Experiment 2, we compared a larger speaker model with hidden size $H=512$ to a reduced speaker model with hidden size $H=256$. 
This reduced-capacity setting was selected through preliminary parameter searches comparing $H=512$, $H=256$, and $H=128$ under the same training setup. 
We used two criteria for this selection. 
First, the reduced model had to maintain successful communication, so that any change in word order could not simply be attributed to failure to learn the task. 
Second, among models that communicated successfully, we looked for stable differences in production behavior, by which we mean differences in the distribution of word orders produced by the speaker. 
The $H=128$ model reduced communicative accuracy too strongly and produced less stable behavior, making its word order patterns difficult to interpret. 
By contrast, the $H=256$ model retained sufficient communicative accuracy while showing production patterns that diverged from the larger $H=512$ model. 
We therefore used $H=256$ as the limited capacity condition.

% a short motivation
Our motivation stems from cognitive constraints observed in human communication. Limited memory and processing resources make it difficult for speakers to track all potential word combinations or variations when communicating with partners. Such constraints can naturally promote the use of efficient language structures \citep{verhof2011cultural, cornish2017sequence, isbilen2020chunk}. 
% various sources from the early kirby et al. studies and morten christiansen % cornish2017sequence，
By analogy, we hypothesized that reducing a speaker model’s capacity could simulate such cognitive limitations and serve as an architectural bottleneck, thereby promoting the emergence of more efficient linguistic patterns, such as shorter dependency lengths, during the evolution of communication protocols.

\subsection{Incremental Processing}
\label{sec:impa_arch}

Human language comprehension is often characterized by the prediction mechanism \citep{altmann2009incrementality} by which human listeners process incoming input sentences incrementally without delay and seek to guess the intended content when the input is still incomplete (i.e., impatient listener) \citep{kamide2003time, altmann2009incrementality}. 
In the context of agent communication, incremental accuracy measured the extent to which an utterance's meaning could be anticipated by a listening agent before the utterance had been fully processed.
In Experiment 3 (\S~\ref{sec:exp3}), we incorporated incremental accuracy during training as a component of the reward function to optimize speaking and listening agents. 
Our motivation was that rewarding utterances that could be partially understood as they unfold in real time might promote linearizations that facilitated incremental interpretation, often linked to shorter dependency lengths. This approach encouraged word orders that reduced processing demands, potentially favoring structures that minimized dependency length in line with efficiency pressures observed in natural languages.
Specifically, we adapted the Impatient Listener architecture proposed by \citet{rita-etal-2020-lazimpa}, which modeled the prediction mechanism by allowing the listener to reconstruct the intended meaning as soon as possible during communication. Instead of guessing the set of meaning items after consuming the entire utterance $\hat{\mathbf{u}} = (\hat{u_{0}}, ..., \hat{u_{t}})$ produced by the Speaker, the Impatient Listener predicted the right meaning candidate as soon as possible when processing each symbol $\hat{u_{k}}$ during communication, where $k$ denoted the length of the utterance. 
% t is the moving index

% At supervised training, a prediction of Impatient Listener, at a position k, is a Categorical distribution $L(u_{:k})$, constructed using a shared n parallel linear layers followed
% by softmax layers (with $L(u_{:k}) = (u_{0}, ..., u_{k})$). Eventually, we get a sequence of $t + 1$ distributions $L(u) =
% (L(u_{0}), ..., L(u_{t}))$, one for each reading position of
% the message.

\begin{figure}[ht]
\centering
    \includegraphics[width=0.4\linewidth]{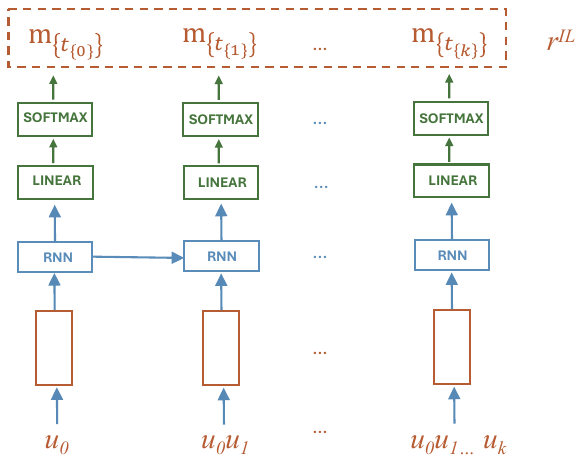}
    \caption{Impatient listening agent architecture. 
    }
    \label{fig:impalst}
\end{figure}

\textbf{Impatient Loss:\ }
During communication training, a prediction of an Impatient Listener, at a position $k$, was a Categorical distribution $L(\hat{\mathbf{u}}_{0:k})$, constructed using $n$ parallel linear layers followed
by softmax layers (with $\hat{\mathbf{u}}_{0:k} = (\hat{u}_{0}, ..., \hat{u}_{k})$). Eventually, we got a sequence of $t + 1$ distributions $L(\hat{\mathbf{u}}) =
(L(\hat{u}_{0}), ..., L(\hat{u}_{t}))$, one for each reading position of the message (see Figure~\ref{fig:impalst}).

In this case, optimizing communication success was equivalent to minimizing the impatient loss, defined as the average cross-entropy loss between the ground-truth input meaning $\mathbf{m}$ and all of the listener’s intermediate predictions 
$L(\hat{\mathbf{u}})=(L(\hat{\mathbf{u}}_{0:0}), \ldots, L(\hat{\mathbf{u}}_{0:t}))$. Both agents received higher rewards when the speaking agent's production yielded higher incremental listening accuracy, as shown in the following reward functions. 
Specifically, at each timestep $t$, we calculated a reward $r^L(m, \hat{\mathbf{u}}_{0:t})$, and then averaged these rewards over all timesteps.

The original reward $r^L(\mathbf{m}, \hat{\mathbf{u}})$ defined in NeLLCom and used in our standard setting was: 
\begin{equation}
r^L(\mathbf{m}, \hat{\mathbf{u}}_{0:k})=\sum_{e \in \mathbf{m}=\{A, a, p, \mu_a, \mu_p\}} \log p_{\theta_L}(e \mid \hat{\mathbf{u}}_{0:k})
\end{equation}

To incorporate incremental comprehension, we defined the incremental reward $r^{IL}$ as the average listener reward over all prefixes of the utterance:

\begin{equation}
r^{IL} = \frac{1}{k+1} \sum_{t=0}^{k} r^L(\mathbf{m}, \hat{\mathbf{u}}_{0:t})
\label{incre_r}
\end{equation}

Here, \( k \) denoted the length of the utterance \( \hat{\mathbf{u}} \), and \( \hat{\mathbf{u}}_{0:t} \) denoted the prefix of the utterance up to timestep \( t \).
Hence, all intermediate distributions contributed to the loss function: earlier correct predictions yielded larger communication rewards.

\subsection{Evaluation Metrics}

During the SL phase, \textbf{speaking accuracy} was computed based on whether a generated utterance fully matched the one defined by the artificial language grammar (acc=1) or not (acc=0) for any given meaning. For languages that combined short and long dependencies and therefore had two acceptable utterances to express the same meaning, we counted a response as correct (1) if it matched either of the two gold utterances; otherwise, it was counted as incorrect (0) \footnote{This corresponded to the \textit{permissive accuracy} approach used to evaluate speakers in \citet{lian2023communication}.}. 
\textbf{Listening} agents were evaluated based on two metrics: \textbf{whole-meaning accuracy}, and meaning \textbf{item-level accuracy}. We calculated item-level accuracy by dividing the number of correctly predicted meaning items by the total number of items in the ground-truth meaning, averaged over all test utterances.
During the RL phase, we instead measured \textbf{communication success} through meaning reconstruction accuracy, which corresponded to the listener's accuracy in recovering the intended meaning from a speaker-generated utterance.    
Agents were always assessed on a reserved set of meanings that were unseen during any training phase. 
All plotted accuracies were computed over the unseen test set and averaged over all random seeds unless otherwise stated.

%Per-category listening accuracy is calculated for all meaning categories ($\{A,a,p, \mu_a, \mu_p\}$). 

% TODO or N/A 
% To perform statistical significance testing regarding the experimental measures, we fit Bayesian multilevel linear models using the \texttt{brms} \cite{burkner2017brms} package in R \citep{r}. Our data involves complex structures where multiple observations of accuracy measures are nested within various experimental conditions and these models provide a robust framework for analyzing such data. We investigate the impact of epochs (centered at the final epoch), conditional word distributions (categorical: skew or uniform), noise levels (categorical: dropout 0, 0.1, or 0.2), and dependency conditions (categorical: short, mixed, or long) on both speaking and listening accuracy. Values for the epoch were centered so that the intercept is focused on agents' learning performance at the end, thereby ensuring that measured effects represent relationships between variables concerning the final stable stage. Therefore, in our statistical analysis, the reference point of epoch 0 represents the final point of the epochs instead of the first epoch. Importantly, we accounted for all possible interaction terms among these factors to capture relationships within the data. Default priors of the \texttt{brms} package were used for all models. All models included four chains with 4,000 iterations each and a total of 2,000 post-warm-up samples.

To perform statistical significance testing of the experimental measures, we fitted Bayesian multilevel linear models using the \texttt{brms} package \citep{burkner2017brms} in R \citep{r}. 
For accuracy measures, our data involved complex structures where multiple observations were nested within various experimental conditions, and these models provided a robust framework for analyzing such data. We investigated the impact of epochs (centered at the final epoch), conditional word distributions (categorical: skew or uniform), noise levels (categorical: dropout 0, 0.1), and dependency conditions (categorical: short, mixed, or long) on both speaking and listening accuracy. Values for epoch were centered so that the intercept reflected agents' learning performance at the end, thereby ensuring that measured effects represented relationships between variables concerning the final stable stage. Therefore, in our statistical analysis, the reference point of epoch 0 represented the final point of the epochs instead of the first epoch. Importantly, we also included all possible interaction terms among these factors to capture relationships within the data. 
% % We converted this proportion to a percentage:
% \(P_{\%}(\mathrm{local}) =
% 100 \times N_{\mathrm{local}} /
% (N_{\mathrm{local}} + N_{\mathrm{long}})\).
To directly assess whether agents exhibited DLM, we examined the proportion of short-dependency utterances within each condition. 
Productions that could not be classified as either local- or long-dependency utterances were excluded from the denominator. We therefore calculated the local-dependency proportion as 
\(P_{\%}(\mathrm{local}) =
100 \times N_{\mathrm{local}} /
(N_{\mathrm{local}} + N_{\mathrm{long}})\),
such that 50\% indicated equal proportions of the two dependency-length alternatives.
For this, separate Bayesian multilevel models were fitted for each communication stage, treating condition as a fixed effect and agent seed as a random intercept. In this framework, the intercept corresponded to the average proportion of short-dependency utterances in the reference condition. To determine whether agents significantly preferred short dependencies over the chance level (50\%), we explicitly tested the hypothesis that the posterior distribution of the mean short-dependency proportion was greater than 50 using the \texttt{hypothesis()} function in \texttt{brms} \citep{burkner2017brms}. 
Pairwise contrasts between conditions were obtained by re-leveling the condition factor within the model and applying the same hypothesis-testing approach. This allowed us to quantify both absolute effects (preference for short dependencies vs. chance) and relative effects (differences between conditions) within the same Bayesian framework.
% No additional multiple-comparison correction was applied, as Bayesian inference relies on posterior distributions and credible intervals rather than frequentist p-values. Posterior summaries include the mean proportion of short-dependency utterances and 95\% credible intervals, providing a clear probabilistic interpretation of the effects.
% We also compared the proportion of short-dependency utterances across conditions and communication phases using Bayesian multilevel regression models. Separate models were fitted for each phase (comm0, comm30, comm60), with condition as a fixed effect and random intercepts for seed number. Pairwise contrasts were obtained by re-leveling the condition factor and via the hypothesis function in \texttt{brms} \citep{burkner2017brms}. No additional multiple-comparison correction was applied, as Bayesian inference relies on posterior distributions. 
% % We report posterior means and 95\% credible intervals, interpreting effects by the proportion of posterior mass above or below zero.
The default priors of the \texttt{brms} package were used for all models. All models included four chains with 4,000 iterations each and a total of 2,000 post-warm-up samples.
This approach ensured that our analysis directly addressed whether agents produced a significantly higher proportion of short dependencies than expected by chance, and whether this preference varied across experimental conditions or phases.

\subsection{Training Details}

We trained the agents on six different language types (\textit{skewed short, skewed mixed, skewed long, uniform short, uniform mixed, uniform long}) to study the interplay between DLM and the above-mentioned factors. Language names denoted the exact proportion of local-dependency (DLM-optimal) utterances in the training data: 100\% in \textit{short}, 0\% in \textit{long}, and exactly 50\% in \textit{mixed}. Because the \textit{mixed} condition provided a perfectly neutral baseline with no statistical bias toward DLM during SL, we focused our analysis primarily on the \textit{skewed mixed} languages. Here, deviation from the 50\% baseline during the communication phase indicated a structural preference actively developed by the agents, rather than one memorized from the input distribution.

To evaluate the agents' ability to convey new meanings, we randomly split the dataset into 66.7\% for training and 33.3\% for testing. Within both the training and testing phases, we ensured an equal distribution of utterances with long and short dependencies. 
We trained both speaking and listening agents for 60 epochs each in supervised learning, followed by 60 epochs dedicated to communication. The maximum production length was set to 9.
% , one unit longer than the ground-truth length. 

Experiment 1 was conducted with 20 random seeds per condition and later expanded to 40 random seeds per condition in experiments 2 and 3 to ensure robust and reliable results. Training details are provided in the respective sections.

\subsection{Meaning Space}
\label{sec:meaning_space}

In Experiment 1, we used the whole meaning space (half describing subject-modified scenes, and the other half describing object-modified scenes) shown in Figure~\ref{fig:lang} to generate the artificial languages. In other words, for all languages, half of the meanings were subject-modified (SM), while the other half were object-modified (OM). Therefore, a meaning was defined by nine items ($\{A,a,p, \mu_a^1,\mu_a^2,\mu_a^3, \mu_p^1,\mu_p^2,\mu_p^3\}$, % $n=9$ 
representing Action, agent, patient, three agent modifiers, and three patient modifiers respectively ($n=9$). The filler was set as NONE if that aspect of the meaning was absent. 
In Experiment 1, approximately 5000 meaning-utterance pairs were generated for each language. 

Experiments 2 and 3 were simplified versions of Experiment 1. In these follow-up experiments, each language consistently had only one entity being modified, rather than alternating between SM and OM meanings. This allowed us to reduce the meaning space to six items ($n=6$): for SM meanings, the slots were $A, a, p, \mu_a^1,\mu_a^2,\mu_a^3$; for OM meanings, the slots were $A, a, p, \mu_p^1,\mu_p^2,\mu_p^3$. \footnote{We also experimented with graph convolutional networks (GCNs) \citep{kipf2017semi} to model dependency relationships within the language structures. In this approach, linguistic elements were represented as nodes and the dependencies between them as edges, allowing information to flow through the network based on these established connections. However, our preliminary results showed overall similar performance and trends as those observed in the original models, primarily due to the simplicity of our language designs, which included only simple modifier-noun and verb-argument dependency patterns.} 
In experiments 2 and 3, a total of 2543 meaning-utterance pairs were generated for each language. 
Data generation details and distributions were provided in Appendix~\ref{app:date_gen}.

\subsection{Summary of Experiments}
Before presenting the detailed experiments, we provide a schematic overview of the three experiments in Table~\ref{tab:experiment-overview}, including their meaning spaces, architectures, manipulations, and main findings. The experiments were designed to progressively isolate the factors that shaped production preferences in neural-agent communication. 
In Experiment 1, word order regularization overrides DLM in the full meaning-space setup; in Experiment 2, a short-before-long ordering bias overrides DLM in the half meaning-space setup; and in Experiment 3, explicitly modeling incremental processing through the Impatient Listener elicits DLM production preferences in both verb-initial and verb-final languages.

\begin{table*}[h!]
\centering
\begin{threeparttable}
\caption{Schematic overview of the three experiments, including their meaning spaces, architectures, manipulations, and main findings.}
\label{tab:experiment-overview}
\scriptsize
\renewcommand{\arraystretch}{1}
\setlength{\tabcolsep}{5pt}

\begin{tabularx}{\textwidth}{
>{\raggedright\arraybackslash}p{0.09\textwidth}
>{\raggedright\arraybackslash}p{0.17\textwidth}
>{\raggedright\arraybackslash}p{0.09\textwidth}
>{\raggedright\arraybackslash}p{0.19\textwidth}
Y
}
\toprule
\textbf{Experiment} &
\textbf{Meaning space} &
\textbf{Architecture} &
\textbf{Manipulations} &
\textbf{Findings} \\
\hline

\cellcolor{lightblue}
{\color{expblue}\textbf{Exp. 1}}
&
\cellcolor{lightblue}
\vspace{0.2em}
\textbf{Full meaning space}\par
\cellitems{
  \item SM $+$ OM
} 
&
\cellcolor{lightblue}
\cellitems{
  \item RNN
  \item GRU\tnote{a}
}
&
\cellcolor{lightblue}
\cellitemsfirst{
  \item Baseline
  \item Noise
}
&
\cellcolor{lightblue}
\vspace{0.2em}
\textbf{Word order regularization overrides DLM}
\cellitems{
  \item Individual seeds converge on one order
  \item Average local-dependency proportion remains near 50\%
}
\\
\hline

\cellcolor{lightgreen}
{\color{expgreen}\textbf{Exp. 2}}
&
\cellcolor{lightgreen}
\vspace{0.2em}
\textbf{Half meaning space}\par
\cellitems{
  \item SM only or OM only
}
&
\cellcolor{lightgreen}
\cellitems{
  \item RNN
  \item GRU
}

&
\cellcolor{lightgreen}
\cellitemsfirst{
  \item Baseline
  \item Noise
  \item Lower speaker capacity
  \item Lower speaker capacity + Noise
}
&
\cellcolor{lightgreen}
\vspace{0.2em}
\textbf{Short-before-long overrides DLM}
\cellitems{
  % \item Verb-initial languages show more local dependencies
  % \item Verb-final languages show more long dependencies
    \item Agents develop a short-before-long ordering bias
    \item This bias is DLM-compatible in verb-initial languages but anti-DLM in verb-final languages
    \item Noise and lower speaker capacity tend to amplify the bias
}
\\
\hline

\cellcolor{lightpurple}
{\color{exppurple}\textbf{Exp. 3}}
&
\cellcolor{lightpurple}
\vspace{0.2em}
\textbf{Half meaning space}\par
\cellitems{
  \item SM only or OM only
}
&
\cellcolor{lightpurple}
\cellitems{
  \item RNN
}
&
\cellcolor{lightpurple}
\cellitemsfirst{
  \item Baseline
  \item Impatient listener
  \item Impatient listener + Noise
}
&
\cellcolor{lightpurple}
\vspace{0.2em}
\textbf{Incremental processing elicits DLM}
\cellitems{
  \item DLM preference emerges in both verb-initial and verb-final languages
}
\\
\bottomrule
\end{tabularx}

\begin{tablenotes}[flushleft]
\scriptsize
\item[a] The GRU architecture in Experiment 1 follows the same setup as \citet{yuqing-2024-dlm} and reproduces the same ordering patterns.
\end{tablenotes}

\end{threeparttable}
\end{table*}

\section{Experiment 1: Word Order Regularization Overrides DLM}
\label{sec:exp1}

% In our initial experimental setup, the design includes both subject-modified and object-modified meaning spaces. Production preferences for DLM means producing two distinct word orders for different meaning types.
% We find that RNN-based speakers and listeners show no learning advantage for the shorter-dependency languages over longer-dependency counterparts, in terms of both learning speed and final accuracy. Both speakers and listeners reach perfect accuracy after several epochs. During communication, neural learners fail to display a DLM preference. Instead, they tend to regularize toward a single word order rather than optimizing for dependency length, indicating that word order entropy minimization is a stronger competing pressure \citep{kharitonov2020entropy}.

% To isolate the effects of DLM, we simplified the design by separating meaning types, thus removing the competing pressure from word order entropy minimization. Under these conditions, neural agents show a preference for reducing dependency length in production, particularly when processing limitations are strengthened in learning verb-initial languages.

The setup of Experiment 1 matched the overview outlined in \S~\ref{sec:miniaturelang}, \ref{con_distr} and \ref{sec:meaning_space},  and followed the setup of \cite{zhang2024neural} using the complete meaning space. 
As the only difference, we included both the subject and the object markers to rule out a general agent preference for generating the marked constituent first (cf. \S~\ref{sec:miniaturelang}).
Training details were summarized in Table~\ref{tab:training_setup}.

\begin{table}[h!]
\centering
\begin{tabular}{ll}
\toprule
\textbf{Component}        & \textbf{Details}                      \\ 
\midrule
Speaker Network   & 16-dimensional embedding layer, 1024-dimensional hidden layer \\ 
Listener Network  & 512-dimensional embedding layer, 1024-dimensional hidden layer \\ 
Batch Size       & 32                                    \\ 
Maximum Epochs   & 60                                    \\ 
Learning Rate    & 0.0001                                \\ 
Phases           & Supervised learning and communication  \\ 
\bottomrule
\end{tabular}
\caption{Training setup details for Experiment 1.}
\label{tab:training_setup}
\end{table}

\subsection{Communication Success and Production Preferences}

\begin{figure}[ht]
    \centering

    % =====================================================
    % Left: Communication accuracy
    % =====================================================
    \begin{subfigure}[b]{0.48\textwidth}
        \centering
        \includegraphics[width=\textwidth]{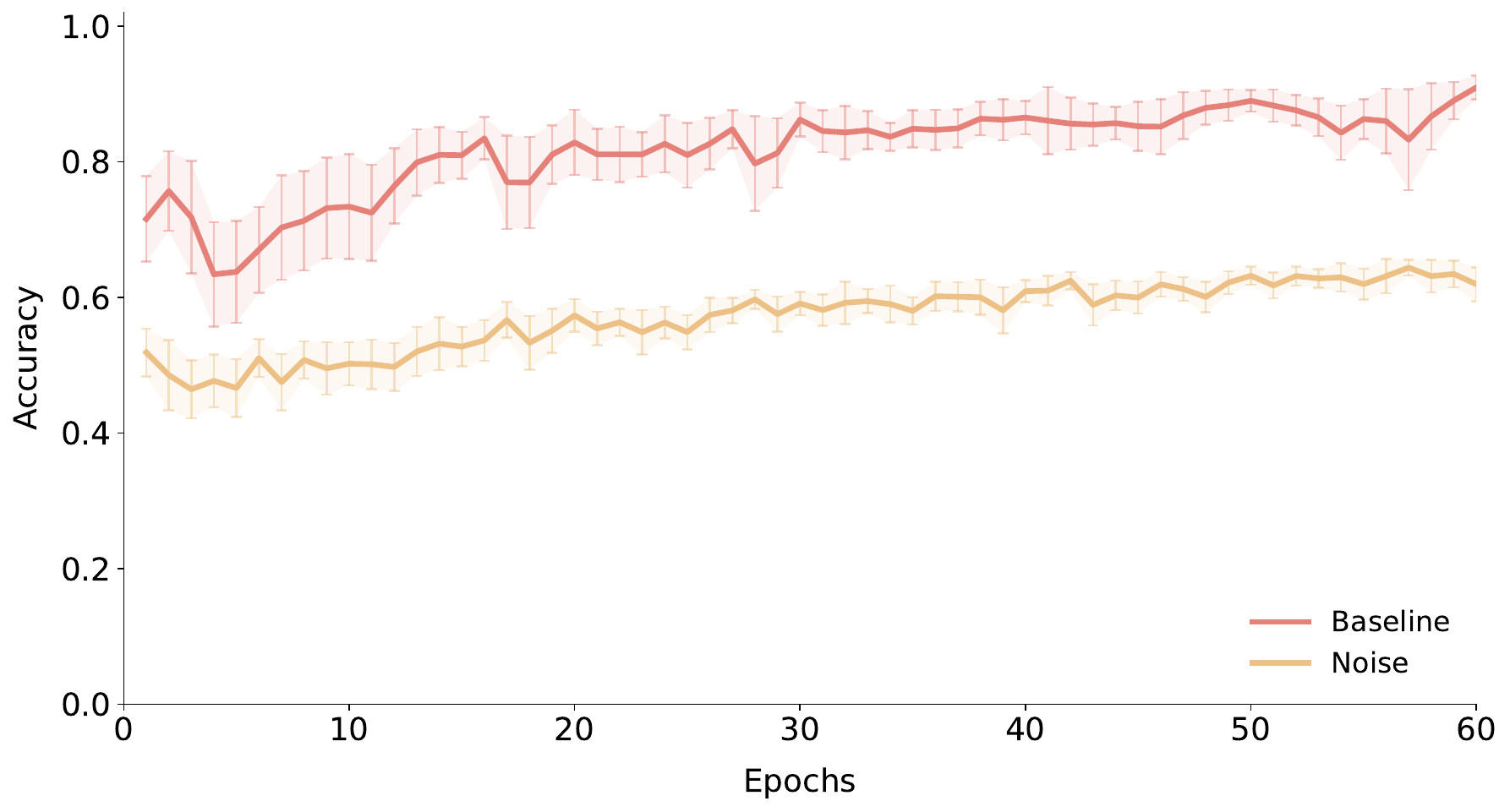}
        \caption{Communication accuracy}
        \label{fig:comm_initial_rnn_ALL_acc}
    \end{subfigure}
    \hfill
    % =====================================================
    % Right: Production preference
    % =====================================================
    \begin{subfigure}[b]{0.48\textwidth}
        \centering
        \includegraphics[width=\textwidth]{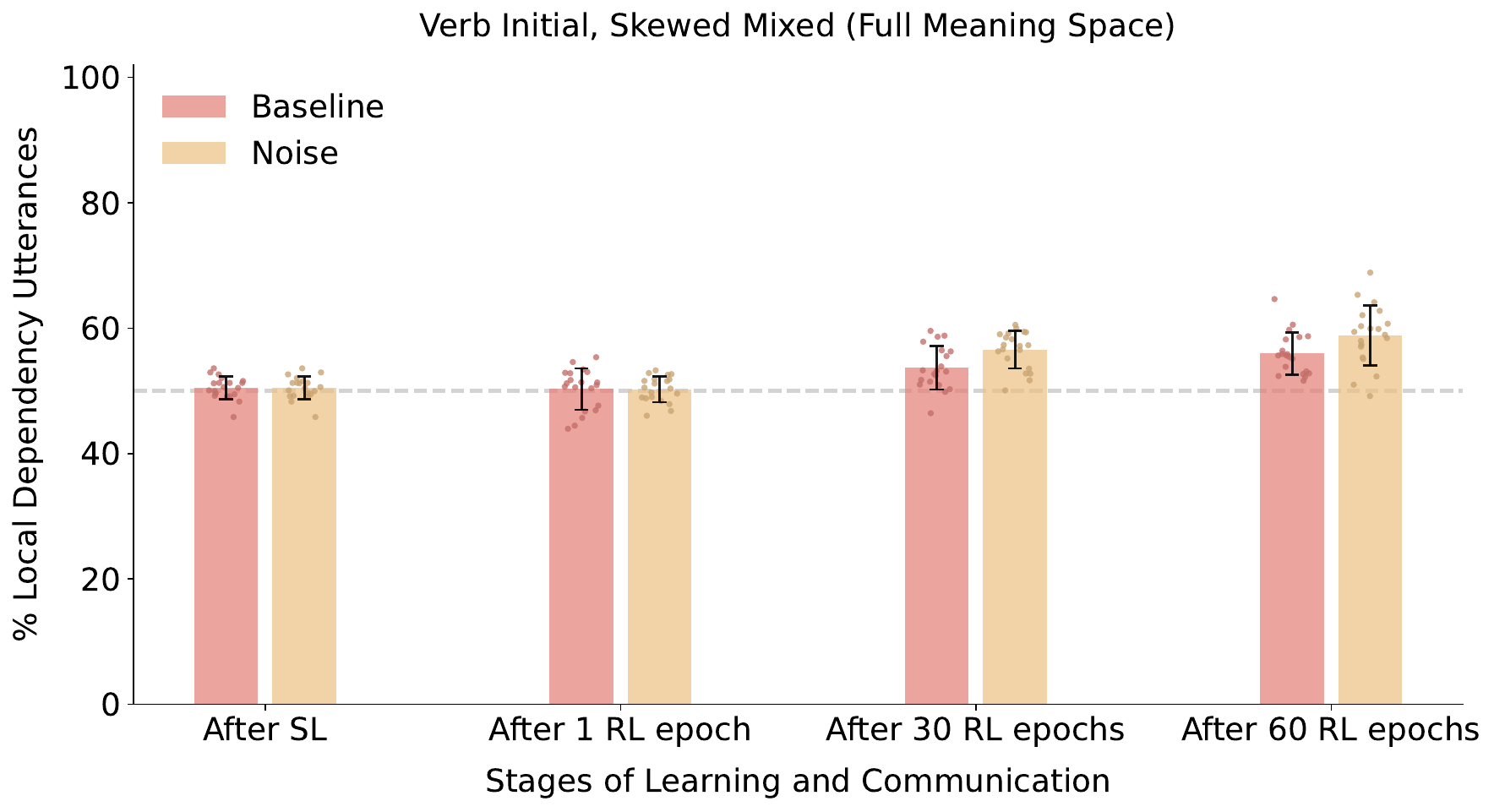}
        \caption{Local-dependency production}
        \label{fig:comm_initial_rnn_ALL_prod}
    \end{subfigure}

    \caption{
    Communication accuracy and production preferences of
    \textbf{RNN-based agents} using the \textbf{verb-initial} skewed mixed
    language with the \textbf{full meaning space} in the baseline and noise conditions.
    \textbf{(a)} Communication accuracy across communication epochs.
    \textbf{(b)} Proportion of local-dependency utterances at different stages of learning and communication. Each dot in panel (b) represents one random
    seed, and error bars indicate the standard error across seeds.
    Proportions in panel (b) are calculated over utterances classified as
    either local- or long-dependency utterances, excluding other productions
    and renormalizing the two categories to sum to 100\%.
    The dashed line indicates equal proportions of local- and long-dependency
    utterances (50\%).
    The x-axis labels in panel (b) denote training phases:
    \textit{After SL} represents the model's output at the end of the SL phase (60 epochs), while \textit{After 1 RL epoch}, \textit{After 30 RL epochs}, and \textit{After 60 RL epochs} refer to outputs after 1, 30, and 60 epochs of the communication phase, respectively.
    }
    \label{fig:comm_initial_rnn_ALL}
\end{figure}

As shown in Figure~\ref{fig:comm_initial_rnn_ALL_acc}, communication accuracy increased across RL epochs in both conditions, with higher accuracy in the \textit{Baseline} than in the \textit{Noise} condition. Given the generally high communication accuracy, we then examined whether agents developed production preferences during communication.
The production patterns observed when agents learned \textit{skewed} and \textit{uniform} verb-initial or verb-final languages were broadly similar across conditions, consistent with the results reported in \citet{yuqing-2024-dlm}. 
Moreover, production patterns under the noise condition followed the same overall trends as those under the no-noise condition, with no systematic deviations (\textit{b} = 1.02, 95\% CI [-1.01, 3.16]). Therefore, we focused the main analysis on the verb-initial \textit{skewed mixed} language under the no-noise condition, because it was representative of the broader pattern. Detailed production patterns under the noise condition were provided in Appendix~\ref{app:individual_prod}.

\subsubsection{Supervised Learning Phase}
\label{sec:full_sl}

\begin{figure*}[t]
\centering
\includegraphics[width=\linewidth]{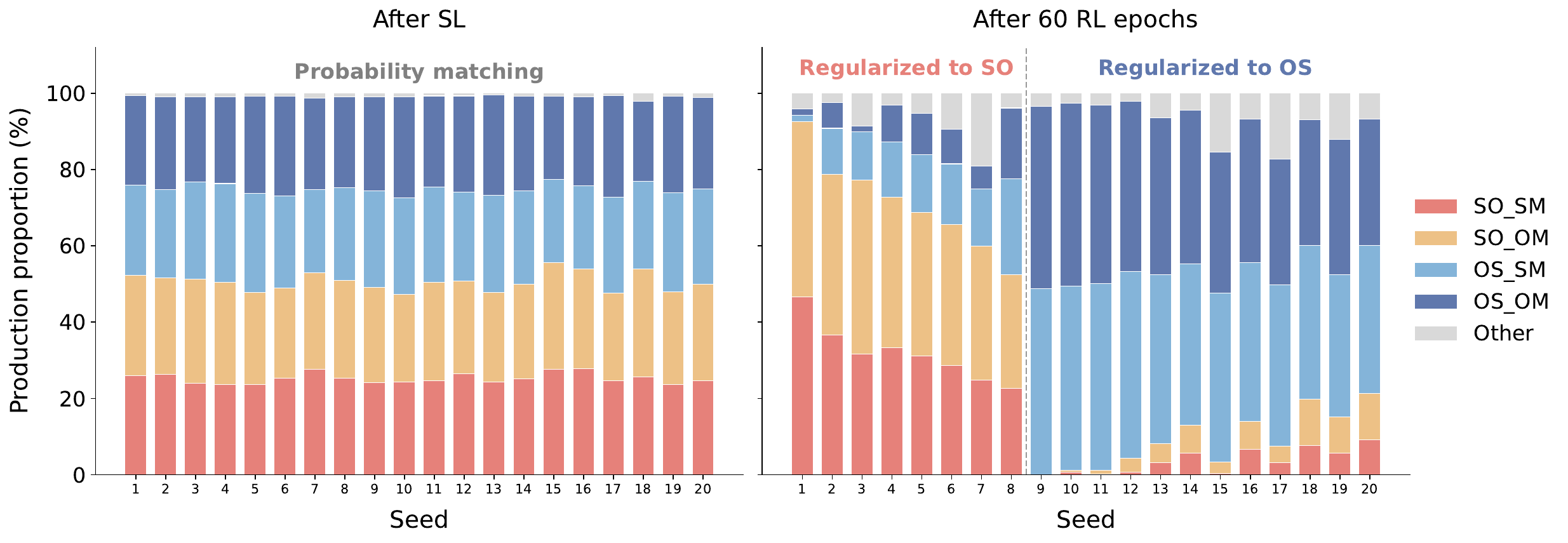}
\caption{
Seed-level production patterns before and after communication for the \textbf{verb-initial} \textit{skewed mixed} language with the \textbf{full meaning space}, \textbf{no dropout} (Baseline condition).
Each vertical bar represents one random seed and sums to 100\% of that seed's productions. 
Colors encode both word order and meaning type: SO\_SM, SO\_OM, OS\_SM, and OS\_OM denote productions using SO or OS order for subject-modified (SM) or object-modified (OM) meanings, respectively; grey denotes other or incorrect productions. 
The left panel shows productions after SL, where agents largely preserve the mixed input distribution, consistent with probability matching. 
The right panel shows productions after communication, where seeds are sorted by their dominant word order and individual agents regularize toward either SO or OS. 
This pattern shows that communication induces word-order regularization at the seed level, even though the aggregate proportion of local-dependency utterances can remain close to the 50\% input baseline.
}
\label{fig:spk_comm_regularization}
\end{figure*}

As visualized by the proximity to the dashed 0.5 baseline in Figure~\ref{fig:comm_initial_rnn_ALL_prod}, production preferences at the end of SL training showed that RNN-based learners preserved the balanced distribution found in the training data. 
% Bayesian multilevel modeling also confirms that 
The proportion of short-dependency utterances did not meaningfully deviate from chance level during this phase.
% the difference is very small effect, we run stat test, difference is less than 1%
Figure~\ref{fig:spk_comm_regularization} (left panel) showed the average production preferences of 20 speaking agents trained on the verb-initial skewed mixed language after SL, further confirming that agents maintained the original distribution. 
This result contrasted with the findings of \cite{chaabouni2019word} for sequence-to-sequence agents, possibly due to differences in artificial language design. It was also consistent with previous studies showing that neural learners exhibit strong probability-matching behavior after SL \citep{lian2021effect, lian2023communication}.

\subsubsection{Communication Phase}

Figure~\ref{fig:spk_comm_regularization} (right panel) shows the production preferences of 20 speaking agents after communicating with the verb-initial skewed mixed language under the no-noise condition.\footnote{Production patterns under other settings, including uniform languages, with dropout, with varying speaker capacity, showed similar regularization strategies. We were unable to successfully train communication using the Impatient Listener architecture in the full meaning space setting.} 
Agents showed different patterns from those observed after SL. Specifically, they developed two opposite regularization strategies: one group of agent pairs stopped producing VOS utterances and regularized to VSO order, while the other regularized to VOS order. 
In both cases, the distribution of short versus long dependencies remained balanced. As indicated by values close to the dashed 0.5 baseline in Figure~\ref{fig:comm_initial_rnn_ALL_prod} (After 60 RL epochs), agents’ productions remained close to the 50\% input level.
This pattern was observed in both the baseline and noise conditions.
These findings suggested that, during communication, an individual agent consistently regularized toward a single order, without showing a preference for minimizing dependency length.

\subsection{Discussion}

The neural agents in Experiment 1 were exposed to languages containing an equal proportion of long- and short-dependency utterances. 
Different word order alternatives were available for describing subject-modified and object-modified scenes. 
In this setting, minimizing dependency length required agents to produce different word orders for different types of scenes, making it impossible for the agents to regularize both word order \textit{and} dependency lengths simultaneously.
Under this setting, our speaking agents appeared to prefer the former over the latter, consistent with the findings of \cite{yuqing-2024-dlm}. Introducing both subject and object markers into the language or modifying model parameters did not affect the overall pattern. 

This outcome aligned with prior work in emergent communication showing that neural learners under communicative pressure tend to reduce word order entropy \citep{kharitonov2020entropy, lian2023communication}, which may explain the absence of DLM in the current setup. Interestingly, this preference contrasts with human behavior: in human experiments, when word order regularization and dependency length minimization were in competition, the latter could still emerge \citep{fedzechkina2018human}. One possible explanation was that human participants may not strongly regularize word order when case marking was available, whereas our neural agents appeared to do so.
The tendency toward a fixed word order revealed an important aspect of our experimental setup: even with all markers available, neural learners favored word order consistency as a strategy to reduce variability, which could override potential efficiency gains from DLM under communicative pressure. We further refined the experimental design and examined agents' preferences in experiments 2 and 3.

\section{Experiment 2: Short-Before-Long Overrides DLM}
\label{sec:exp2}

We aimed to stay as close as possible to the human experimental design in \citet{fedzechkina2018human}. However, as demonstrated in section \ref{sec:exp1}, order regularization dominated DLM when the agents interacted in the full meaning space setting.
To explore whether neural agents showed a preference for reducing DL when order regularization pressures were minimized, we modified the miniature language design and used only half of the original meaning space (i.e., only OM or SM meanings were present, but never both at the same time). In this setup, both types of regularization (order and dependency length) could happen concurrently. 
Preliminary results indicated no significant differences between neural agents trained with either OM or SM meaning type, as the meaning slots in our representations were order-invariant. Therefore, for simplicity, from this point onward, all subsequent analyses and plots presented in the text were based on the SM setup. 
% we will use the term "one-entity-modified meanings" to refer to either SM or OM meanings. Additionally, the term "modified constituents" will be used to describe the modified agent or the modified patient \footnote{All subsequent plots and analyses presented in the text are based on the SM setup.}

\subsection{Experimental setup}

The setup of this experiment matched the overview in \S~\ref{sec:miniaturelang} and \ref{con_distr}. In this experiment, we focused on two factors related to processing limitations, noise during comprehension on the listener's side and limited encoding capacity during production on the speaker's side, 
% processing capacity can include perception, memory, decision-making, comprehension, and production.
to investigate whether these factors encourage agents to favor DLM.  The experiment included four conditions: \textit{Baseline}, \textit{Noise}, \textit{Lower speaker capacity}, and a combined condition, \textit{Lower speaker capacity + Noise}. 

The parameter settings for Experiment 2 were summarized in Table~\ref{tab:exp_setup}.
% This paragraph is mostly just reading out the content of the table --- not needed.
% It could make sense to discuss *some* of the numbers in the table, but only as long as it serves to explain why you chose them or add some useful context about them.
% Under the \textit{Baseline} condition, the RNN speaker network has a 128-dimensional embedding layer followed by a 512-dim. hidden layer. The listener has a 256-dim. embedding followed by a 512-dim. hidden layer. The learning rate was set to 0.0001 for both the supervised learning and communication phases. 
To explore the architectural effect on production preferences, we also trained GRU-based agents for comparison. 
% The \textit{Baseline} settings include a hidden size of 64 for the speaker, a hidden size of 128 for the listener, a meaning embedding dimension of 8, a listener embedding size of 8, and a learning rate of 0.005 for the supervised learning phase and 0.002 for the communication phase. 
In the \textit{Lower speaker capacity} condition, the speaker hidden size was decreased from 512 to 256 for RNN agents and from 64 to 32 for GRU agents. 

\begin{table}[h]
    \centering
    \begin{tabular}{lccccccc}
        \toprule
        Condition & \multicolumn{2}{c}{Speaker} & \multicolumn{2}{c}{Listener} & \multicolumn{2}{c}{Learning Rate} & Dropout \\
        & Hidden Size & Embedding Size & Hidden Size & Embedding Size & SL & RL & \\
        \midrule
        \multicolumn{8}{l}{\textbf{RNN Agents}} \\
        Baseline & 512 & 128 & 512 & 256 & 0.0001 & 0.0001 & 0.0 \\
        Lower speaker capacity & 256 & 128 & 512 & 256 & 0.0001 & 0.0001 & 0.0 \\
        Noise & 512 & 128 & 512 & 256 & 0.0001 & 0.0001 & 0.1 \\
        Lower speaker capacity + Noise & 256 & 128 & 512 & 256 & 0.0001 & 0.0001 & 0.1 \\
        \midrule
        \multicolumn{8}{l}{\textbf{GRU Agents}} \\
        Baseline & 64 & 8 & 128 & 8 & 0.005 & 0.002 & 0.0 \\
        Lower speaker capacity & 32 & 8 & 128 & 8 & 0.005 & 0.002 & 0.0 \\
        Noise & 64 & 8 & 128 & 8 & 0.005 & 0.002 & 0.1 \\
        Lower speaker capacity + Noise & 32 & 8 & 128 & 8 & 0.005 & 0.002 & 0.1 \\
        \bottomrule
    \end{tabular}
    \caption{Experimental setup for different conditions. Dropout was set to 0.1 in the noise conditions.}
    \label{tab:exp_setup}
\end{table}

\subsection{Learning Accuracy}
\label{sec:half_initial_acc}

\begin{figure}[ht]
\centering
    \begin{subfigure}[b]{0.19\textwidth}
        \centering
        \includegraphics[width=\textwidth]{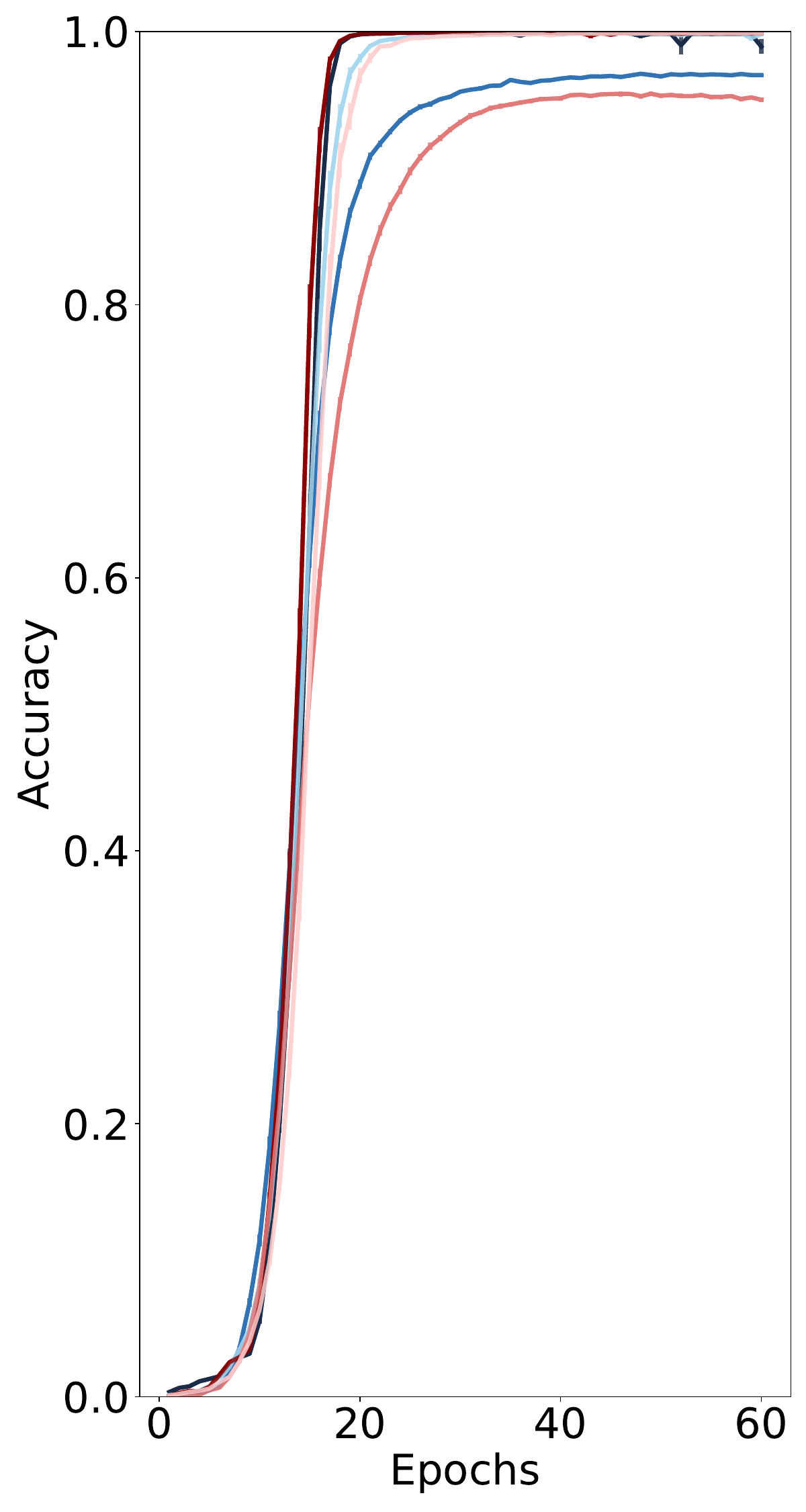}
        \caption{Speaking}
        \label{fig:initial_spk_acc_rnn}
    \end{subfigure}
    \hfill
    \begin{subfigure}[b]{0.19\textwidth}
        \centering
         \includegraphics[width=\textwidth]{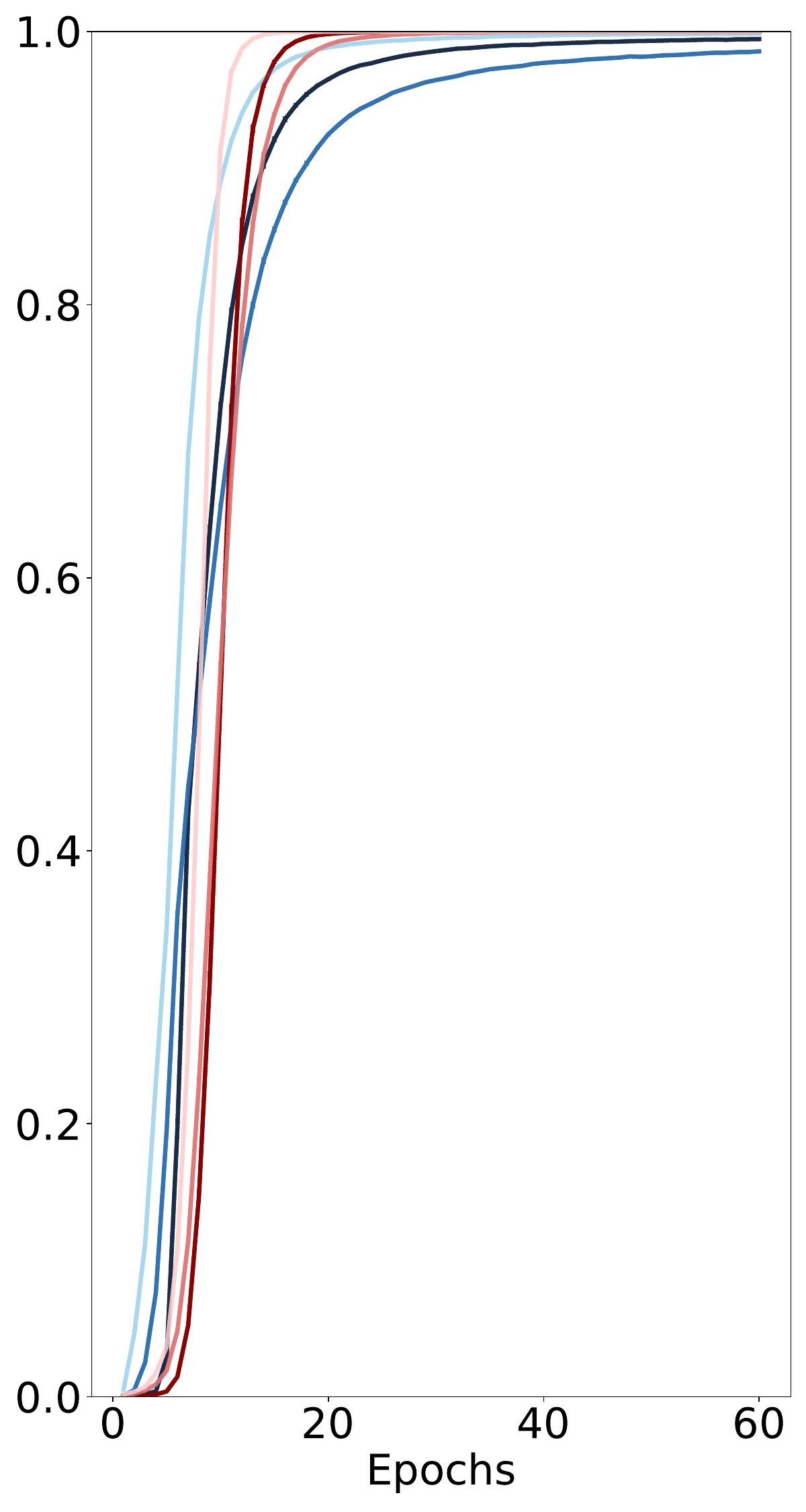}
        \caption{Listening (noise 0)}
        \label{fig:initial_lst_acc_rnn}
    \end{subfigure}
    \hfill
    \begin{subfigure}[b]{0.19\textwidth}
        \centering
        \includegraphics[width=\textwidth]{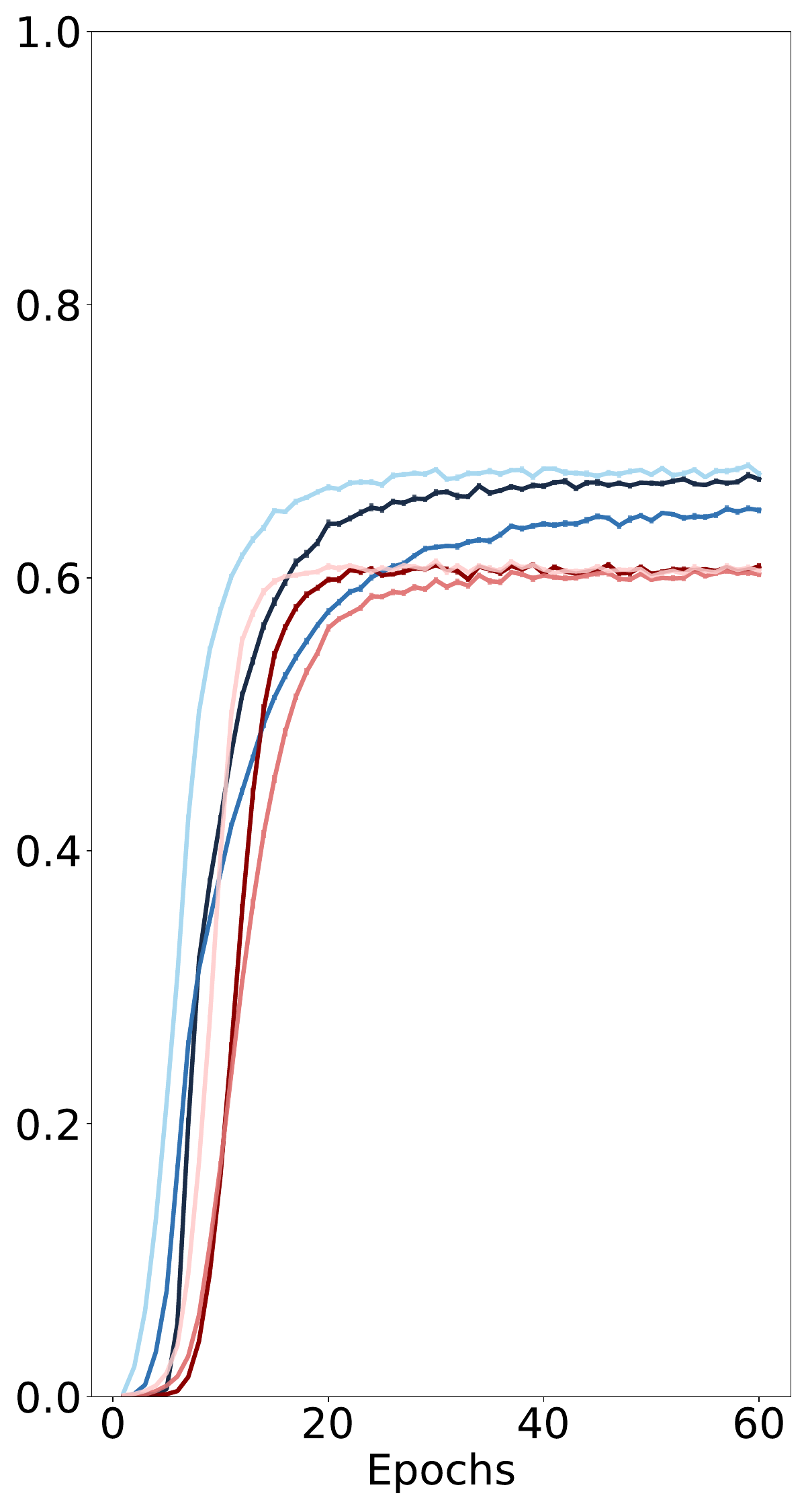}
        \caption{Listening (noise 0.1)}
        \label{fig:initial_lst_acc_rnn_0.1}
    \end{subfigure}
    \hfill
    \begin{subfigure}[b]{0.19\textwidth}
        \centering
        \includegraphics[width=\textwidth]{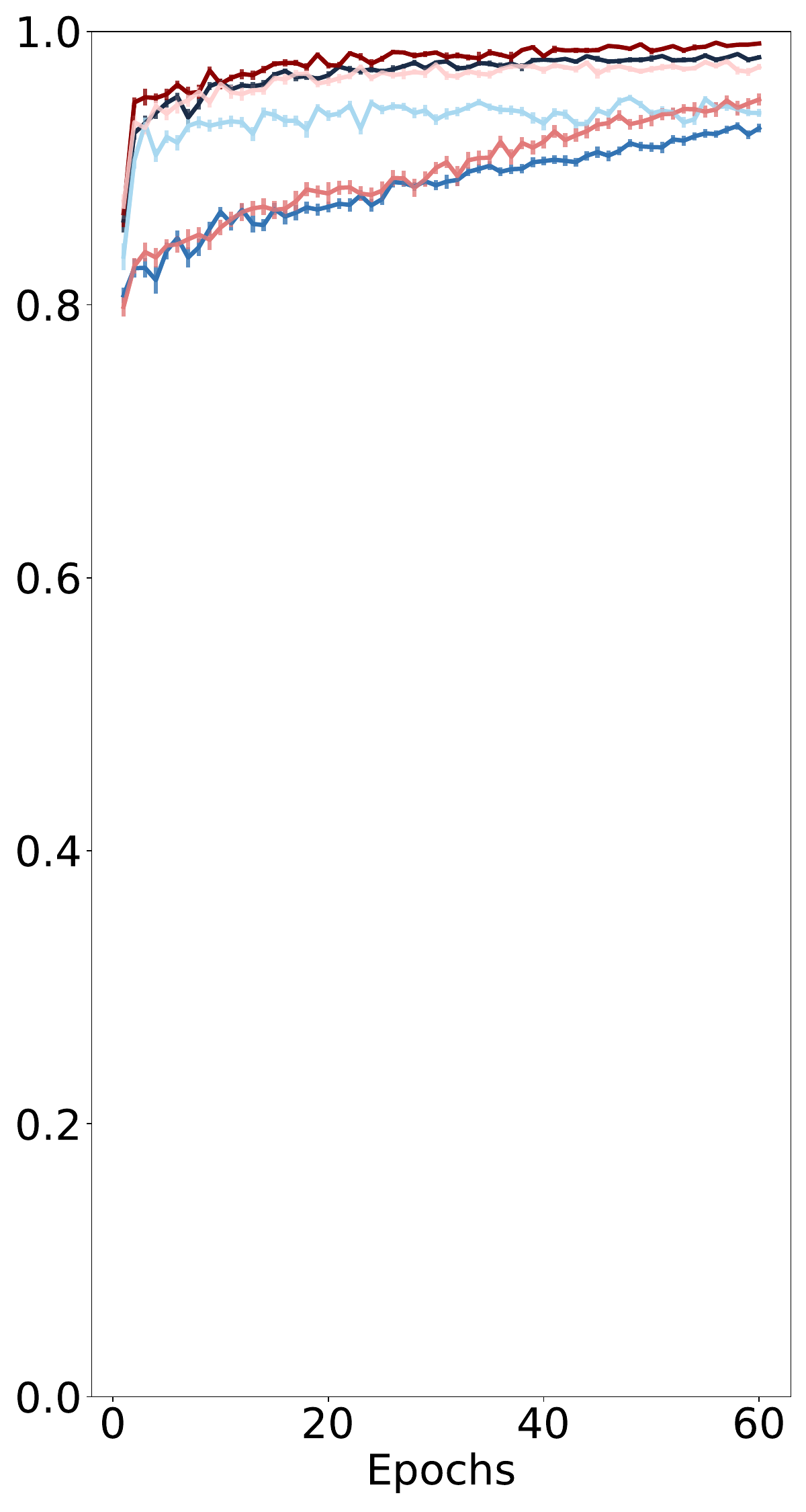}
        \caption{Communication (noise 0)}
        \label{fig:initial_comm_acc_rnn}
    \end{subfigure}
    \hfill
    \begin{subfigure}[b]{0.19\textwidth}
        \centering
        \includegraphics[width=\textwidth]{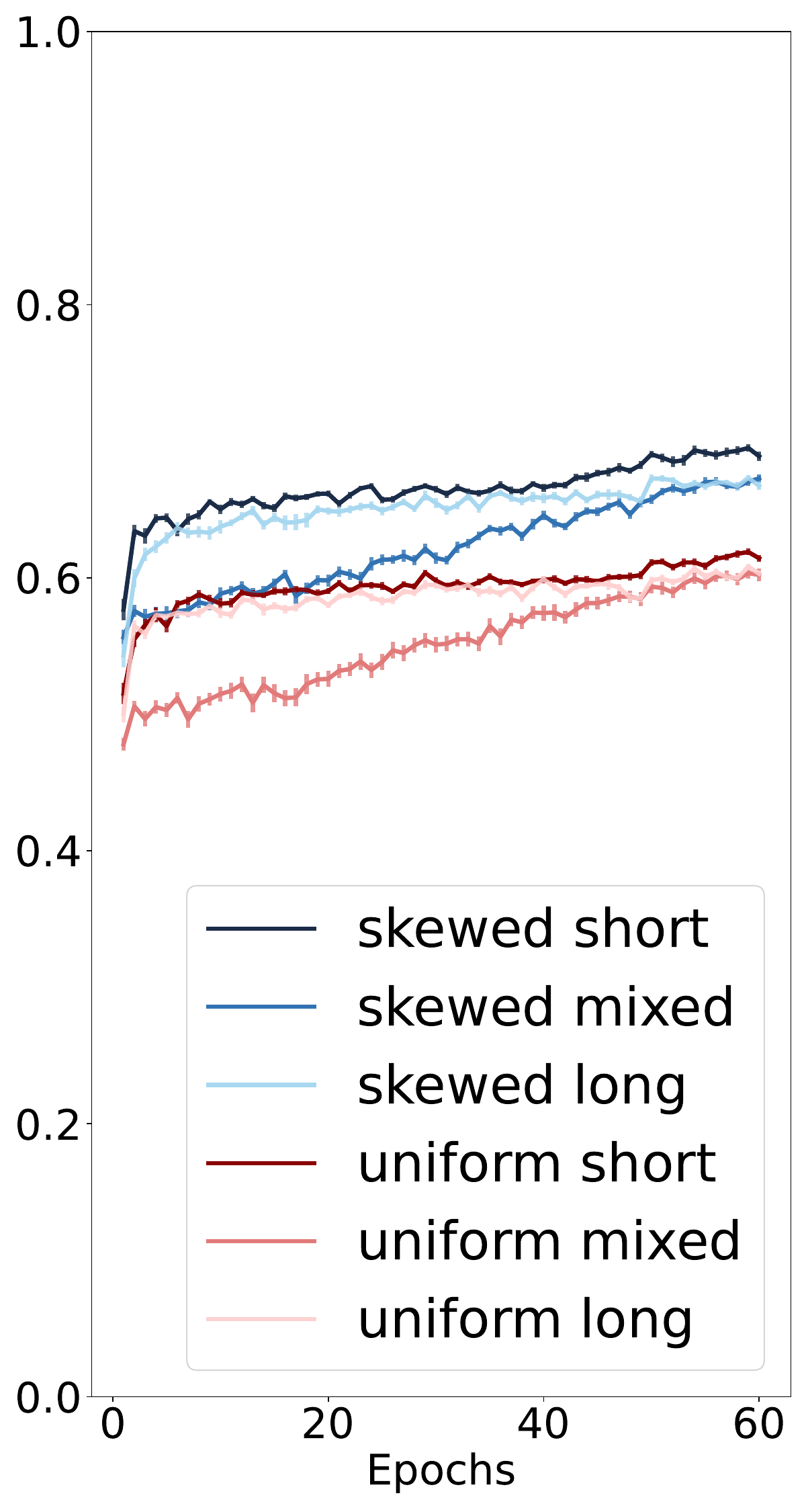}
        \caption{Communication (noise 0.1)}     \label{fig:initial_comm_acc_rnn_0.1}
    \end{subfigure}
    \caption{Accuracy of RNN-based agents in speaking, listening with noise levels 0 and 0.1, and communication with noise levels 0 and 0.1 during learning and communication in the \textbf{verb-initial language} under \textbf{the \textit{Baseline} and the Noise conditions}. Error bars represent the standard error for each line. Results for verb-final languages show similar patterns (see Appendix~\ref{fig:final_acc_rnn_combined}).}
    % \AB{Results or verb-final languages show similar patterns (see Appendix XXX)}
    \label{fig:initial_acc_rnn_combined}
\end{figure}

Figure~\ref{fig:initial_spk_acc_rnn}, \ref{fig:initial_lst_acc_rnn} and \ref{fig:initial_lst_acc_rnn_0.1} present the learning accuracies of speaking and listening agents trained on verb-initial languages with varying proportions of short- and long-dependency constructions for one-entity-modified meanings. 
Most findings of learning and communication accuracies 
discussed here were largely consistent across verb-final and verb-initial languages. Detailed results for verb-final languages were provided in Appendix~\ref{app:acc_final}.

% \subsubsection{Speaking Accuracy}
Figure~\ref{fig:initial_spk_acc_rnn} shows the speaking accuracy of verb-initial languages during the supervised learning phase. 
All language types reached over 95\% accuracy after around 20 epochs.
However, languages with mixed dependency lengths were more challenging for speaking agents to learn. 
Specifically, statistical analyses showed significantly lower speaking accuracies in verb-initial languages with mixed dependency utterances compared to those with short dependency utterances (\textit{b} = -.163, Bayesian 95\% Credible Interval [-.172, -.153]).
These differences possibly resulted from the higher variation and entropy of the mixed languages.
Compared to short-dependency languages, languages with long dependencies were slightly harder for speakers to learn (\textit{b} = -.038, 95\% CI [-.047, -.028]). 
Languages with short dependencies exhibited slightly faster learning speeds than those with long dependencies. 
Similarly, skewed mixed languages showed a modestly faster learning speed compared to those with a uniform distribution.

Figure~\ref{fig:initial_lst_acc_rnn} and \ref{fig:initial_lst_acc_rnn_0.1} present meaning-level listening accuracy across different noise conditions. 
% mix vs. consistent
% When languages are constructed with both long- and short-dependency utterances, listening accuracy is slightly higher but the difference is very small and not significant (\textit{b} = 0.001, 95\% CI [-.030, -.033]).  
% noise vs. no noise
Listening accuracy declined significantly under noisy conditions, with accuracy levels dropping considerably when noise was present (\textit{b} = -.329, 95\% CI [-.361, -.297]). Noise disrupted the input, preventing agents from accurately decoding the intended meanings, leading to lower overall accuracy compared to noise-free settings.

% skewed vs. uniform
Languages with skewed conditional word distributions were more robust to noise than those with uniform distributions, resulting in higher listening accuracy overall for skewed mixed languages under noisy conditions, as evidenced by the two-way interaction between noise and skewed language type (\textit{b} = .052, 95\% CI [.007, .099]), which confirmed our expectations. 
A closer examination of accuracy for specific meaning items in the verb-initial language revealed that the advantage of \textit{skewed} languages under noise was concentrated in the Action, Agent, and Patient categories, with no effect on the modifiers. This suggested that selectional preferences offered extra cues for recovering intended meanings from context.\footnote{For further details, Figure~\ref{fig:per-category-acc-epoch-initial} in Appendix~\ref{app:per_cat_acc_initial} shows listening accuracy for individual meaning items and average meaning reconstruction accuracy as a function of training epochs under noisy conditions.}

% short vs. long
% noise makes short dependency utterances advantageous for listeners of verb-initial languages --- not true now
% Finally, listening accuracy for languages with short dependencies is similar to that of languages with long dependencies in terms of the final listening accuracy, though learning speeds differ.  --- not true
Results showed a small but significant listening advantage of the long-dependency language type (\textit{b} = .034, 95\% CI [.001, .066]).
This suggested that shorter dependency lengths in our artificial languages did not provide learnability advantages for RNN-based neural agents during listening, even in noisy conditions (\textit{b} = -.005, 95\% CI [-.050, .039]). 
% TV: Doesn't this contradict with the previous sentence which suggests there is an advantage of learning speed? And also it contradicts our conclusion from the Coling paper, right? (I may be wrong)
% if that's the case (results different from the Coling ones) we should comment on that -- see below
Notably, this finding differed from the previous conclusion \citep{yuqing-2024-dlm}, where shorter dependency lengths were reported to provide overall learnability advantages. The discrepancy might arise from differences in experimental setup and the inclusion of new manipulations (i.e., both subject and object markers) in this study. 
This discrepancy motivated us to explore whether communication influenced agents’ preference for producing short-dependency utterances, even in the absence of a clear learning advantage during SL.

\subsection{Communication Success}

Figure~\ref{fig:initial_comm_acc_rnn} and \ref{fig:initial_comm_acc_rnn_0.1} show communication accuracy during the communication learning phase across different noise conditions. Across all language types, communication accuracy gradually improved and reached above 95\% at convergence, indicating that agents were able to communicate effectively using their shared language.
% mix vs. consistent 
Languages with mixed dependency lengths showed slightly lower communication accuracy compared to those with consistent dependency lengths (\textit{b} = -.081, 95\% CI [-.086, -.076]). This difference likely arose from the increased variability and entropy in mixed languages, which made it harder for agents to consistently interpret messages.

% noise vs. no noise 
Noise had a detrimental effect on agents' communication success, as shown by the lower accuracies in Figure~\ref{fig:initial_comm_acc_rnn_0.1} (\textit{b} = -.335, 95\% CI [-.340, -.330]).
One possible explanation was that, given the decreased accuracy in listening, the listener functioned less effectively as a scaffolding agent. Consequently, the speaker faced challenges in adapting its production, and the listener failed to decode the intended meanings, leading to a decline in communication accuracy.

% skewed vs. uniform 
Under noisy conditions, languages with skewed word distributions demonstrated greater robustness than those with uniform distributions, as evidenced by the two-way interaction between noise and skewed language type (\textit{b} = .062, 95\% CI [.055, .069]), maintaining higher communication accuracy overall. This suggested that skewed mixed languages provided stronger contextual cues, which helped agents partly mitigate the negative effects of noise.

% long vs. short 
Results indicated a slight communication advantage for short dependency languages over long dependency languages (\textit{b} = -.018, 95\% CI [-.023, -.012]). This advantage became more evident under the skewed condition (\textit{b} = -.011, 95\% CI [-.018, -.003]). 
% The communication accuracy curves for languages with short and long dependencies overlap,--- not true in the model; small but significant, indicating that communication accuracy is similar for both types. 
This suggested that dependency length influenced communication success, especially when the languages had skewed word frequency distributions. A complementary analysis of production errors across meaning-item categories, dependency types, and experimental conditions is provided in Appendix~\ref{app:error_analyses}.
%skewed conditional word distributions make the preference stronger;

\subsection{Production Preferences for Verb-initial Languages}

\begin{figure}[ht]
    \centering

    % =====================================================
    % Top row: RNN
    % =====================================================
    \begin{subfigure}[b]{0.48\textwidth}
        \centering
        \includegraphics[width=\textwidth]{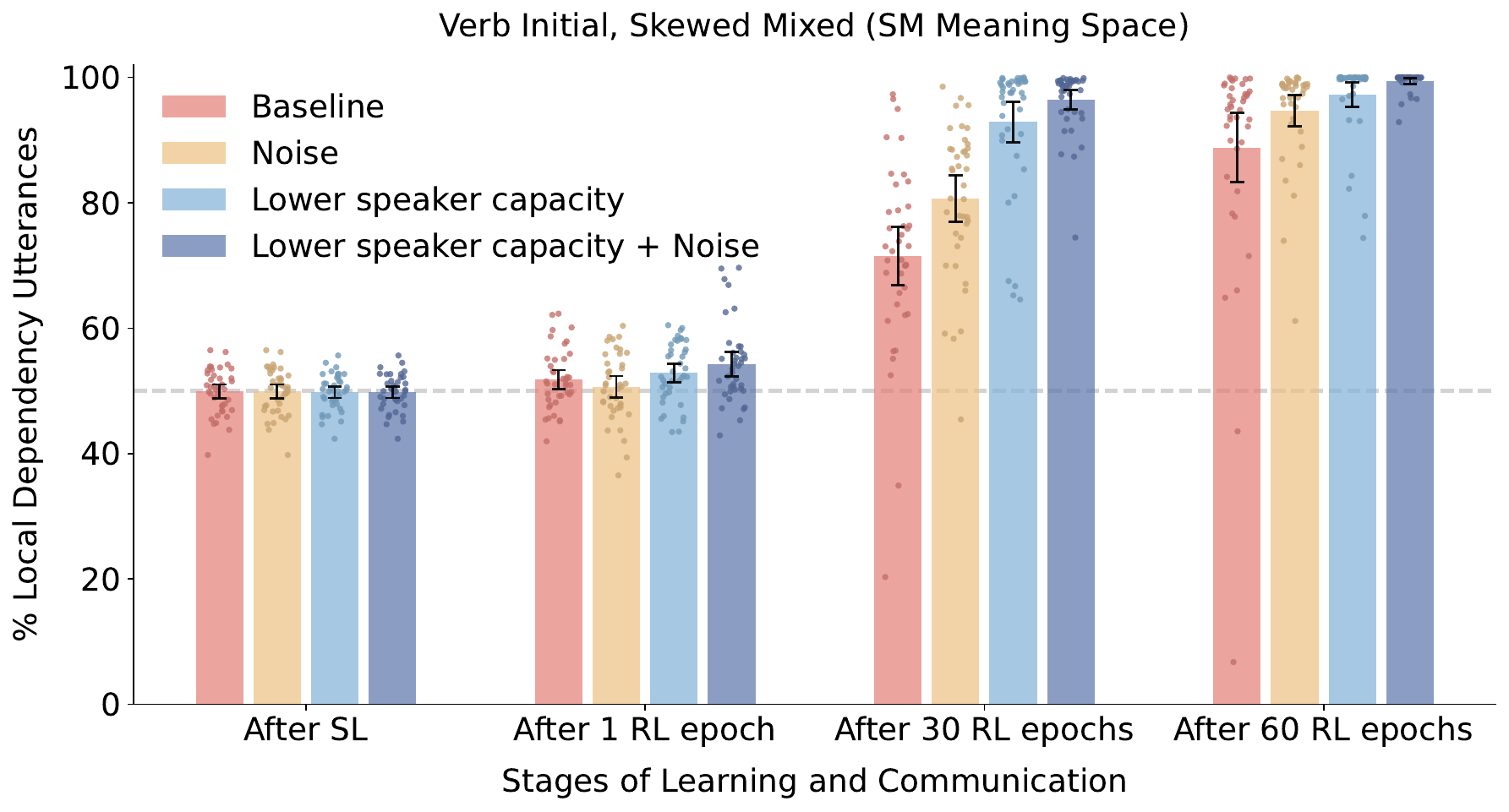}
        \caption{Verb-initial, RNN-based agents}
        \label{fig:comm_initial_rnn_skewed}
    \end{subfigure}
    \hfill
    \begin{subfigure}[b]{0.48\textwidth}
        \centering
        \includegraphics[width=\textwidth]{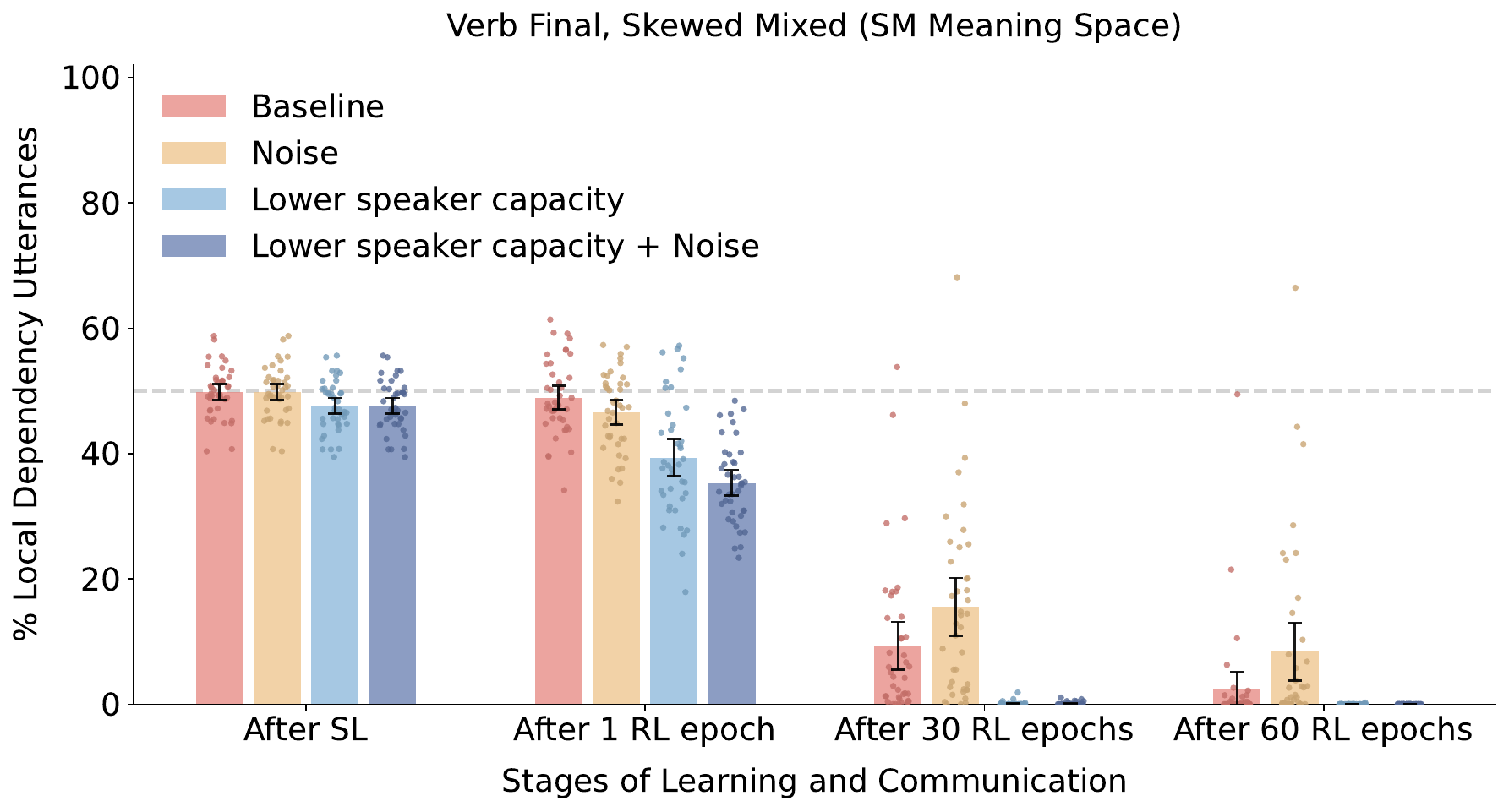}
        \caption{Verb-final, RNN-based agents}
        \label{fig:comm_final_rnn_skewed}
    \end{subfigure}

    \vspace{0.5em}

    % =====================================================
    % Bottom row: GRU
    % =====================================================
    \begin{subfigure}[b]{0.48\textwidth}
        \centering
        \includegraphics[width=\textwidth]{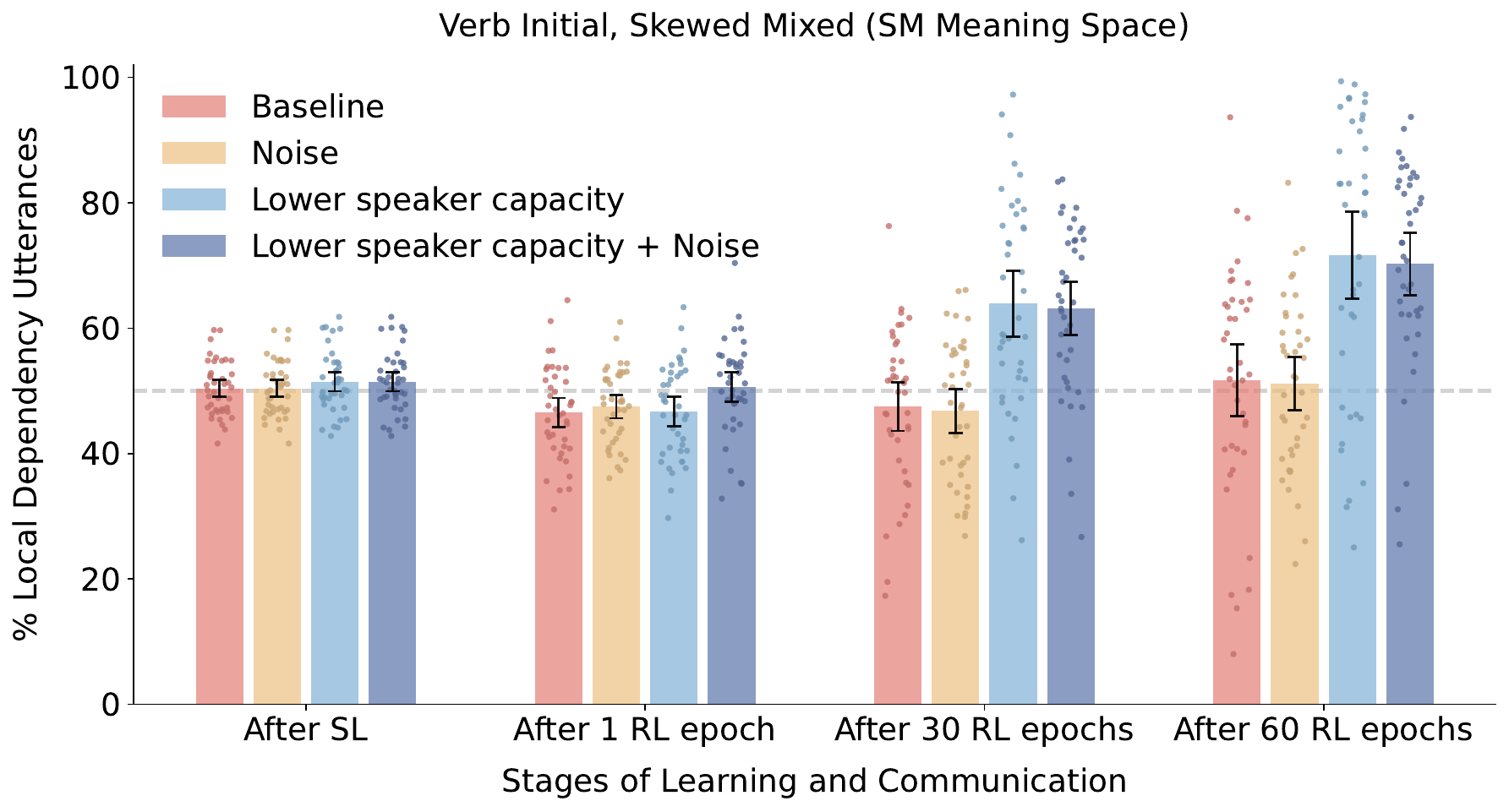}
        \caption{Verb-initial, GRU-based agents}
        \label{fig:comm_initial_gru_skewed}
    \end{subfigure}
    \hfill
    \begin{subfigure}[b]{0.48\textwidth}
        \centering
        \includegraphics[width=\textwidth]{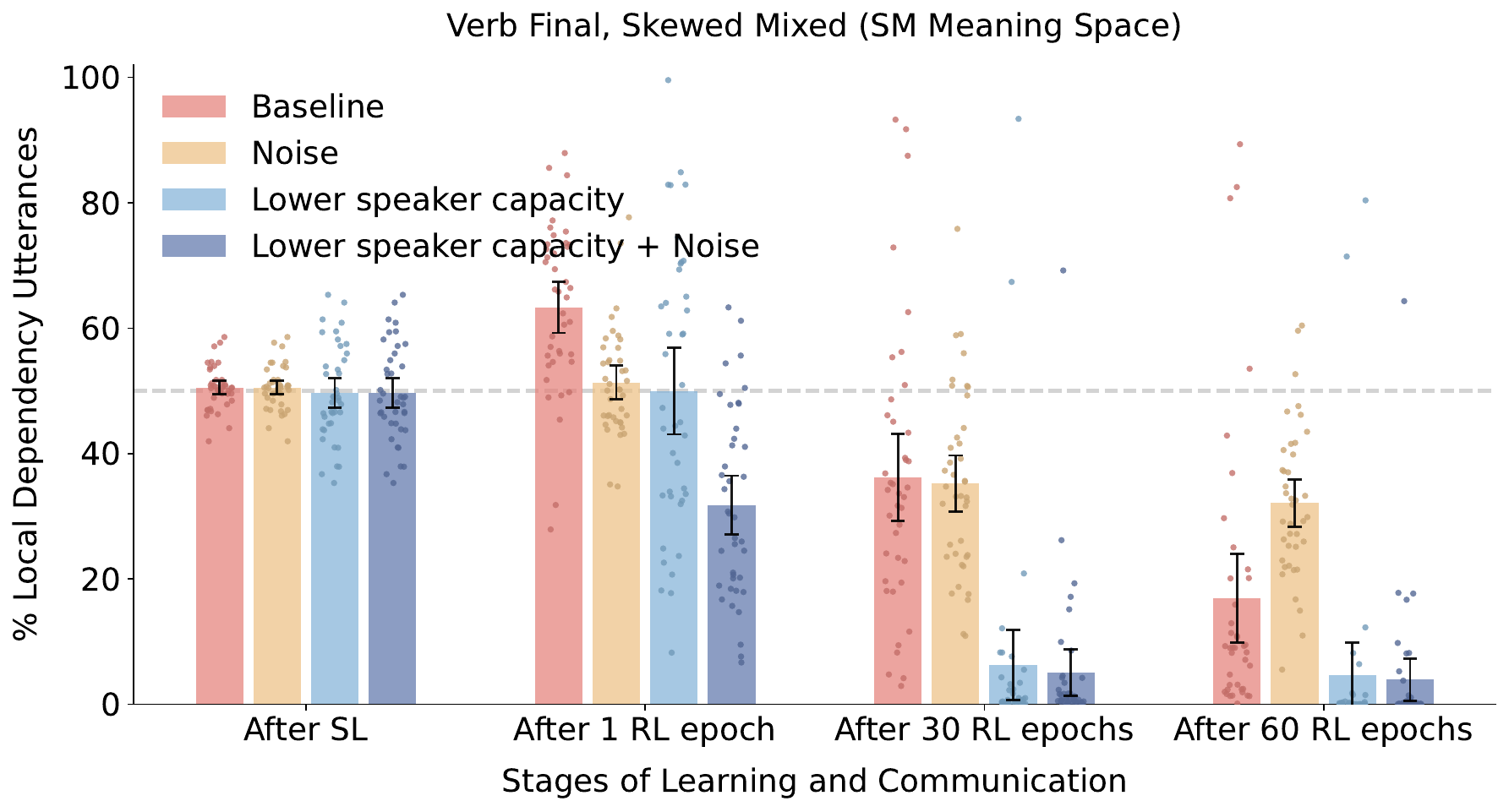}
        \caption{Verb-final, GRU-based agents}
        \label{fig:comm_final_gru_skewed}
    \end{subfigure}

    \caption{
    Proportion of local-dependency utterances produced by \textbf{RNN-based} and \textbf{GRU-based agents} at different stages of learning and communication using the \textbf{skewed mixed language} with the \textbf{half meaning space}. Panels (a) and (b) show the results for RNN-based agents in the \textbf{verb-initial} and \textbf{verb-final} languages, respectively, while panels (c) and (d) show the corresponding results for GRU-based agents. The four experimental conditions are: \textit{Baseline}, \textit{Noise}, \textit{Lower speaker capacity}, and \textit{Lower speaker capacity + Noise}. Local dependencies increase in the verb-initial language but decrease in the verb-final language. Because short-before-long ordering corresponds to local dependencies in the verb-initial language but to long dependencies in the verb-final language, both patterns reflect a common short-before-long ordering preference.
    }
    \label{fig:comm_skewed}
\end{figure}

As demonstrated in \citet{yuqing-2024-dlm} and \S~\ref{sec:half_initial_acc}, agents leveraged dependency relationships through the statistical patterns in skewed mixed languages, as evidenced by higher listening accuracy and communication success under noisy conditions. 
This suggested that artificial language designs with realistic skewed patterns yielded more reliable results when examining linguistic phenomena related to dependency relationships. Therefore, we focused on testing and reporting results from languages featuring skewed distributions.
We explored whether the slight speaking advantage observed for short-dependency languages in verb-initial skewed mixed languages during SL could lead to a progressively stronger DLM across RL phases (Figure~\ref{fig:comm_initial_rnn_skewed}).
Statistical comparison results are reported in Table~\ref{tab:verb_initial}.

\textbf{Production after SL} As shown in Figure~\ref{fig:comm_initial_rnn_skewed} (After SL) and supported by Bayesian hypothesis testing, for verb-initial languages the production preferences of speaking agents at the end of SL did not meaningfully differ from the 50\% baseline. Under all conditions, RNN-based learners tended to preserve the distribution of long and short dependencies present in the training data, and therefore did not exhibit DLM in a purely SL setting.
% % % at the spk60 phase, before communication begins, the mean proportion of local dependency length (DL) utterances produced by speakers was close to 50\% (Baseline: b = 48.91, 95\% CI [48.08, 49.73]; Noise 0.1; Lower speaker capacity; Lower speaker capacity \& noise 0.1).
The result was consistent with the full meaning space setup in \S~\ref{sec:full_sl}.

% \begin{figure}[ht]
% \centering
%     \includegraphics[width=0.7\textwidth]{figs/comm_initial_rnn_skewed.pdf}
%     \caption{
% Proportion of short-dependency utterances produced by \textbf{RNN-based agents} during learning and communication in the \textbf{verb-initial skewed mixed language}. 
% Dashed line indicates the baseline proportion of short dependencies in the input data.
% Labels such as \textbf{Spk60} denote the model's output after 60 epochs of SL, while \textbf{Comm1}, \textbf{Comm30}, and \textbf{Comm60} refer to outputs after 1, 30, and 60 epochs of communication-based training, respectively.
% }
% \label{fig:comm_initial_rnn_skewed}
% \end{figure}

\begin{table}[ht]
\centering
\resizebox{\textwidth}{!}{
\begin{tabular}{llccc}
\toprule
\textbf{Language} & \textbf{Comparison} & \textbf{After 1 RL epoch} & \textbf{After 30 RL epochs} & \textbf{After 60 RL epochs} \\
\midrule
\multirow{6}{*}{V-initial}
& Noise vs. Baseline          & \textit{b} = -1.13, 95\% CI [-3.19, 0.91] & \textbf{\textit{b} = 9.22, 95\% CI [4.47, 14.03]} & \textbf{\textit{b} = 5.99, 95\% CI [1.41, 10.56]} \\
& Limited vs. Baseline        & \textit{b} = 1.05, 95\% CI [-1.00, 3.13] & \textbf{\textit{b} = 21.42, 95\% CI [16.66, 26.21]} & \textbf{\textit{b} = 8.56, 95\% CI [4.16, 12.93]} \\
& Noise+Limited vs. Baseline  & \textbf{\textit{b} = 2.48, 95\% CI [0.44, 4.51]} & \textbf{\textit{b} = 24.95, 95\% CI [20.20, 29.67]} & \textbf{\textit{b} = 10.66, 95\% CI [6.22, 15.22]} \\
& Limited vs. Noise        & \textbf{\textit{b} = 2.18, 95\% CI [0.17, 4.22]} & \textbf{\textit{b} = 12.28, 95\% CI [7.58, 17.05]} & \textit{b} = 2.56, 95\% CI [-1.96, 7.12] \\
& Noise+Limited vs. Noise  & \textbf{\textit{b} = 3.62, 95\% CI [1.50, 5.73]} & \textbf{\textit{b} = 15.79, 95\% CI [11.02, 20.60]} & \textbf{\textit{b} = 4.67, 95\% CI [0.28, 9.06]} \\
& Noise+Limited vs. Limited   & \textit{b} = 1.46, 95\% CI [-0.59, 3.54] & \textit{b} = 3.56, 95\% CI [-1.03, 8.37] & \textit{b} = 2.09, 95\% CI [-2.42, 6.43] \\

\bottomrule
\end{tabular}}
\caption{Pairwise statistical comparisons for verb-initial languages across communication phases. Significant results (95\% CI not containing 0) are shown in bold.}
\label{tab:verb_initial}
\end{table}

\textbf{Production under \textit{Baseline} Communication Condition}
As shown in Figure~\ref{fig:comm_initial_rnn_skewed} (Baseline), during the first round of communication (After 1 RL epoch), the mean proportion of local DL utterances produced by speakers increased slightly to b = 51.78 (95\% CI [50.35, 53.21]), and by the final communication phase (After 60 RL epochs), it reached b = 89.07 (95\% CI [86.38, 91.77]), indicating that speakers gradually produced a higher proportion of short dependency utterances. 
Analyzing individual seed behaviors, we found that 38 out of 40 random seeds regularized to shorter DLs in the baseline condition. This pattern is visible from the individual seed points in Figure~\ref{fig:comm_initial_rnn_skewed}. 
% The full seed-level production patterns for all conditions are provided in Appendix~\ref{app:individual_prod_exp2}.
% explain individual patterns: ref: One type of rule participants could have imposed would be to produce determiners categorically, either all or none of the time. (Regularizing Unpredictable Variation: The Roles of Adult and Child Learners in Language Formation and Change)

\textbf{Production under \textit{Noisy} Communication Condition}
To simulate realistic communication, we trained neural agents to communicate under noisy conditions.
Figure~\ref{fig:comm_initial_rnn_skewed} (Noise) shows the production patterns of speaker networks after communication under the noisy condition for the verb-initial language with SM meanings. 
When noise was present, utterances with shorter DL quickly became the dominant pattern, and utterances with longer DL were reduced more strongly than in the no-noise condition.
Specifically, all 40 random seeds produced a higher percentage of shorter-dependency utterances, indicating that noise yielded a slightly higher preference towards reducing DL. 

\textbf{Production When Speakers Have \textit{Lower Capacity}}
% architectural features (depth, width, and number of parameters)
To examine how limited processing capacity affected production, we manipulated the architectural features of the speaker networks.
Specifically, we reduced the hidden layer size to simulate agents with constrained memory and cognitive resources.
Figure~\ref{fig:comm_initial_rnn_skewed} (Lower speaker capacity) presents the resulting production patterns: the speaker networks trained with limited capacity showed a noticeable preference for DLM in their productions.
Moreover, all random seeds converged on a preference for shorter DL. Compared to the \textit{Baseline} communication condition, this bias toward DLM was further strengthened under lower capacity.
This aligned with observations in human learners, who tended to favor simpler, shorter structures when cognitive load is high. In contrast, higher-capacity networks exhibited a more varied range of production patterns, suggesting that increased processing capacity allowed for a greater tolerance of longer dependencies in utterances. 

\textbf{Production when Communication is \textit{Noisy} and Speaker Capacity is \textit{Limited}}
We additionally tested a combined condition with both noise and reduced capacity. Results revealed a cumulative effect: the joint presence of noise and low capacity produced the strongest bias toward shorter dependencies across all seeds.

\textbf{Experiments with Different Neural Architectures} 
% To investigate potential architectural effects on speakers' production, we experiment with different neural-network architectures: vanilla RNN and GRU.
As shown in Figure~\ref{fig:comm_initial_gru_skewed}, the overall trends observed for GRU agents were similar to those seen in RNN agents, albeit with some notable differences. 
GRU-based agents exhibited similar preferences after repeated communication with the verb-initial skewed mixed language under the \textit{Baseline} communication condition, with no notable effects of noise. Across multiple rounds of communication, speakers, on average, produced an equal proportion of short- and long-dependency utterances in both conditions. 
% however, for the verb-final languages (not shown in the paper, but I looked back at graphs in Discord) there does seem to be an effect of noise on increasing long-dependency utterances? --- true, still don't know why
This indicated that GRU-based agents were generally more robust to external factors introduced during learning and communication, such as noise, showing fewer differences in production behavior between the \textit{Baseline} and noise conditions as compared to RNN-based agents.
This apparent robustness of GRUs might be attributed to their gating mechanisms, specifically the update and reset gates, that control the retention and refresh of hidden states. Through the update gate, GRUs can preserve input distribution more reliably, making them less susceptible to perturbations during communication \citep{cho-etal-2014-learning, can2020gating}. In contrast, vanilla RNNs are prone to the vanishing gradient problem and thus tend to be more sensitive to such manipulations \citep{pascanu2013difficulty}.

\subsection{Production Preferences for Verb-final Languages}
Figure~\ref{fig:comm_final_rnn_skewed} and Figure~\ref{fig:comm_final_gru_skewed} show the production patterns for the verb-final language, and Table~\ref{tab:verb_final} presents the statistical testing results. Different from what we observed for the verb-initial language, after several communication rounds, speakers gradually showed a preference for producing more long-dependency utterances. 
% Figure~\ref{fig:final_baseline_exp2} illustrates production at the final communication phase under the baseline condition.
This pattern was observed for both RNN- and GRU-based agents. Across both architectures, agents consistently produced the OSV word order.
These results suggested that DLM alone could not provide a unified account of the patterns observed in both verb-initial and verb-final languages. Importantly, in the verb-final language, long-dependency utterances corresponded to short-before-long orderings, whereas in the verb-initial language, short-before-long orderings corresponded to local-dependency utterances (see Figure~\ref{fig:lang}). When interpreting the results under this mapping, we found that agents consistently exhibited a \textbf{short-before-long preference} across both language types. Therefore, the observed DLM asymmetry between verb-initial and verb-final languages was likely driven by a general short-before-long bias \citep{behaghel1909beziehungen}.

\begin{table}[ht]
\centering
\resizebox{\textwidth}{!}{
\begin{tabular}{llccc}
\toprule
\textbf{Language} & \textbf{Comparison} & \textbf{After 1 RL epoch} & \textbf{After 30 RL epochs} & \textbf{After 60 RL epochs} \\
\midrule
\multirow{6}{*}{V-final}
& Noise vs. Baseline          & \textit{b} = -2.31, 95\% CI [-5.24, 0.68] & \textbf{\textit{b} = 6.17, 95\% CI [2.20, 10.16]} & \textbf{\textit{b} = 5.88, 95\% CI [2.32, 9.41]} \\
& Limited vs. Baseline        & \textbf{\textit{b} = -9.58, 95\% CI [-12.58, -6.67]} & \textbf{\textit{b} = -9.23, 95\% CI [-13.21, -5.27]} & \textit{b} = -2.47, 95\% CI [-6.04, 1.10] \\
& Noise+Limited vs. Baseline  & \textbf{\textit{b} = -13.60, 95\% CI [-16.48, -10.71]} & \textbf{\textit{b} = -9.22, 95\% CI [-13.19, -5.25]} & \textit{b} = -2.49, 95\% CI [-5.96, 1.03] \\

& Limited vs. Noise          & \textbf{\textit{b} = -7.26, 95\% CI [-10.26, -4.33]} & \textbf{\textit{b} = -15.39, 95\% CI [-19.27, -11.59]} & \textbf{\textit{b} = -8.35, 95\% CI [-11.87, -4.79]} \\
& Noise+Limited vs. Noise    & \textbf{\textit{b} = -11.32, 95\% CI [-14.31, -8.33]} & \textbf{\textit{b} = -15.39, 95\% CI [-19.34, -11.43]} & \textbf{\textit{b} = -8.37, 95\% CI [-11.89, -4.82]} \\
& Noise+Limited vs. Limited  & \textbf{\textit{b} = -4.02, 95\% CI [-7.00, -1.05]} & \textit{b} = -0.02, 95\% CI [-4.10, 3.92] & \textit{b} = -0.02, 95\% CI [-3.52, 3.49] \\
\bottomrule
\end{tabular}}
\caption{Pairwise statistical comparisons for verb-final languages across communication phases for RNN-based agents. Significant results (95\% CI not containing 0) are shown in bold.}
\label{tab:verb_final}
\end{table}

\subsection{Discussion}

Unlike the first experiment, where a strong tendency to regularize word order prevented the appearance of other, more subtle, ordering preferences, agents in this experiment showed a strong short-before-long ordering bias across both verb-initial and verb-final languages and for both RNNs and GRUs.
While these results showed that neural agents could demonstrate behaviors that were consistent with human-like DLM preferences in verb-initial languages, this behavior might be epiphenomenal, arising from more general short-before-long ordering biases rather than from an explicit pressure to minimize dependency length.

Differences across architectures were also notable: RNN agents gradually developed stronger preferences with added noise, while GRU agents showed greater stability and resistance to noise.  
This highlights the role of architectural design in shaping linguistic behavior. It is worth noting that the small size of the artificial languages in this study may render gated architectures like GRUs disproportionately powerful, which could account for their resistance to experimental manipulations.
% Mechanistically, the short-before-long bias may reflect a production-side sequencing pressure in recurrent speaker agents \citep{cho-etal-2014-learning}. Generating the longer constituent first may make early production more fragile, as early errors can propagate to later positions and the model must maintain the other unresolved constituent while completing the utterance. Producing the shorter constituent first may therefore make sequence generation more stable under the pressure to maximize communication success.} 
One possible explanation for the short-before-long bias is that it reflects a production-side sequencing pressure in recurrent speaker agents. Because RNN-based speakers generate utterances incrementally, different constituent orders correspond to different generation trajectories. Starting with the shorter constituent may provide a simpler and more stable early generation sequence, which could favor short-before-long ordering during communication. However, this interpretation remains tentative, as we did not directly examine the speakers' hidden-state dynamics.
In these standard communication settings, speaker-side production pressures appear to play a stronger role in shaping word order than DLM. This suggests that DLM does not emerge as a default production preference, and that DLM may instead require specific communicative pressures driven by the listener's comprehension constraints, such as incremental processing. Given that key differences between the artificial learners in our study and the human participants in \citet{fedzechkina2018human} persisted in verb-final languages, in the next experiment, we explored the impact of communicating with an Impatient Listener \citep{rita-etal-2020-lazimpa}, an architecture resembling the human online prediction mechanism, on speakers' production preferences during communication.
% This suggests that pressures, such as noise on the listener side or processing capacity on the speaker side, promote efficient communication strategies similar to those of humans. 
% In contrast, verb-final languages exhibited no preference for DLM, which is consistent with prior studies showing that head-final languages display weaker DLM effects than head-initial ones \citep{futrell2015large}, and with psycholinguistic evidence of “anti-DLM” effects in head-final structures \citep{jing2022dependency}.

% "RNNs potentially representing a more cognitively plausible model" ---\AB{here I'd be careful, maybe we can  connect this specifically to the fact that we're working with very small languages for which gating RNN may be "too powerful"} --- done

\section{Experiment 3: Incremental Processing Elicits DLM} 
\label{sec:exp3}

Building on Experiment 2, Experiment 3 further investigated factors that promoted a preference for DLM.
Because Experiment 2 showed a short-before-long preference, consistent with DLM only in the verb-initial language, it remained important to test whether specific communicative pressures could nevertheless strengthen DLM tendencies in both languages.
To this end, we tested speakers' production preferences when they interacted with an Impatient Listener. Specifically, we modified the reward function by incorporating cumulative rewards, assigning higher rewards to speaker productions that yielded greater incremental accuracy (see \S~\ref{sec:impa_arch})\footnote{As noted in \S \ref{sec:adaptation_nellcom}, Experiment 3 used only vanilla RNN agents, so it tested the effect of incremental processing within the RNN architecture rather than providing a full cross-architecture comparison.}.

\subsection{Experimental Setup} 
Preliminary experiments with Impatient Listeners and the same parameters as in Experiment 2 resulted in low communication accuracy. To address this issue, we made minor parameter adjustments for Experiment 3.
Specifically, since impatient listeners were more difficult to train, we increased the model size and used a smaller learning rate during RL to improve overall performance. 
For comparison, agents were also trained using the standard listener architecture with the same parameters. 
There were three conditions in total: \textit{Baseline} (similar to the \textit{Baseline} in Experiment 2 but with different initialization parameters), \textit{Impatient}, and \textit{Impatient + Noise}. \footnote{The \textit{Lower speaker capacity} condition is excluded because the current speaker model represents the minimal size capable of achieving reasonable communication accuracy.}

\begin{table}[h]
    \centering
    \begin{tabular}{lccccccc}
        \toprule
        Condition & \multicolumn{2}{c}{Speaker} & \multicolumn{2}{c}{Listener} & \multicolumn{2}{c}{Learning Rate} & Dropout \\
        & Hidden Size & Embedding Size & Hidden Size & Embedding Size & SL & RL & \\
        \midrule
        \multicolumn{8}{l}{\textbf{RNN-based Agents}} \\
        Baseline & 1024 & 128 & 1024 & 256 & 0.0001 & 0.00001 & 0 \\
        Impatient & 1024 & 128 & 1024 & 256 & 0.0001 & 0.00001 & 0 \\
        Impatient + Noise & 1024 & 128 & 1024 & 256 & 0.0001 & 0.00001 & 0.1 \\
        \bottomrule
    \end{tabular}
    \caption{Experimental setup for different conditions.}
    \label{tab:exp_setup_2}
\end{table}

\subsection{Communication Success}

\begin{figure}[th]
\centering
    \hspace{0.05\textwidth}
    \begin{subfigure}[b]{0.40\textwidth}
        \centering
        \includegraphics[width=\textwidth]{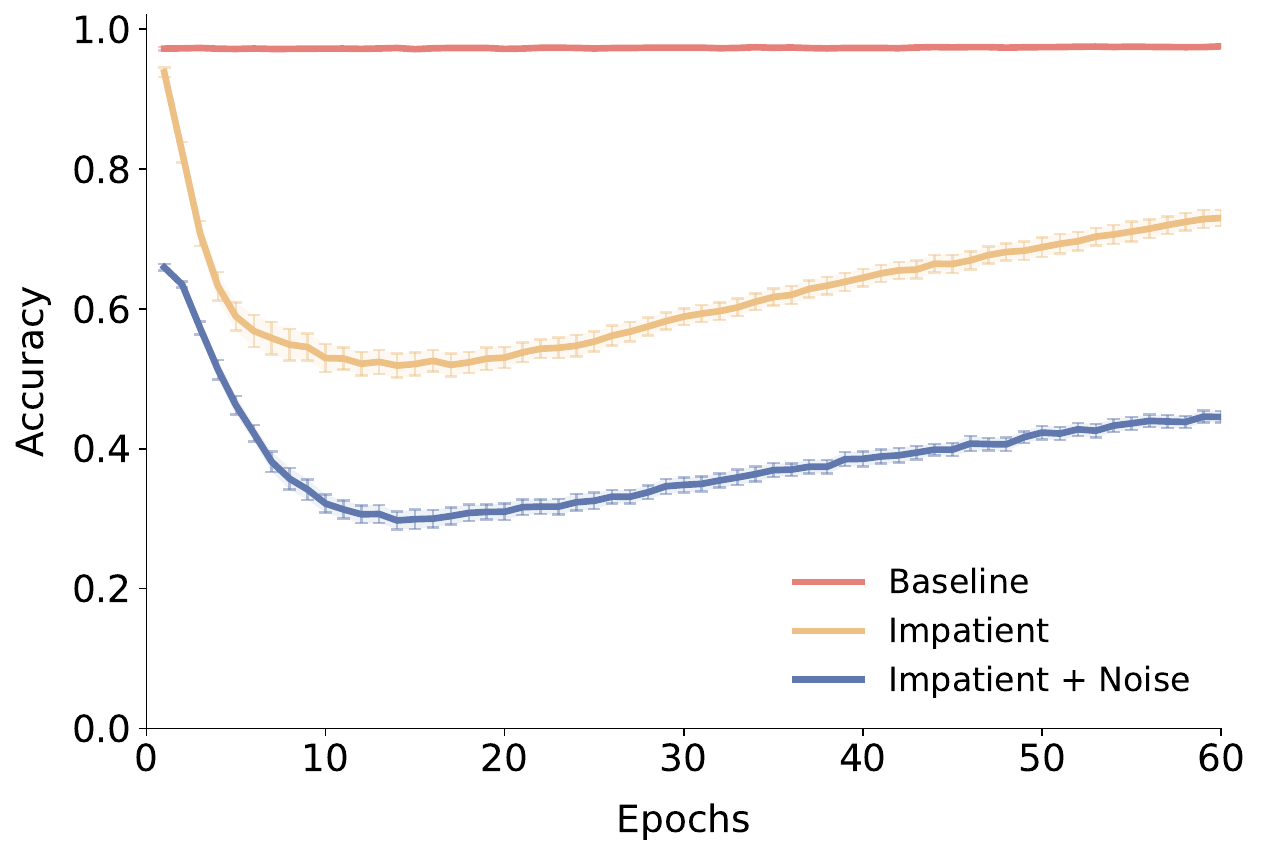}
        \caption{Verb Initial, Skewed Mixed}
        \label{fig:initial_impa_acc}
    \end{subfigure}
    \hfill
    \begin{subfigure}[b]{0.40\textwidth}
        \centering
        \includegraphics[width=\textwidth]{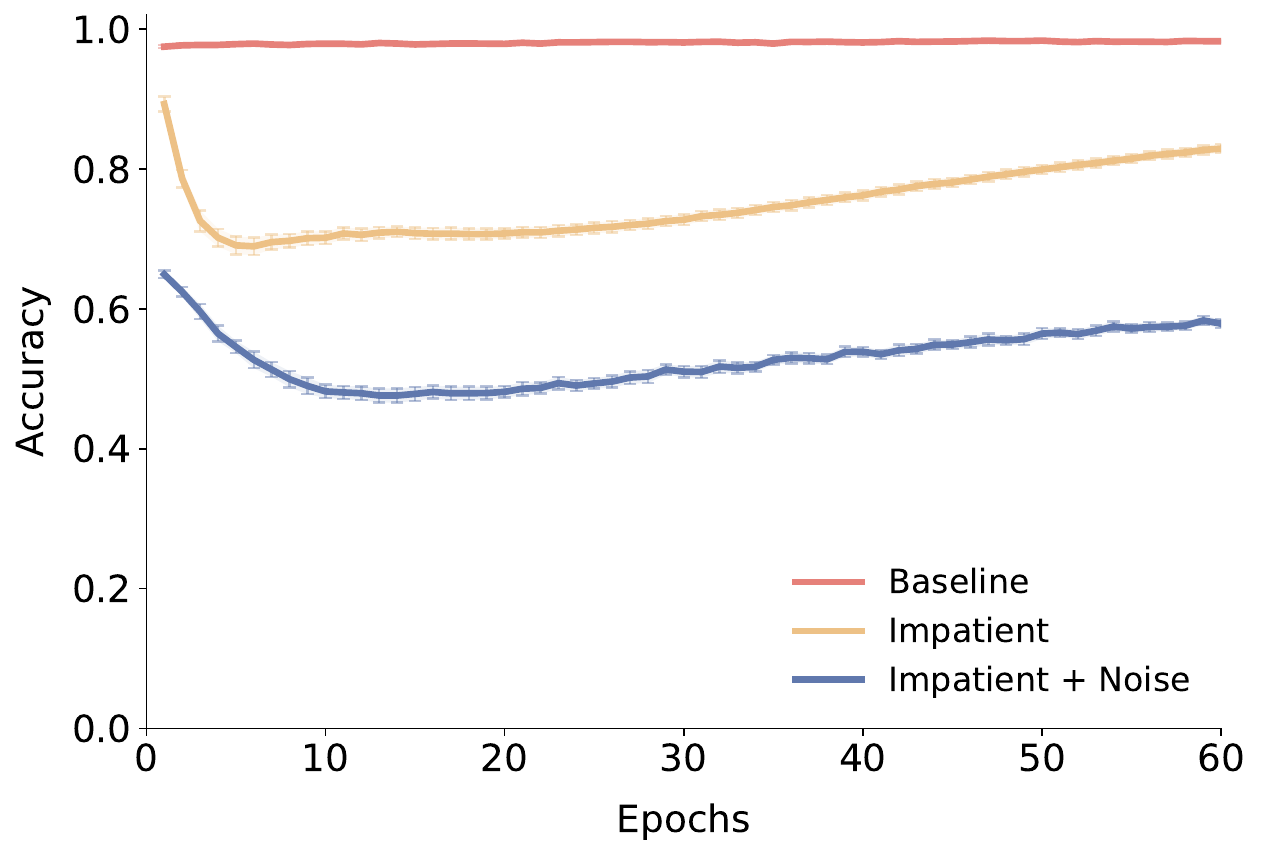}
        \caption{Verb Final, Skewed Mixed}
        \label{fig:final_impa_acc}
    \end{subfigure}
    \hspace{0.05\textwidth}
    \caption{Communication accuracy of Speakers and Impatient Listeners for the verb-initial skewed mixed language (a) and the verb-final skewed mixed language (b). 
    % with noise (top) and without noise (bottom).
    }
    \label{fig:comm_impa_combined_acc}
\end{figure}

Figure \ref{fig:comm_impa_combined_acc} illustrates the communication accuracy in the baseline condition and in the case of speakers interacting with impatient listeners for both verb-initial (Figure \ref{fig:initial_impa_acc}) and verb-final (Figure \ref{fig:final_impa_acc}) languages. 
In both conditions with impatient listeners, communication accuracy initially dropped but later recovered, approaching its initial level over several rounds of communication.
These results demonstrated robust communication performance, allowing us to proceed with an analysis of the speakers' production preferences.

\subsection{Production Preferences} 

\begin{figure}[th]
\centering
    \hspace{0.05\textwidth}
    \begin{subfigure}[b]{0.44\textwidth}
        \centering
        \includegraphics[width=\textwidth]{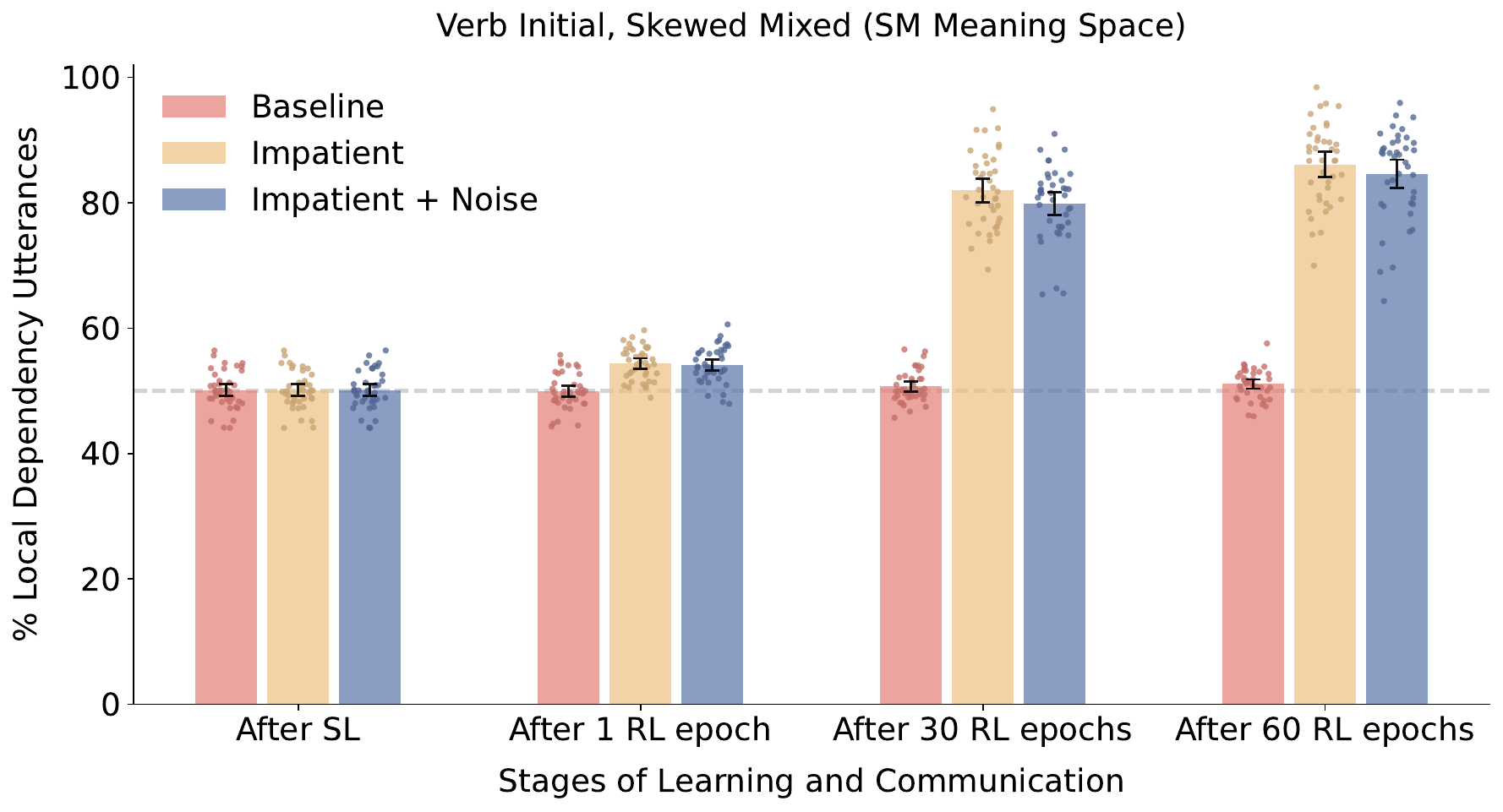}
        \caption{Verb Initial, RNN-based agents}
        \label{fig:comm_initial_impa}
    \end{subfigure}
    \hfill
    \begin{subfigure}[b]{0.44\textwidth}
        \centering
        \includegraphics[width=\textwidth]{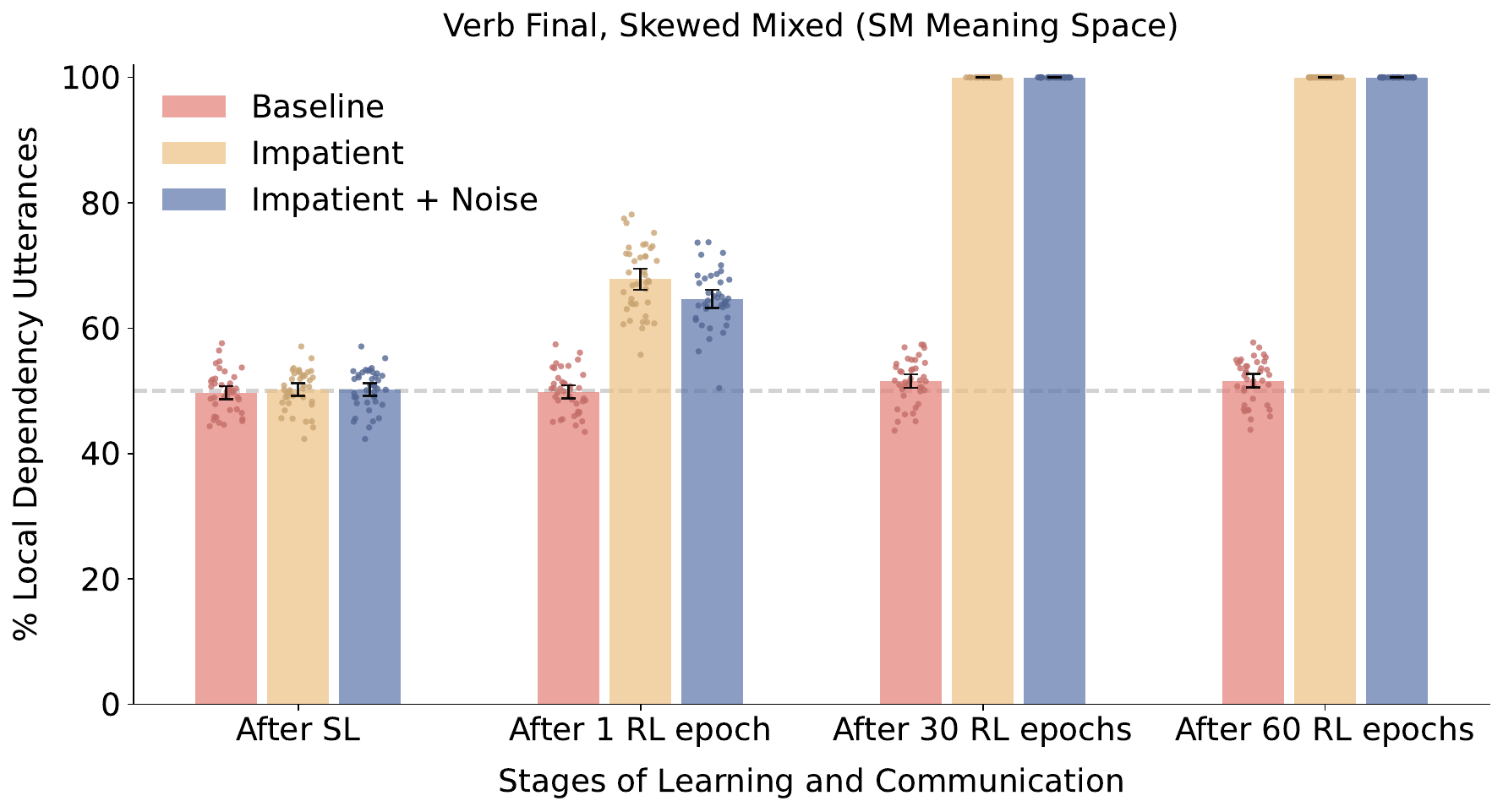}
        \caption{Verb Final, RNN-based agents}
        \label{fig:comm_final_impa}
    \end{subfigure}
    \hspace{0.05\textwidth}
    \caption{Proportion of local-dependency utterances produced by \textbf{RNN-based agents} at different stages of learning and communication using the (a) \textbf{verb-initial} or (b) \textbf{verb-final} skewed mixed language with the \textbf{half meaning space} in the three conditions: \textit{Baseline}, \textit{Impatient}, \textit{Impatient + Noise}.
    }
    \label{fig:comm_impa_combined}
\end{figure}

As shown in Figure~\ref{fig:comm_impa_combined} and Table~\ref{tab:main2}, communicating with the Impatient Listener facilitated the production of short dependency-length utterances in both verb-initial and verb-final languages.
% (Verb-initial: Comm1: (\textit{b} = xxx, 95\% CI [xxx, .xx]; Comm30: (\textit{b} = xxx, 95\% CI [xxx, .xx]; Comm60: (\textit{b} = xxx, 95\% CI [xxx, .xx]). 
Compared with the baseline condition, the Impatient Listener condition led to a higher proportion of short-dependency utterances after several RL epochs in both language types. The same pattern was observed in the \textit{Impatient + Noise} condition.
By contrast, noise did not introduce a systematic additional effect at later communication stages.
These results suggest that the primary driver of the observed behavior was the incremental processing pressure introduced by the Impatient Listener architecture.
Specifically, it incrementally processes utterances and updates its predictions word-by-word, with a reward signal that helps listeners build accurate interpretations and makes partial predictions more reliable. While verb-final languages place the verb at the end, minimizing linear dependency distance (i.e., grouping action, agent, patient together) reduces temporary ambiguity on the listener side. 
As a result, utterances that enable efficient resolution, such as those minimizing the dependency length between the verb and its arguments, yield higher rewards. 
This architectural choice not only enhances cognitive plausibility by mirroring predictive mechanisms in human online comprehension, but also naturally gives rise to more human-like patterns in production\footnote{There appears to be a consistent DLM preference, but with reversed asymmetry: verb-final languages exhibit higher local dependencies. Under the current parameter settings, the Impatient Listener setup yields higher overall communicative accuracy for verb-final languages (Figure~\ref{fig:comm_impa_combined_acc}), suggesting that the hyperparameters are effectively optimized for these agents. Importantly, modest adjustments to the learning parameters can also improve performance for agents communicating in verb-initial languages, and under certain settings, verb-initial agents can similarly converge to 100\% local-dependency production.}. 
% Individual production patterns under the \textit{Impatient + Noise} condition at the final communication stage are shown in Figure~\ref{fig:initial_plotdrop0.1_spk1024_impa} (verb-initial) and Figure~\ref{fig:final_plotdrop0.1_spk1024_impa} (verb-final). Production patterns for the remaining conditions are provided in Appendix~\ref{app:individual_prod_exp3}.

\begin{table}[ht]
\centering
\begin{tabular}{llccc}
\toprule
\textbf{Language} & \textbf{Comparison} & \textbf{After 1 RL epoch} & \textbf{After 30 RL epochs} & \textbf{After 60 RL epochs} \\
\midrule
\multirow{3}{*}{V-initial}
& Impatient vs. Baseline         & \textbf{\textit{b} = 4.44, 95\% CI [3.86, 5.02]} & \textbf{\textit{b} = 31.27, 95\% CI [29.17, 33.33]} & \textbf{\textit{b} = 35.02, 95\% CI [32.58, 37.50]} \\
& Impatient+Noise vs. Baseline   & \textbf{\textit{b} = 4.19, 95\% CI [3.61, 4.76]} & \textbf{\textit{b} = 29.20, 95\% CI [27.16, 31.27]} & \textbf{\textit{b} = 33.55, 95\% CI [31.08, 35.95]} \\
& Impatient+Noise vs. Impatient       & \textit{b} = -0.25, 95\% CI [-0.81, 0.33] & \textit{b} = -2.07 , 95\% CI [-4.21, 0.03] & \textit{b} = -1.46, 95\% CI [-3.88, 0.98] \\
\midrule
\multirow{3}{*}{V-final}
& Impatient vs. Baseline         & \textbf{\textit{b} = 17.97, 95\% CI [16.32, 19.64]} & \textbf{\textit{b} = 48.45, 95\% CI [47.57, 49.31]} & \textbf{\textit{b} = 48.39, 95\% CI [47.51, 49.26]} \\
& Impatient+Noise vs. Baseline   & \textbf{\textit{b} = 14.85, 95\% CI [13.23, 16.49]} & \textbf{\textit{b} = 48.45, 95\% CI [47.58, 49.30]} & \textbf{\textit{b} = 48.40, 95\% CI [47.50, 49.27]} \\
& Impatient+Noise vs. Impatient       & \textbf{\textit{b} = -3.11, 95\% CI [-4.73, -1.44]} & \textit{b} = -0.00, 95\% CI [-0.86, 0.88] & \textit{b} = -0.00, 95\% CI [-0.89, 0.86] \\
\bottomrule
\end{tabular}
\caption{Pairwise statistical comparisons across different communication phases.}
\label{tab:main2}
\end{table}

\subsection{Discussion}

The results of Experiment 3 provided valuable insights into how speaker agents adapted their production strategies when interacting with Impatient Listeners. By modifying the reward function to emphasize incremental accuracy, the agents were encouraged to produce utterances with shorter dependency lengths. 
This pattern was consistent with previous findings suggesting that shorter dependency lengths are efficient and preferred in human communication.
Interestingly, the production patterns for verb-final languages differed from those observed under other conditions. Whereas verb-final language speakers in Experiment 2 produced relatively more long-dependency utterances, those in the incremental processing condition favored short dependencies. 
This deviation indicated that the Impatient Listener’s incremental processing imposed stronger constraints on speakers' production preferences, yielding more human-like linguistic patterns.
In general, these findings suggest that incorporating processing constraints into neural agents can lead to emergent production behaviors that closely mirror human linguistic preferences. This highlights the potential of computational models to advance our understanding of language processing and evolution.

\section{General Discussion}

Our research demonstrates that incorporating constraints inspired by human cognitive processing into neural network-based agents facilitates the emergence of DLM within an RL-based communication framework. 
Our first contribution lies in applying the recently proposed NeLLCom framework to investigate the evolution of a well-known syntactic principle, DLM. This is achieved by predefining various artificial languages based on the artificial language learning paradigm and analyzing their changes during learning and communication. By adapting NeLLCom, we introduce a powerful learner that can be fully free of linguistic or developmental biases. 
% biases that may arise from pre-training on real-language corpora with specific linguistic properties
This approach aligns with the broader trend in artificial language learning studies, which aim to minimize explicit linguistic biases from participants’ native languages when investigating the emergence of linguistic structures \citep{culbertson_artificial_language_learning}. Moreover, while most emergent communication simulations focus on only one agent architecture, we also examine an alternative architecture in settings where it can be trained successfully, and observe partially similar tendencies.
% % raviv2018systematicity: Children may differ from adults in their biases given age-related differences in general cognitive skills.
% Moreover, adults’ performance on iterated learning tasks may reflect existing (and explicit) linguistic biases,
% partially undermining the generality of the results. % REF: Systematicity, but not compositionality: Examining the emergence of linguistic structure in children and adults using iterated learning
% --- TV: this may not be the most obvious paper to cite here without more explanation. This paper is one example of an ALL study where they looked beyond adult English speakers (by including children), which addresses indeed the current problem in the ALL field of narrow focus on subpopulations and influence of prior language knowledge. For a more broad overview paper that makes the point that ALL findings should be shown to be robust across speakers of different languages, and across stages of development, I would cite Jenny Culbertson's chapter: https://academic.oup.com/edited-volume/49437/chapter/417450728
Secondly, we model dependency relationships through skewed conditional word distributions between agents and patients. Results suggest that introducing selectional preferences strengthens listeners' ability to recover missing elements in noisy environments, highlighting the effectiveness of this approach in modeling more realistic semantic constraints and word sense disambiguation cues.
Thirdly, we study the impact of several processing-related constraints, including noise during listening, limited speaker capacity, and incremental sentence processing. 

The results show three key patterns: 
in the original setup with full-meaning space, word order regularization overrides DLM (Exp. 1); 
in the corrected half-meaning space setup, short-before-long preferences override DLM, a tendency further facilitated by noise and limited speaker capacity (Exp. 2); and 
explicitly modeling incremental sentence processing in the listener successfully elicits DLM (Exp. 3).
These simulation results suggest that DLM can arise from the interaction between speaker-side production choices and listener-side comprehension pressures. 
In the standard communication settings, agents' production preferences are dominated by word-order regularization or short-before-long ordering. By contrast, when the listener is rewarded for interpreting the utterance incrementally, shorter dependencies become communicatively advantageous.
Together, these findings suggest that incorporating such cognitive and processing constraints into neural agents can lead to emergent production behaviors that mirror human-like preferences more closely.
Below, we discuss the conditions under which DLM overrides the short-before-long preference in neural agents, the implications of using neural networks to simulate language emergence, limitations of the current study, and potential future directions.

\subsection{From Short-before-Long Bias to DLM in Neural Agents}

Experiment 2 shows that under conditions of limited processing capacity or increased environmental noise, neural agents tend to position the short unmodified constituent as early in the sentence as possible. These results indicate that, in the absence of an incremental prediction listener, agents consistently prefer to place the short constituent earlier in sentences in both verb-initial and verb-final languages, displaying a short-before-long preference \citep{behaghel1909beziehungen}.
% Under conditions of limited processing capacity or increased environmental noise, the production behavior of neural network-based agents appears to be strongly constrained by form-related factors, i.e., a consistent preference for placing shorter elements before longer ones in both verb-initial and verb-final languages (a short-before-long preference \citep{behaghel1909beziehungen}).

% Regarding the differences in the degree of DLM in verb-initial and verb-final languages, experiment 2 showed that in verb-initial languages, 
% neural agents tend to position the unmodified constituent closer to the head verb and the modified constituent further away. In other words, neural agents have a production bias towards those word orders that yield shorter DL when the longer DL alternative is available to convey the same meaning. In contrast, in verb-final languages, agents prefer to shift the modified constituent closer to the verb, suggesting that agents prefer to produce long dependency-length utterances. 
% These results indicate that neural learners without the incremental prediction mechanism seem to consistently prefer to place the short constituent earlier in sentences. % short-before-long preference, constituent weight-based shift, short to the front

%%%%%%%%%%%%%% human theory
Based on the grammatical encoding theory, human word order choices reflect a balance between two contrary biases: conceptually driven and form-driven processes \citep{bock1982toward}.
% Some choices are more conceptually driven and others are more form-driven \citep{bock1982toward}. 
Conceptually, there is a tendency to place salient elements earlier in sentences \citep{osgood1977salience}, while form-driven decisions involve incrementally selecting lexical items \citep{bock1982toward, ferreira1996better}. In this theory, long phrases have competing characteristics. 
On the one hand, they are semantically richer and therefore more salient, which increases their accessibility in the conceptual domain.
% They are semantically richer because the extra lexical items enhance the meaning of the head noun, which makes them more salient. This semantic richness increases the overall accessibility of the phrase in the conceptual domain. 
On the other hand, in the form domain, they are longer in terms of lexical content, which increases the number of tokens in a sequence. This might increase the likelihood of simpler phrases being ready before complex phrases, and an overall preference for putting short phrases before long phrases in the form domain. In natural languages, these two contrary biases are delicately balanced against each other.
One possible explanation for the observed short-before-long bias lies in the sequential nature of recurrent generation. RNN-based speakers generate utterances incrementally, updating their hidden state after each token. Starting with a shorter constituent may therefore provide a simpler and more stable initial generation trajectory, whereas beginning with a longer constituent requires the model to generate a more extended sequence before transitioning to the remaining material. This asymmetry could create a production-side preference for short-before-long ordering.
However, this interpretation remains tentative. Further analysis of hidden-state dynamics would be needed to determine whether this sequencing account underlies the observed preference.

% Under incremental processing, DLM emerges beyond a short-before-long bias, suggesting that this condition plays a key role in shifting behavior toward genuine minimization.
Importantly, when listener agents are equipped with a prediction mechanism as impatient listeners, speaker agents consistently produce utterances with shorter DLs. This finding highlights the role of listener expectations and processing strategies in shaping language production. 
% A listener model that incrementally predicts upcoming words, creates pressure for the speaker to minimize the processing burden on the listener side. This leads to a shift in production toward short DLs, likely because such structures facilitate more efficient prediction. 
Under incremental listener constraints, the listener must interpret the utterance word-by-word and commit to partial interpretations online. In this setting, the position of the head relative to its dependents directly affects how quickly informative head-dependent relations can be established. Here, minimizing linear dependency distance may reduce temporary ambiguity in the listener. This is possibly why, under incremental processing, we observe DLM effects rather than a simple short-before-long bias.
The result also supports theories of language use where comprehension constraints, especially those involving predictive processing, influence not only how language is understood but also how it is generated. This aligns with psycholinguistic and computational perspectives suggesting that communicative pressures are bidirectional: with speaker and listener behavior mutually shaping one another \citep{clark1986referring, jaeger2010redundancy}.
In addition, DLM is more frequently observed in comprehension-oriented tasks than in production \citep{gibson1998linguistic}. This asymmetry suggests that models may rely on different mechanisms in encoding vs. generating linguistic sequences.

Finally, although we did not successfully adapt Transformer-based agents in the current experiments, our findings still allow us to make tentative predictions about their behavior. 
Because Transformers do not rely on the same recurrent hidden-state dynamics as RNNs, we would not expect them to develop the same short-before-long production bias observed in our recurrent agents. Instead, their production preferences may be more strongly shaped by input distributions and positional biases. 
However, DLM-compatible behavior could still emerge if Transformer-based agents were trained with an incremental listener objective, because in that case shorter dependencies would help the listener better recover the intended meaning. 
A direct comparison between RNN-, GRU-, and Transformer-based agents therefore remains an important direction for future work.

\subsection{Implications on Using Neural Networks to Simulate Language Emergence}

Artificial neural networks rely primarily on statistical learning of language data to develop representations, assuming minimal inductive biases specialized for language.
% though their architecture inherently introduces structural biases. 
Computational simulations with neural networks thus provide a valuable platform for testing which objectives or inductive biases are necessary to reproduce observed patterns in language behavior.

At the same time, while neural networks can sometimes replicate patterns observed in human learners, they may emerge through mechanisms that differ substantially from those underlying human cognition. For example, a neural model might exhibit a linguistic behavior due to optimization strategies aimed at reducing computational cost, rather than constraints analogous to human memory or processing limitations. 
This highlights the importance of caution when interpreting neural-network-based simulation results. 
Importantly, the present study models agent experiments closely after prior human experiments in artificial language learning, as done by \citet{lian2023communication, lian-etal-2024-nellcom, lian2025simulating, kouwenhoven2025searching} 
to identify whether the emergent pattern aligns with hypothesized cognitive mechanisms. In addition, we analyze how specific architectural features or learning objectives influence emergent patterns, which helps disentangle whether neural networks are genuinely modeling human-like processes or simply optimizing for task-specific goals. 
By combining simulations with the validation of human experiments, we can better ensure that neural-network-based simulations provide meaningful insights into the mechanisms of language evolution, rather than reinforcing misleading parallels between artificial and human learners. % leveraging complementary approaches

From another perspective, traditional NLP models often incorporated explicit biases, such as preferences for short dependencies, to enhance performance \citep{collins2003head, nivre2006inductive}. In contrast, modern neural architectures typically avoid such built-in biases to maximize flexibility. While this increases their generality, it may be beneficial to reintroduce inductive biases that encourage linguistically plausible behavior, particularly when simulating human-like language processing. 
% TV: cite some of the work in Tom Griffiths' group, where they tried doing this with a meta-learning approach
Notably, recent work has explored using meta-learning approaches to instill such inductive biases in neural models, enabling them to better simulate human-like reasoning and learning. For instance, meta-learning frameworks have been proposed to develop biases akin to Bayesian priors \citep{griffiths2019doing, grant2018recasting, mccoy2025modeling, lake2023human, goodale-etal-2025-meta}. % lake2023human -- compositionality
 
\subsection{Limitations and Future Extensions}

We acknowledge the importance of studying how processing constraints shape DLM. In this study, we model processing constraints by adapting the Impatient Listener architecture \citep{rita-etal-2020-lazimpa} to the communication game, which incrementally processes input and embodies working memory in a cognitively plausible manner. In addition, we model memory limitation by manipulating the speaker capacity.
Other, more direct ways to model memory constraints could be explored, such as the approach of \citet{timkey2023language}, which extends simple recurrent networks with self-attention, allowing retrieval of past representations. Similarly, \citet{mita-etal-2025-developmentally} model human-like working memory in Transformers using Attention with Linear Biases, which applies a distance-dependent penalty to attention scores. This recency bias allows the model to capture human memory decay and reading-time patterns, providing another cognitively plausible approach to memory limitations in neural learners.
Future work could also explore methods to ensure that neural models develop such biases naturally, aligning their emergent behaviors more closely with those of human learners.

Another key focus should be on modeling dependency relationships and meaning spaces more accurately. Currently, our meanings are restricted to simple transitive events. This simplification lacks hierarchical recursion (e.g., relative clauses) and discourse pragmatics.
Unlike simple sequential dependencies, natural language often involves intricate interdependencies between meaning slots that cannot always be adequately captured by linear representations.
Designing more natural miniature languages with varying dependency types and adopting graph-based representations \citep{kipf2017semi} could provide a more effective framework for handling non-linear and non-sequential dependencies that frequently occur in complex sentences. Future work should also systematically manipulate sequence length. 
Our current setup fixes utterance length, which prevents us from testing whether DLM and competing ordering pressures depend on sequence length. 
Varying utterance length would make it possible to examine whether sentence-length-dependent predictions from the syntactic literature also arise in neural-agent communication.

Another promising direction would involve integrating learning and communication phases more dynamically \citep{lowe2020interaction}. In our current setup, these phases are distinct: first, agents are exposed to linguistic input to acquire the structure and patterns of the language system, and subsequently, they use this knowledge in a communication phase to convey messages within a shared meaning space. Blending learning with communication could provide insight into how real-time usage influences language acquisition and system structure. Another possible direction is to explore character-level modeling, where sequence length and dependency distance are measured in characters rather than words. This would allow investigation of how DLM interacts with morphological compression, such as the tendency for frequent syntactic heads or grammatical markers to shorten, potentially further minimizing character-level dependency distances.
A final limitation concerns the degree of experimental control in neural-agent simulations. Our setup controls key linguistic variables, but training duration, architectural hyperparameters, random initialization, and interaction structure may still affect the results. Future work should test the robustness of these findings across broader model settings and richer multi-agent interaction structures.

% it is good to acknowledge that the method has limitations but end with a strength or suggestion for future work that is more concrete like "the need for testing multiple agent architectures, like we did here, while most emergent communication works don't"? --- done

\section{Conclusions}

% The total dependency length is influenced by:
% Word order (e.g., SVO, VSO, SOV, etc.) --- our exploration is not exhaustive
% Branching direction (head-initial vs. head-final)
% Sentence length and structure

% Anti-DLM effects suggest that DLM is not the only force in shaping word order. Other competing pressures include: Information structure	Topicalization / focus
% Syntactic parallelism	Keeping related structures aligned
% Disambiguation	Longer distance = less short ambiguity
% Prosody or rhythm	Heavy phrases delayed
% Center embedding / parsing constraints	Sometimes longer dependencies help processing
% short Ambiguity Avoidance: Gildea & Temperley (2010) suggest that some word orders increase dependency length to reduce short ambiguity (e.g., disambiguate subject vs. object early).

This study incorporates several factors (noise during listening, limited speaker capacity, incremental sentence processing) to improve the realism and cognitive plausibility of neural-agent simulations of language emergence and applies these to the study of DLM. 
Our experiments with neural learners exposed to an % not creole-like, just one language with unpredictable variations 
artificial language with mixed patterns reveal that noise during listening and limited speaker capacity are not sufficient to elicit a consistent DLM preference across verb-initial and verb-final languages.
%certain pressures are essential for the agents to exhibit productions consistent with a human-like DLM preference. 
% Specific results differ by type of language:
% For learners of verb-initial languages, noise during listening and limited speaker capacity lead to the expected DLM preference. However, for learners of verb-final languages, noise and limited capacity alone do not result in the expected DLM preference but in the opposite pattern. 
%Interestingly, % for both types of languages with a half-meaning space, 
Instead, agent productions appear to be consistent with a preference for short-before-long constituents. This short-before-long bias mirrors production studies where speakers place shorter nominal phrases earlier, often attributed to lexical accessibility or production efficiency (e.g., Behaghel’s law of increasing constituents \citep{behaghel1909beziehungen}). 
Only the introduction of incremental processing leads to agent productions that are consistent with DLM in both types of languages.

The various factors examined in this work form a combination of internal and external constraints on perception, processing, and memory. %Noise is an external factor, while reduced capacity and incremental processing are internal memory constraints.
For agents whose primary objective is to maximize communicative success, each of these constraints exerts selective pressure on the systems that emerge, shaping the strategies and preferences that agents develop.
The interaction between these pressures is highly non-linear, often leading to trade-offs or unexpected synergies. Computational modeling provides a powerful means of exploring these dynamics, allowing us to disentangle their contributions and study how they jointly shape communicative behavior.
%Our study highlights the importance of setting up realistic simulations and testing multiple agent architectures to improve the generalizability of findings from neural-agent simulations to the study of language universals. %especially compared to most emergent communication work, which often relies on a single setup.

\section{Data availability}
All the experiment data and scripts to replicate the experiments are available in the following repository \url{https://github.com/yuqing0304/DLM_exp}.
% \url{https://osf.io/y9f34/?view_only=de229a10a52c4803bb4d45cce28c54af}.

\section{Acknowledgments}
Arianna Bisazza is funded by the Talent Programme of the Dutch Research Council (NWO) under project VI.Vidi.221C.009. 

%\backmatter
% \bmsection*{Author contributions}

% This is an author contribution text. 

% % \AB{TODO add grants!!}

% \bmsection*{Financial disclosure}

% None reported.

\bmsection*{Conflict of interest}

The authors declare no potential conflicts of interest.

\bibliography{wileyNJD-Harvard} 

% \bmsection*{Supporting information}

% Additional supporting information may be found in the
% online version of the article at the publisher’s website.

\appendix

\bmsection{Data generation details}
\label{app:date_gen}
To generate the skewed version of each language, we use the following discrete distributions which roughly correspond to ten equally spaced sample points from the Zipf distribution by setting the number of elements $N$ to 10 and the exponent parameter $s$ to 1.5: $P_{\text{agent}}$ = [0.501, 0.177, 0.096, 0.063, 0.045, 0.034, 0.027, 0.022, 0.019, 0.016] for the probabilities of ten randomly shuffled animate nouns being agent given each verb, and the same distribution for the probabilities of ten randomly shuffled animate nouns being patient given each verb.
By contrast, the uniform versions assign an equal probability to all animate nouns for being agents or patients with each verb. 

In order to make a controlled setting and analyze the impact of skewed distribution on the development of DLM preferences in neural learners, we simplify the artificial language by assuming that all meaning tokens, except for agent and patient, have equal probability.  

For each action, the meanings for both SM and OM spaces are generated by first exhaustively combining all available adpositions, adjectives, and inanimate nouns, resulting in $4*3*3=36$ unique combinations. Each of these combinations is duplicated to augment the data and facilitate the training process.
% \AB{what does this mean?}
Each combination is further expanded by concatenating it with an agent.
% \AB{and this?} 
The agents are sampled based on their respective conditional word distributions with the action, resulting in $72*0.501+72*0.177+72*0.096+72*0.063+72*0.045+72*0.034+72*0.027+72*0.022+72*0.019+72*0.016\approx$67 combinations. % 36+12+6+4+3+2+1+1+1+1=67
This new combination is duplicated and expanded by adding the meaning category of the patient. This is done by sampling patients based on their conditional probability ratios with the action, resulting in $134*0.501+134*0.177+134*0.096+134*0.063+134*0.045+134*0.034+134**0.027+134*0.022+134*0.019+134*0.016\approx$129 combinations. % 67+23+12+8+6+4+3+2+2+2=129
A total of $129*10(actions)*2(SM/OM)=2580$ meanings are generated. Similar to human experiments, we assume that the agent and patient are never the same within a single meaning. Therefore, certain meanings are filtered out, and repeated meanings are excluded as well. We sample half of the SM meanings and then generate short dependency utterances, and then generate long dependency utterances for the other half of the SM meanings. Then we repeat the same steps for OM meanings.

Generating data using this simple method can guarantee that items in categories other than Action, agent, and patient occur with equal probability and that the conditional probabilities of action, agent, and patient obey the assigned discrete skewed distribution.
This process results in a comprehensive set of meaning structures that are representative of the dependency relationships in real languages. 

\begin{figure}[t]
\centering
    \includegraphics[width=0.9\columnwidth]{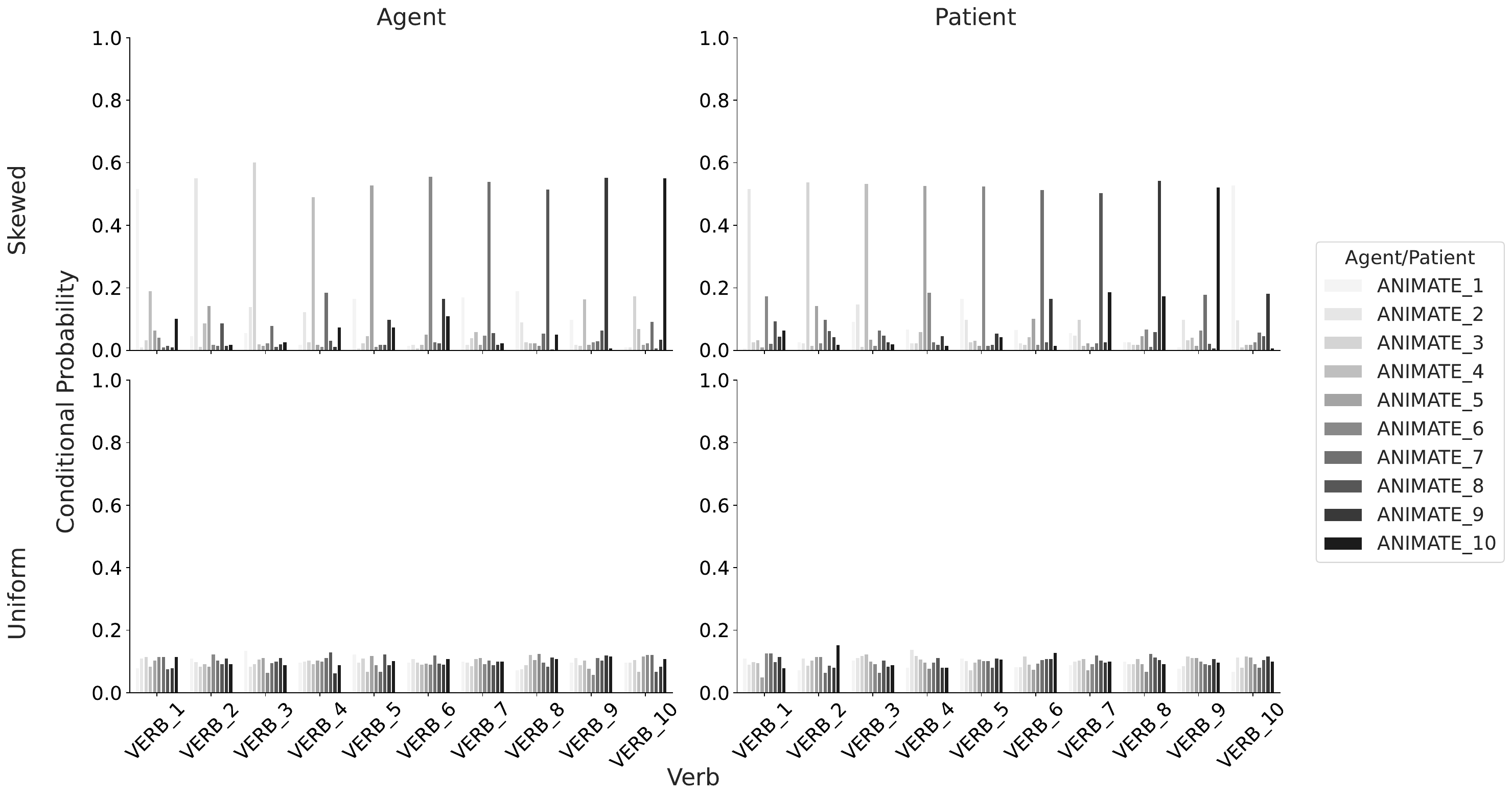}
  \caption{Conditional probability of Agent (left) and Patient (right) given Actions for skewed (top) and uniform (bottom) languages.}
  \label{fig:exp2_data_distr}
\end{figure}

For the uniform version, we generate all possible combinations of verbs, subjects, objects, and their modifiers. Then randomly sample an equal number of meanings. Data distributions are shown in Figure~\ref{fig:exp2_data_distr}.

\begin{figure*}[t]
\centering
\includegraphics[width=\linewidth]{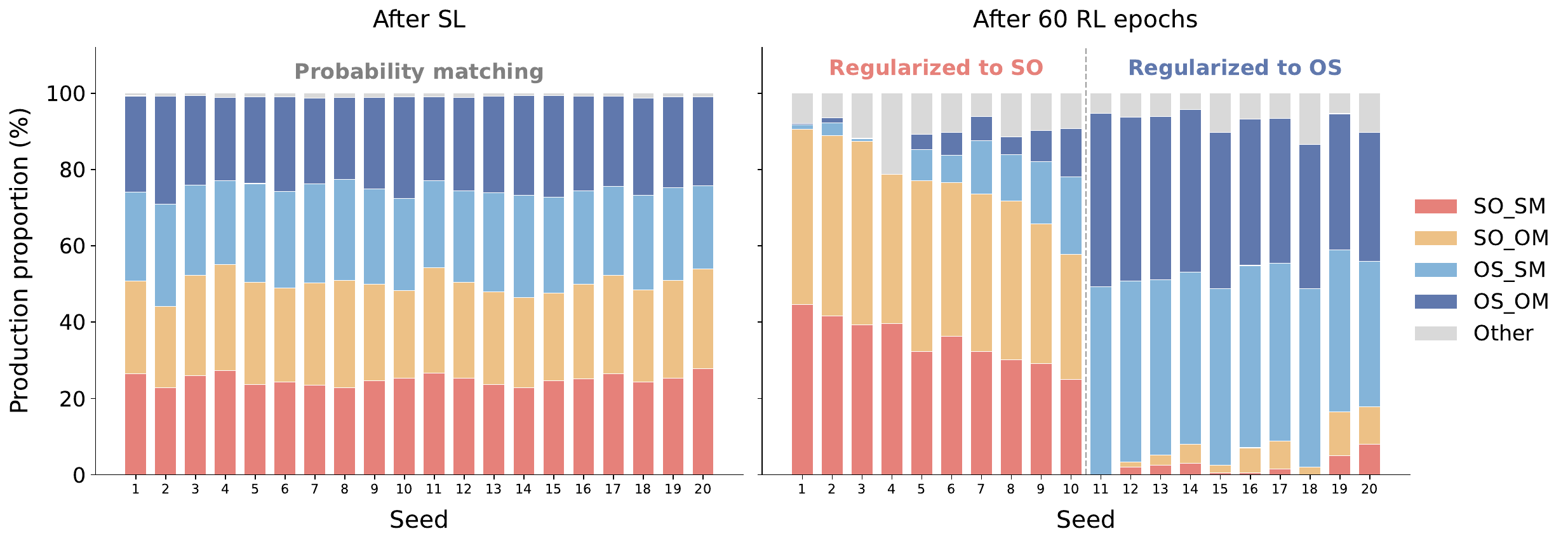}
\caption{
Seed-level production patterns before and after communication for the \textbf{verb-initial} \textit{skewed mixed} language with the \textbf{full meaning space} and \textbf{dropout rate 0.1} (Noise condition).
Each vertical bar represents one random seed and sums to 100\% of that seed's productions. 
Colors encode both word order and meaning type: SO\_SM, SO\_OM, OS\_SM, and OS\_OM denote productions using SO or OS order for subject-modified (SM) or object-modified (OM) meanings, respectively; grey denotes other or incorrect productions. 
The left panel shows productions after SL, where agents largely preserve the mixed input distribution, consistent with probability matching. 
The right panel shows productions after 60 communication epochs, where seeds are sorted by their dominant word order and individual agents regularize toward either SO or OS. 
This pattern shows that, under dropout, communication still induces word-order regularization at the seed level, even though the aggregate proportion of local-dependency utterances can remain close to the 50\% input baseline.
}
\label{fig:spk_comm_regularization_drop01}
\end{figure*}

\bmsection{Individual Production Patterns under the noisy condition in Experiment 1}
\label{app:individual_prod}
This appendix includes the production patterns of 20 speaking agents (random seeds) after learning or communication for the verb-initial \textit{skewed mixed} language, word dropout 0.1 (Figure~\ref{fig:spk_comm_regularization_drop01}). Around half regularize towards SO word order and another half regularize towards OS word order. The production patterns for other settings (verb-final, uniform languages, with dropout) are all similar in terms of regularization strategies and are therefore not shown here.

% \begin{figure*}[h!]
% \centering
%   \includegraphics[width=\linewidth]{figs/spk_comm20drop0.1.jpg}
%   \caption{Production patterns of different speaking agents (random seeds) during communication learning for the \textbf{verb-initial} skewed mixed language with the \textbf{full meaning space}, word dropout = 0.1. Color denotes word order. blue: OS, red: SO. Shading denotes dependency length. dark: short, light: long. Utterances not belonging to these categories are colored in grey. Meaning reconstruction accuracy per seed is shown by the blue line.}
%   \label{spk_commorder_initialskewed0.1}
% \end{figure*}

% \bmsection{Individual Production Patterns for Verb-initial Languages in Experiment 2}
% \label{app:individual_prod_exp2}

This appendix includes the production patterns of 40 speaking agents (random seeds) during communication learning for the verb-initial \textit{skewed mixed} language with only SM meaning space.

\vspace*{12pt}

\bmsection{Per-category listening accuracy in the verb-initial language}
\label{app:per_cat_acc_initial}

\begin{figure}[!ht]
\centering
    \begin{subfigure}[b]{0.19\textwidth}
        \centering
        \includegraphics[width=\textwidth]{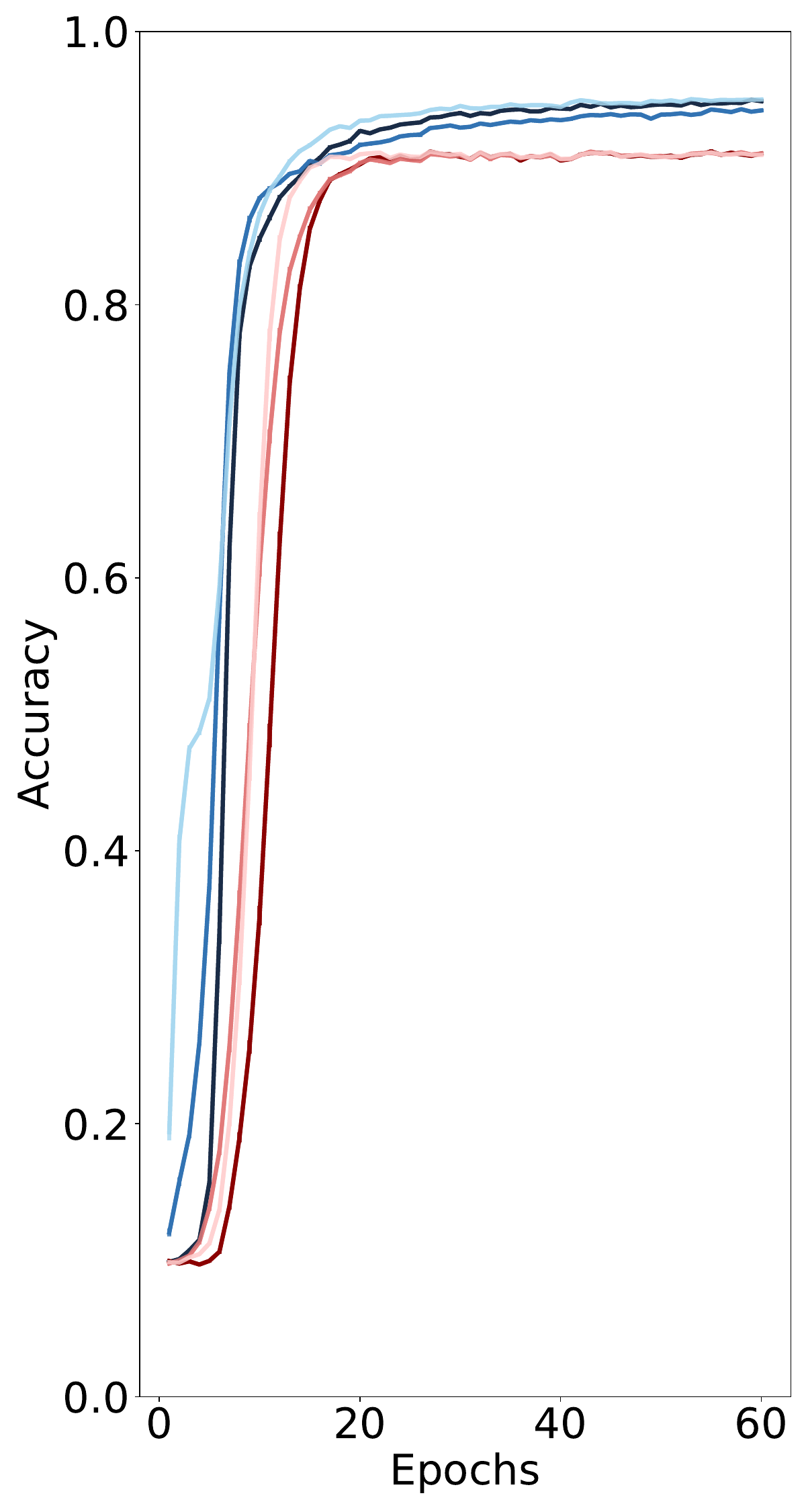}
        \caption{Verb}
        \label{fig:initial_lst_acc_verb_rnn}
    \end{subfigure}
    \hfill
    \begin{subfigure}[b]{0.19\textwidth}
        \centering
        \includegraphics[width=\textwidth]{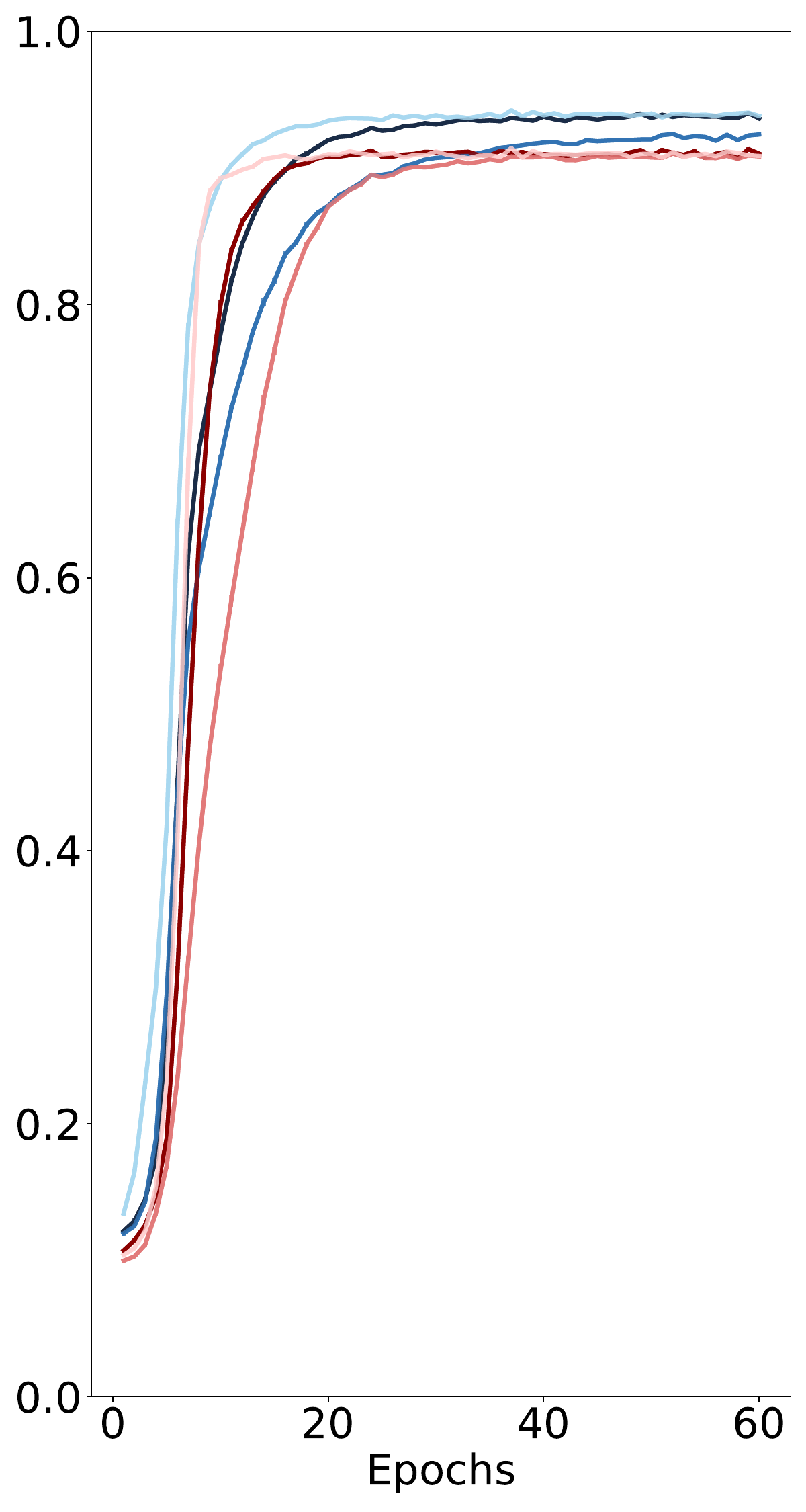}
        \caption{Subject}
        \label{fig:initial_lst_acc_subject_rnn}
    \end{subfigure}
    \hfill
    \begin{subfigure}[b]{0.19\textwidth}
        \centering
        \includegraphics[width=\textwidth]{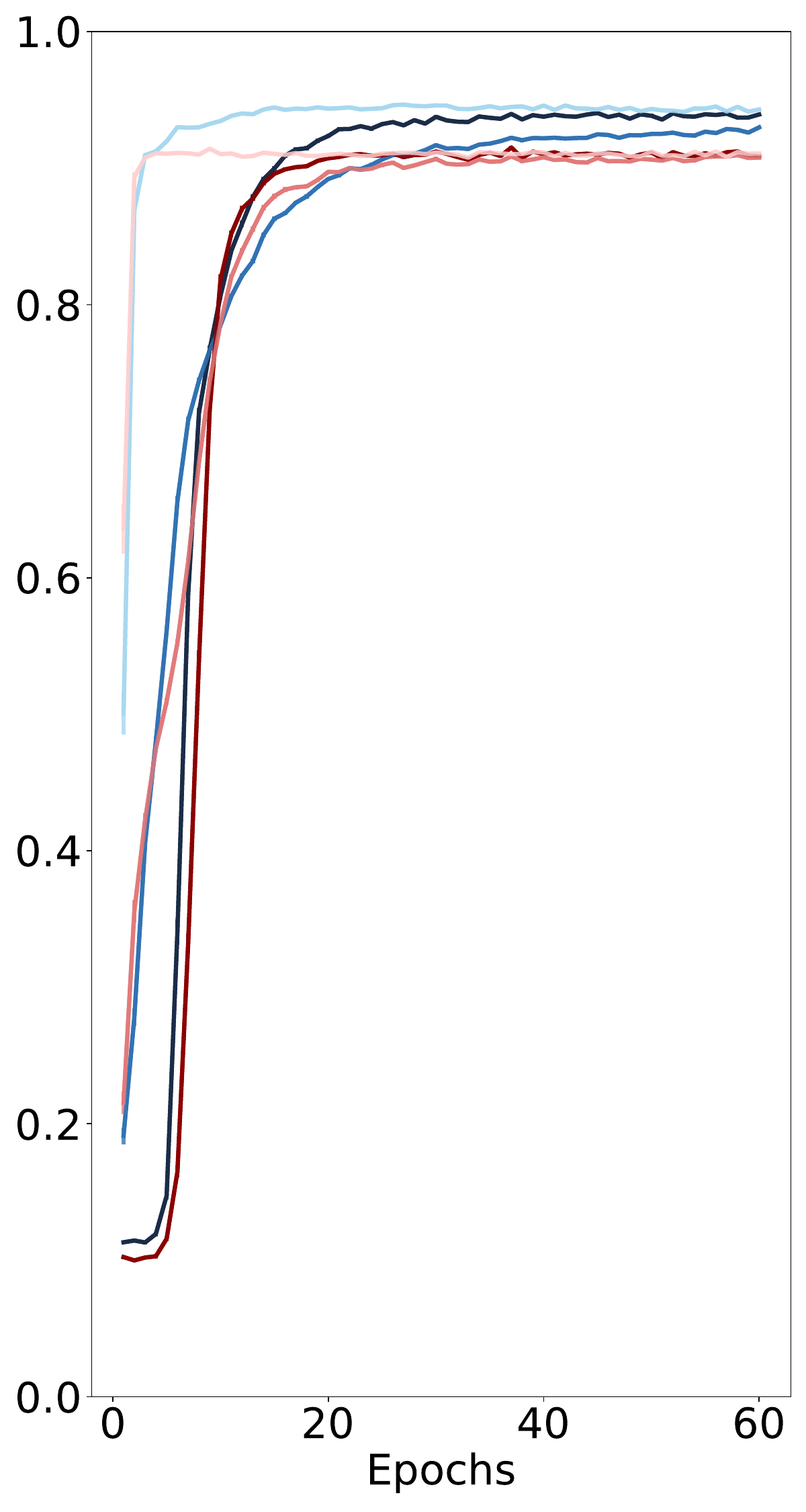}
        \caption{Object}
        \label{fig:initial_lst_acc_object_rnn}
    \end{subfigure}
    \hfill
    \begin{subfigure}[b]{0.19\textwidth}
        \centering
        \includegraphics[width=\textwidth]{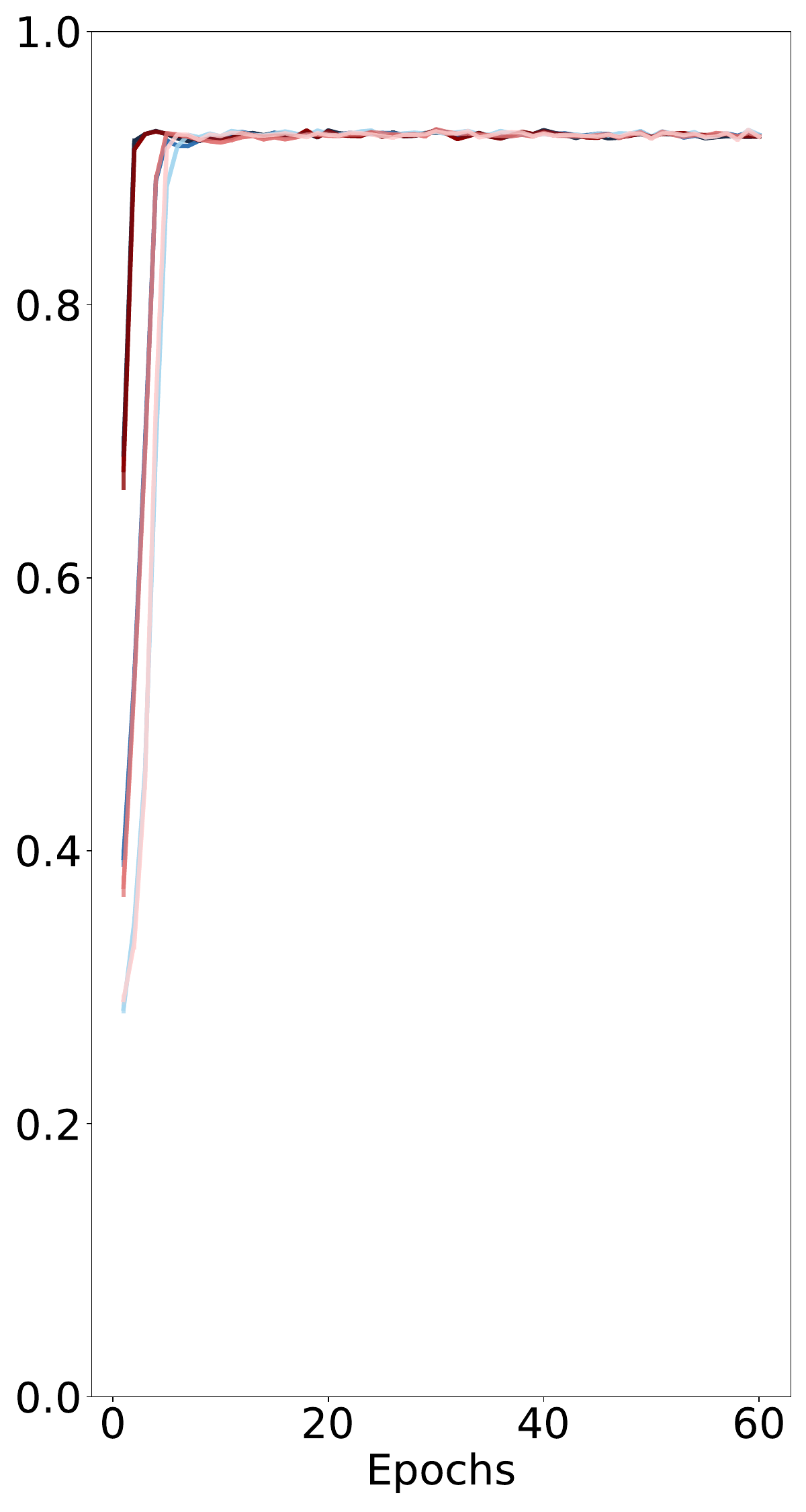}
        \caption{Modifier 1}
        \label{fig:initial_lst_acc_subjectm1_rnn}
    \end{subfigure}
    \hfill
    \begin{subfigure}[b]{0.19\textwidth}
        \centering
        \includegraphics[width=\textwidth]{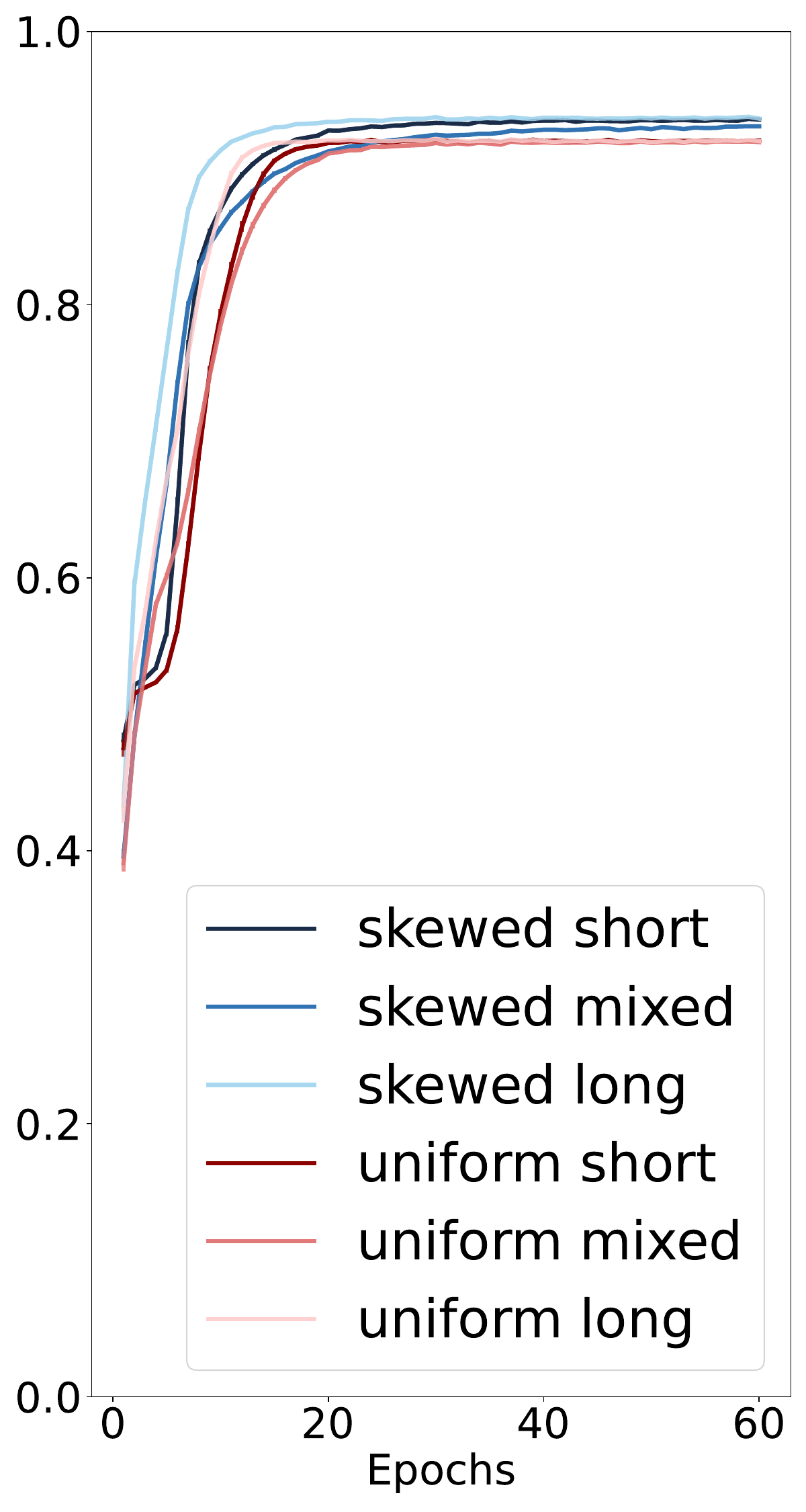}
        \caption{Avg Acc}
        \label{fig:initial_lst_acc_rnn_avg}
    \end{subfigure}
    \caption{Listening accuracy computed separately for each individual meaning item in the verb-initial language, 
  as a function of the training epoch. 
  We see similar results for $\{ \mu_a^1,\mu_a^2,\mu_a^3\}$. Therefore, only $\mu_a^1$ is plotted here for illustration.}
    \label{fig:per-category-acc-epoch-initial}
\end{figure}

This appendix shows the listening accuracy computed separately for each individual meaning item and the average listening accuracy in the verb-initial language as a function of the training epoch (Figure~\ref{fig:per-category-acc-epoch-initial}). 

\vspace*{12pt}

\begin{figure}[ht]
\centering
    \begin{subfigure}[b]{0.19\textwidth}
        \centering
        \includegraphics[width=\textwidth]{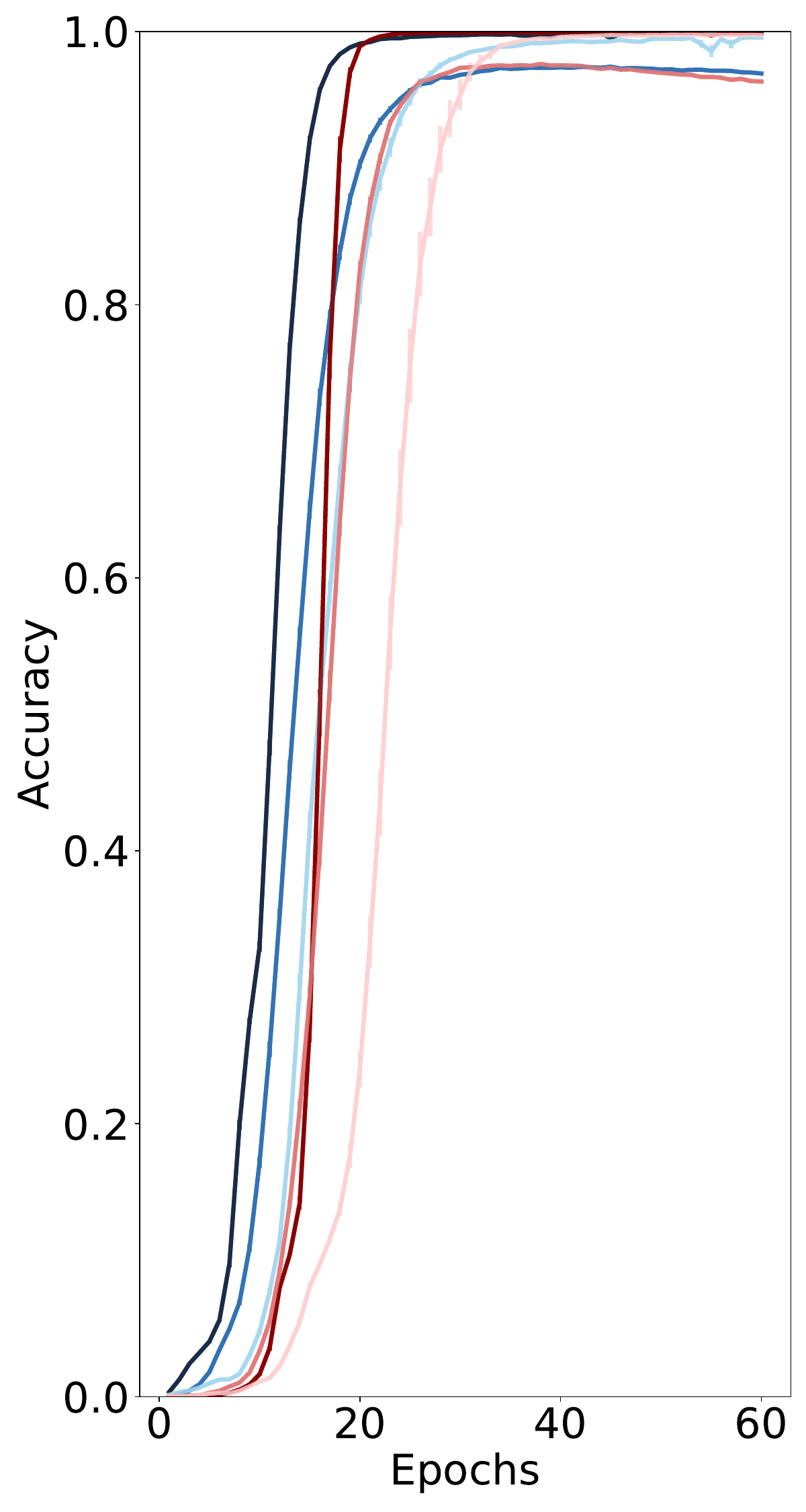}
        \caption{Speaking}
        \label{fig:final_spk_acc_rnn}
    \end{subfigure}
    \hfill
    \begin{subfigure}[b]{0.19\textwidth}
        \centering
        \includegraphics[width=\textwidth]{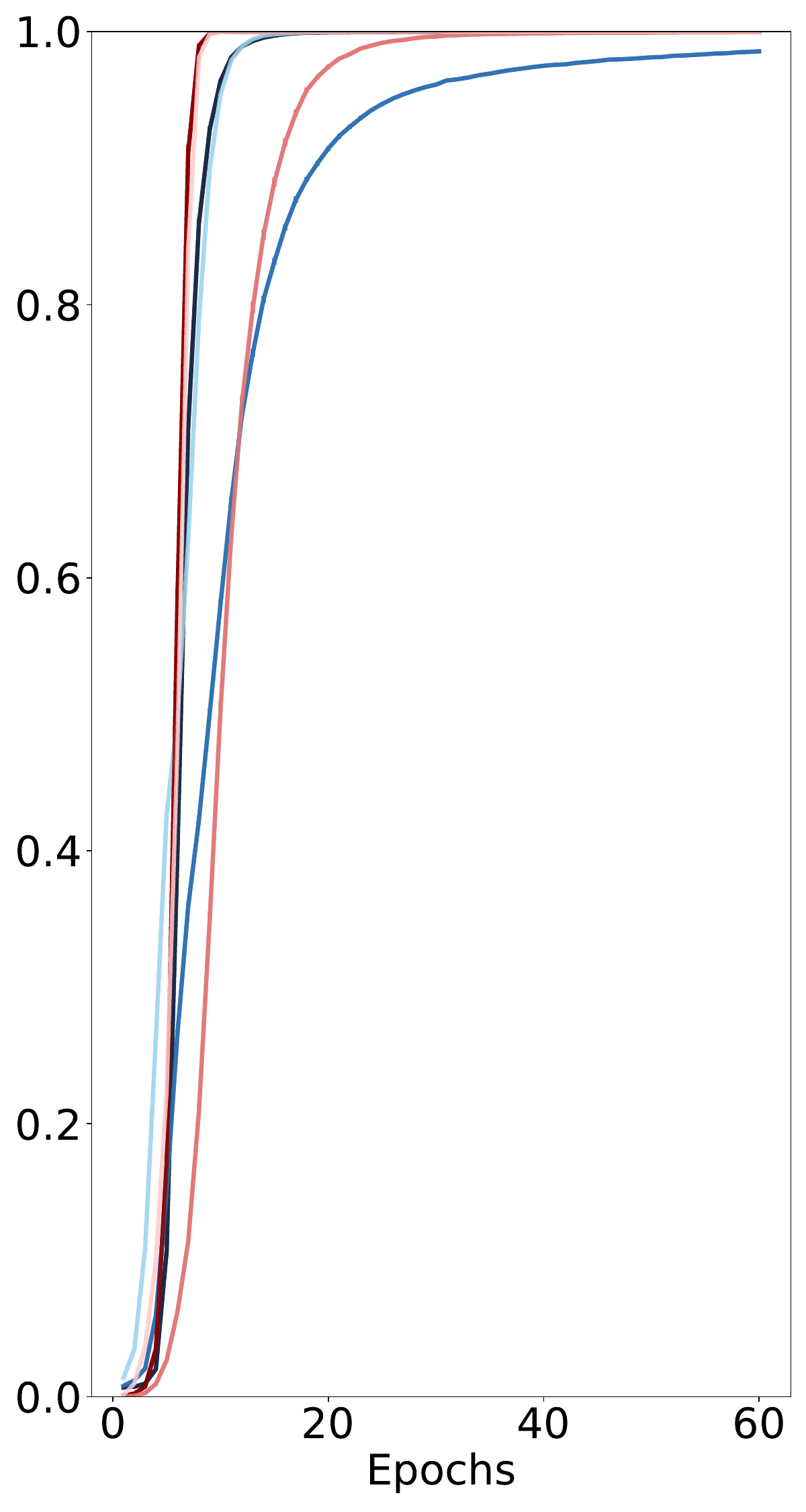}
        \caption{Listening (noise 0)}
        \label{fig:final_lst_acc_rnn}
    \end{subfigure}
    \hfill
    \begin{subfigure}[b]{0.19\textwidth}
        \centering
        \includegraphics[width=\textwidth]{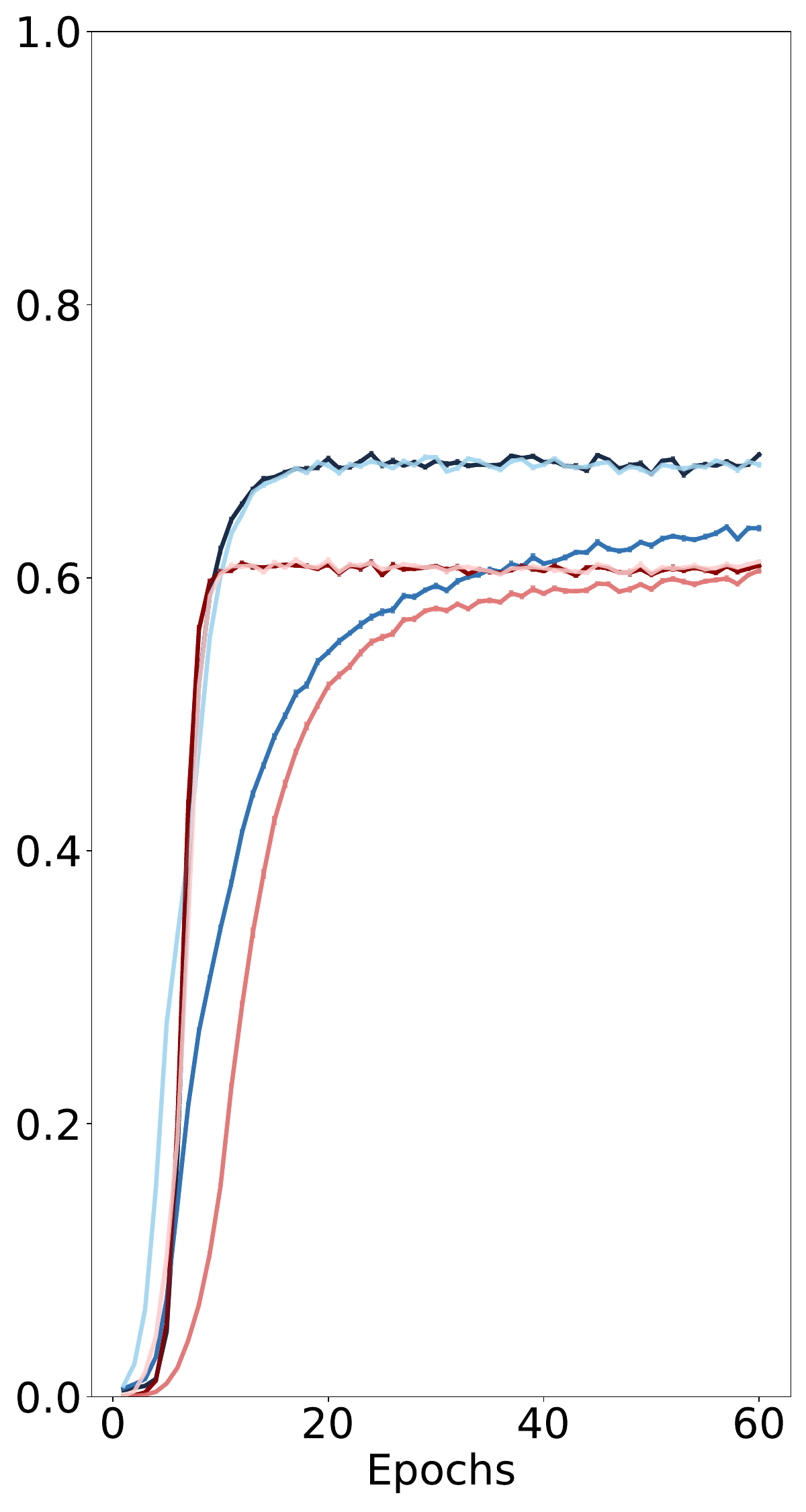}
        \caption{Listening (noise 0.1)}
        \label{fig:final_lst_acc_rnn_0.1}
    \end{subfigure}
    \hfill
    \begin{subfigure}[b]{0.19\textwidth}
        \centering
        \includegraphics[width=\textwidth]{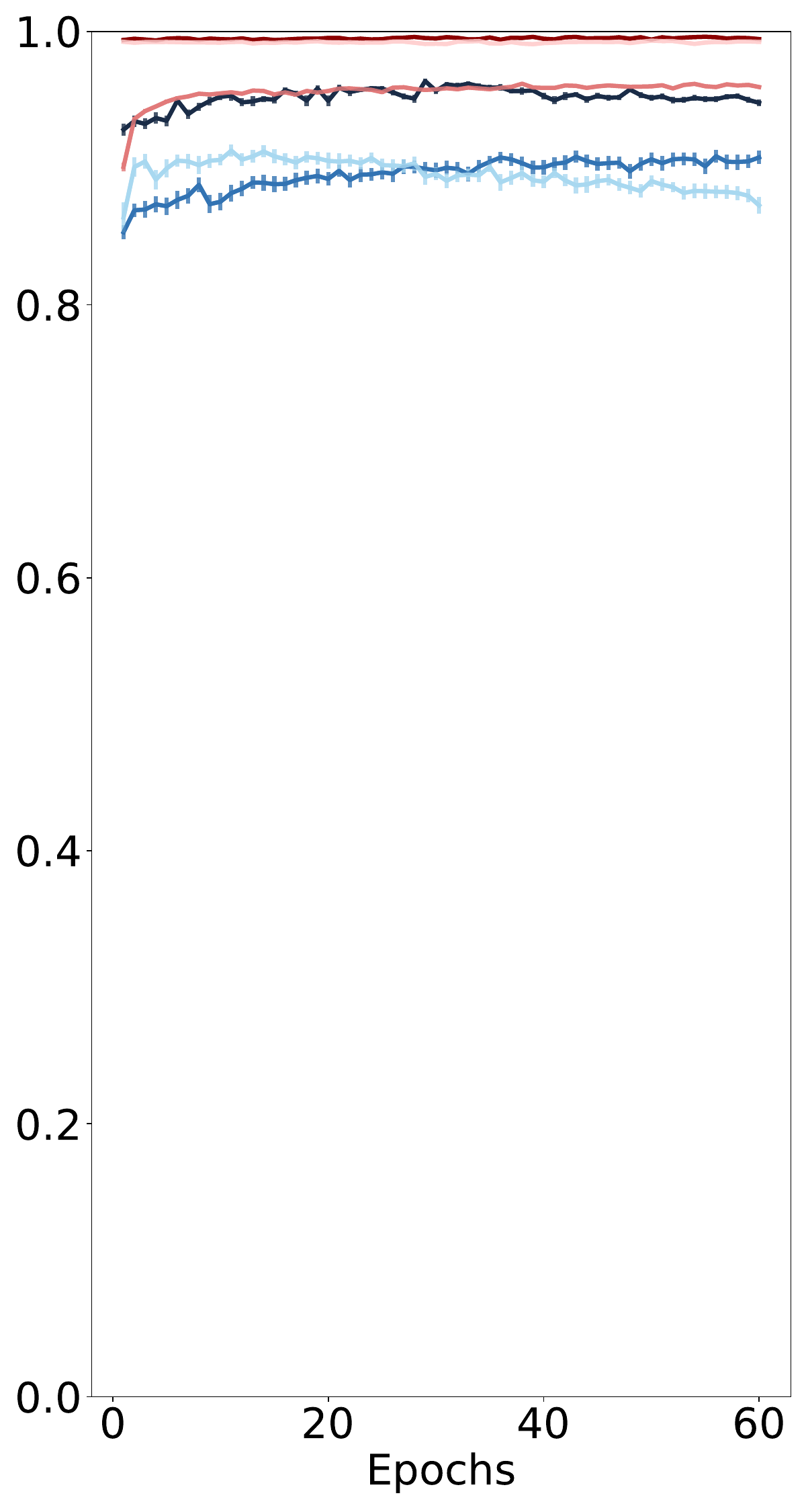}
        \caption{Communication (noise 0)}
        \label{fig:final_comm_acc_rnn}
    \end{subfigure}
    \hfill
    \begin{subfigure}[b]{0.19\textwidth}
        \centering
        \includegraphics[width=\textwidth]{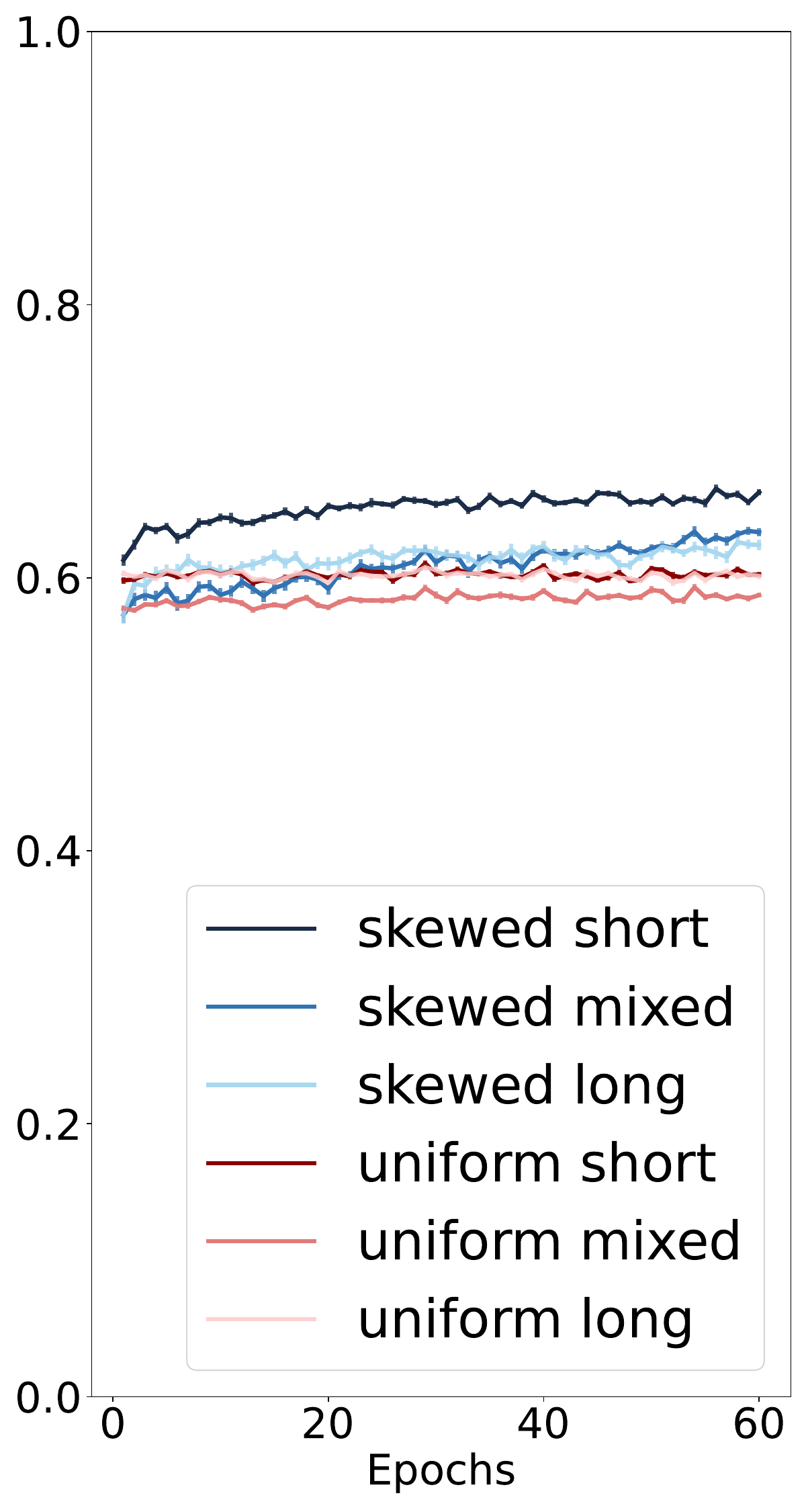}
        \caption{Communication (noise 0.1)}
        \label{fig:final_comm_acc_rnn_0.1}
    \end{subfigure}
    \caption{Performance of RNN-based agents in speaking, listening with noise levels 0 and 0.1, and communication with noise levels 0 and 0.1 during learning and communication in the \textbf{verb-final language} under \textbf{the \textit{Baseline} and the noise condition}.}
    \label{fig:final_acc_rnn_combined}
\end{figure}

\bmsection{Learning and Communication Accuracies in the Verb-final Language}
\label{app:acc_final}
As shown in Figure~\ref{fig:final_acc_rnn_combined}, mostly similar results were found for verb-initial and verb-final languages regarding learning and communication accuracies.  

\bmsection{Error Analyses}
\label{app:error_analyses}

\begin{figure}[t]
\centering
\begin{tabular}{c c c c c}
    % ---------- First row: Baseline ----------
    \raisebox{1.5cm}{\rotatebox{90}{Baseline}} &
    \begin{subfigure}[b]{0.22\textwidth}
        \centering
        \includegraphics[width=\textwidth]{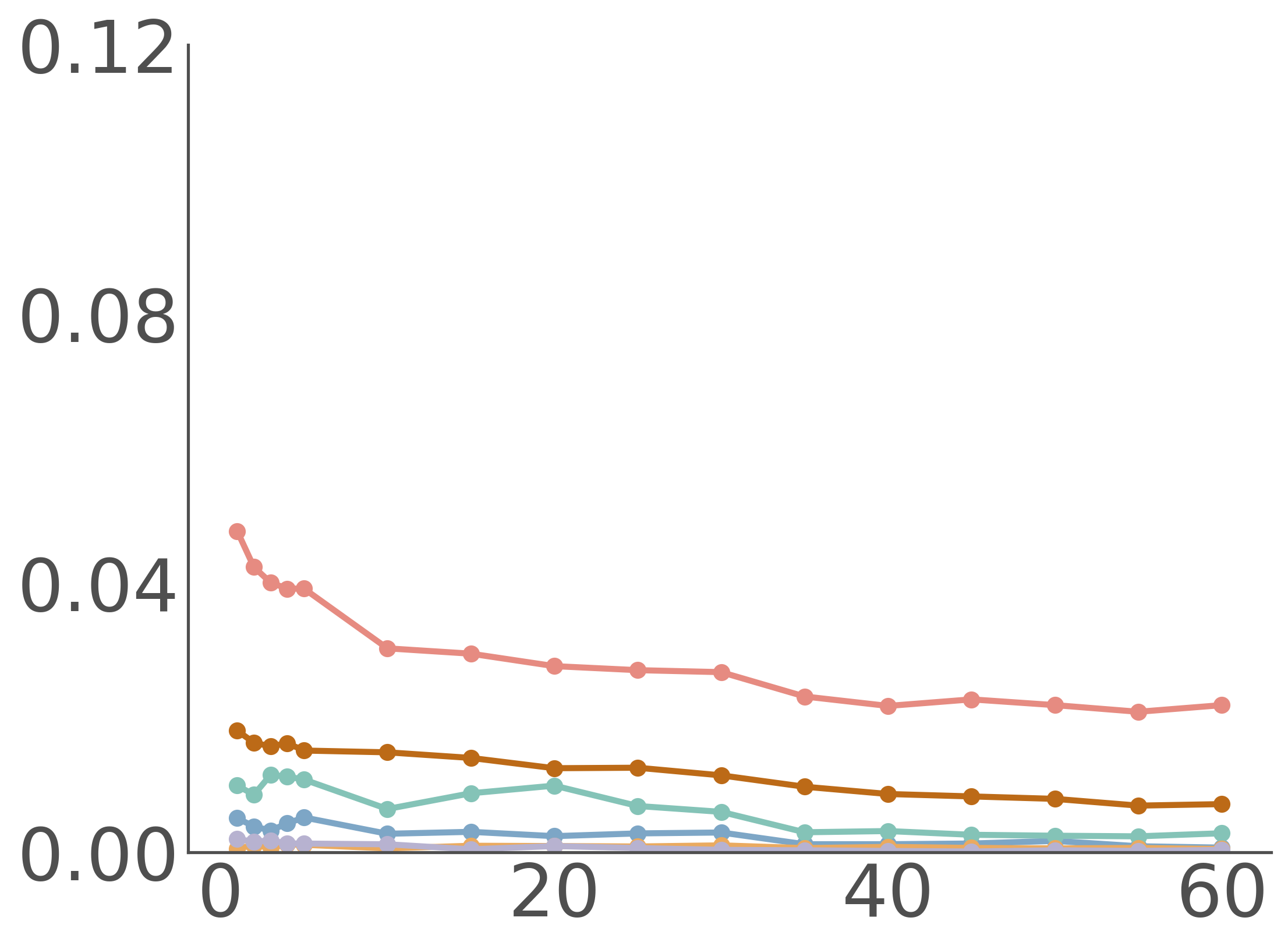}
        \caption{Initial, Short DL}
        \label{fig:os_ratios}
    \end{subfigure} &
    \begin{subfigure}[b]{0.22\textwidth}
        \centering
        \includegraphics[width=\textwidth]{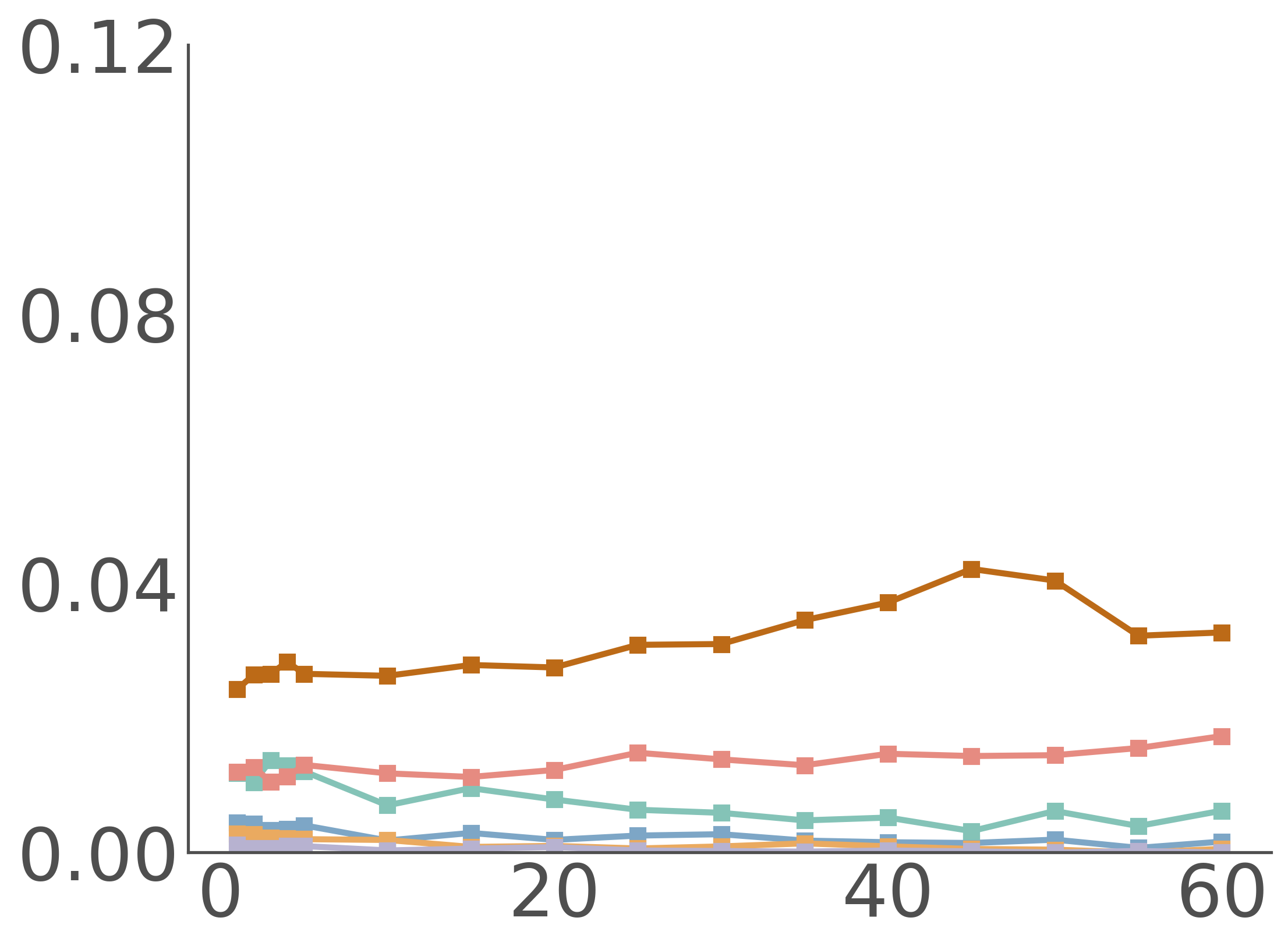}
        \caption{Initial, Long DL}
        \label{fig:so_ratios}
    \end{subfigure} &
    \begin{subfigure}[b]{0.22\textwidth}
        \centering
        \includegraphics[width=\textwidth]{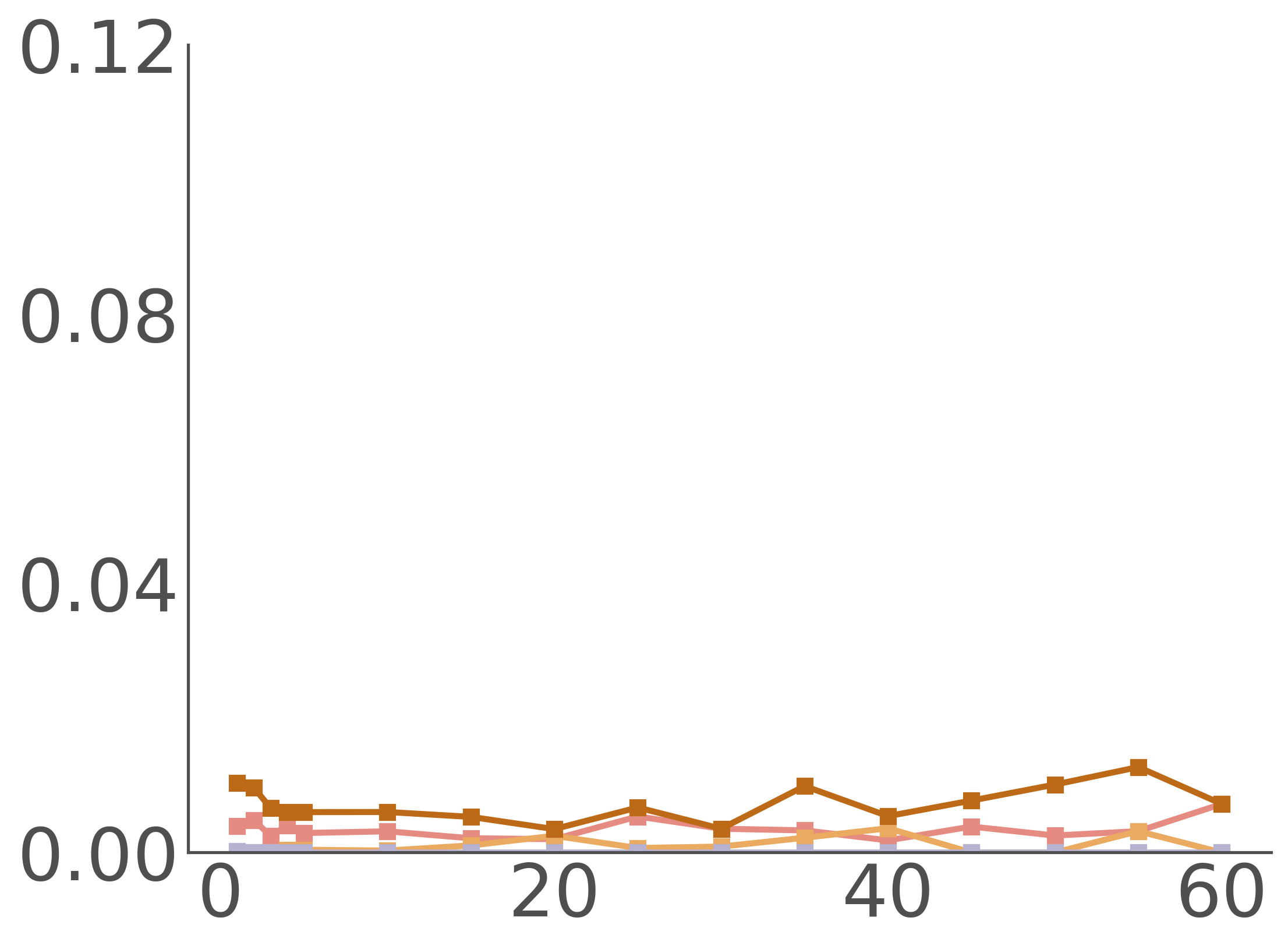}
        \caption{Final, Short DL}
        \label{fig:os_ratios_vf}
    \end{subfigure} &
    \begin{subfigure}[b]{0.22\textwidth}
        \centering
        \includegraphics[width=\textwidth]{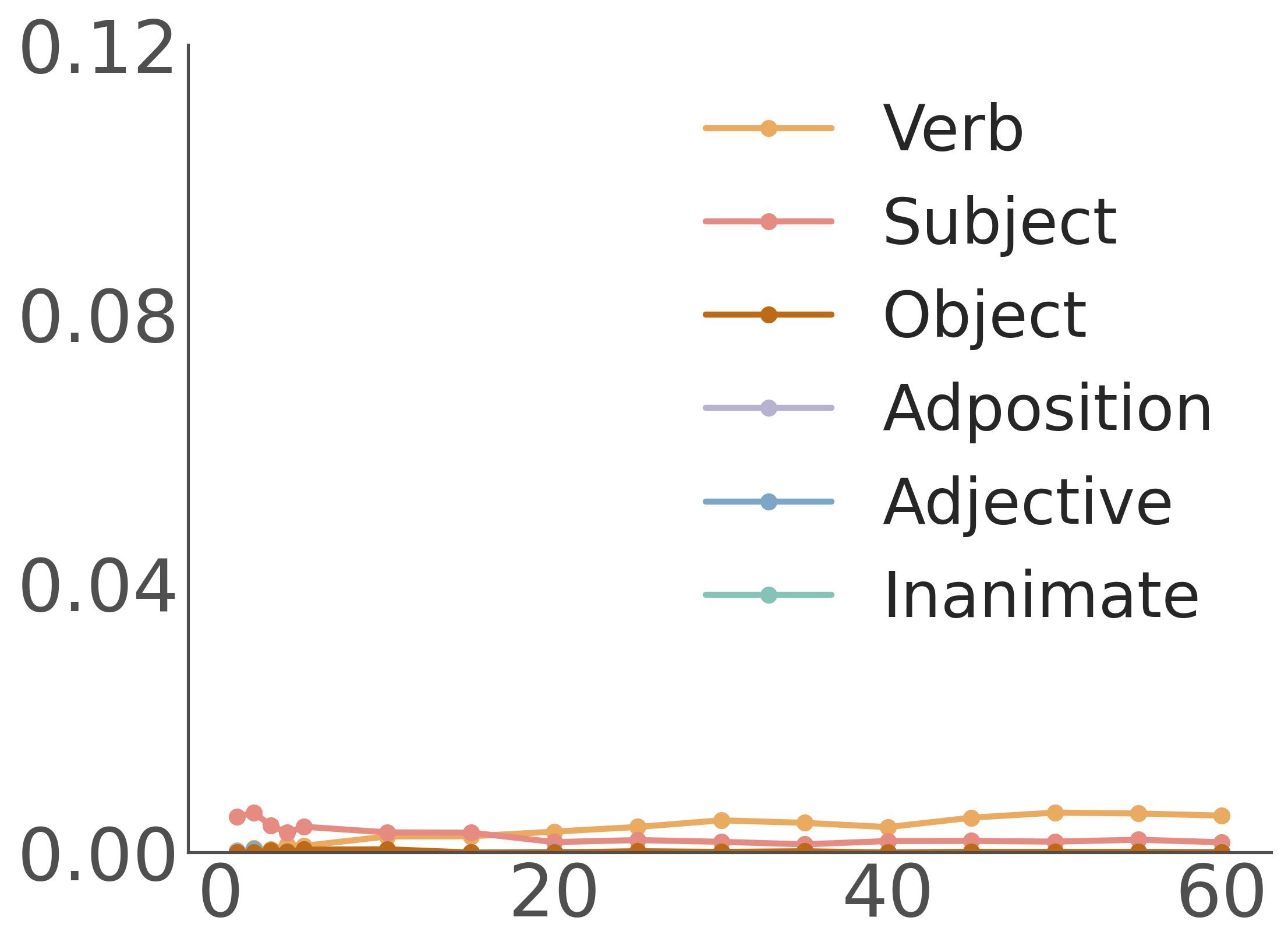}
        \caption{Final, Long DL}
        \label{fig:so_ratios_vf}
    \end{subfigure} \\[1em]
    
    % ---------- Second row: Lower spk capacity ----------
\raisebox{1.5cm}{\rotatebox{90}{Lower speaker capacity}} &
    \begin{subfigure}[b]{0.22\textwidth}
        \centering
        \includegraphics[width=\textwidth]{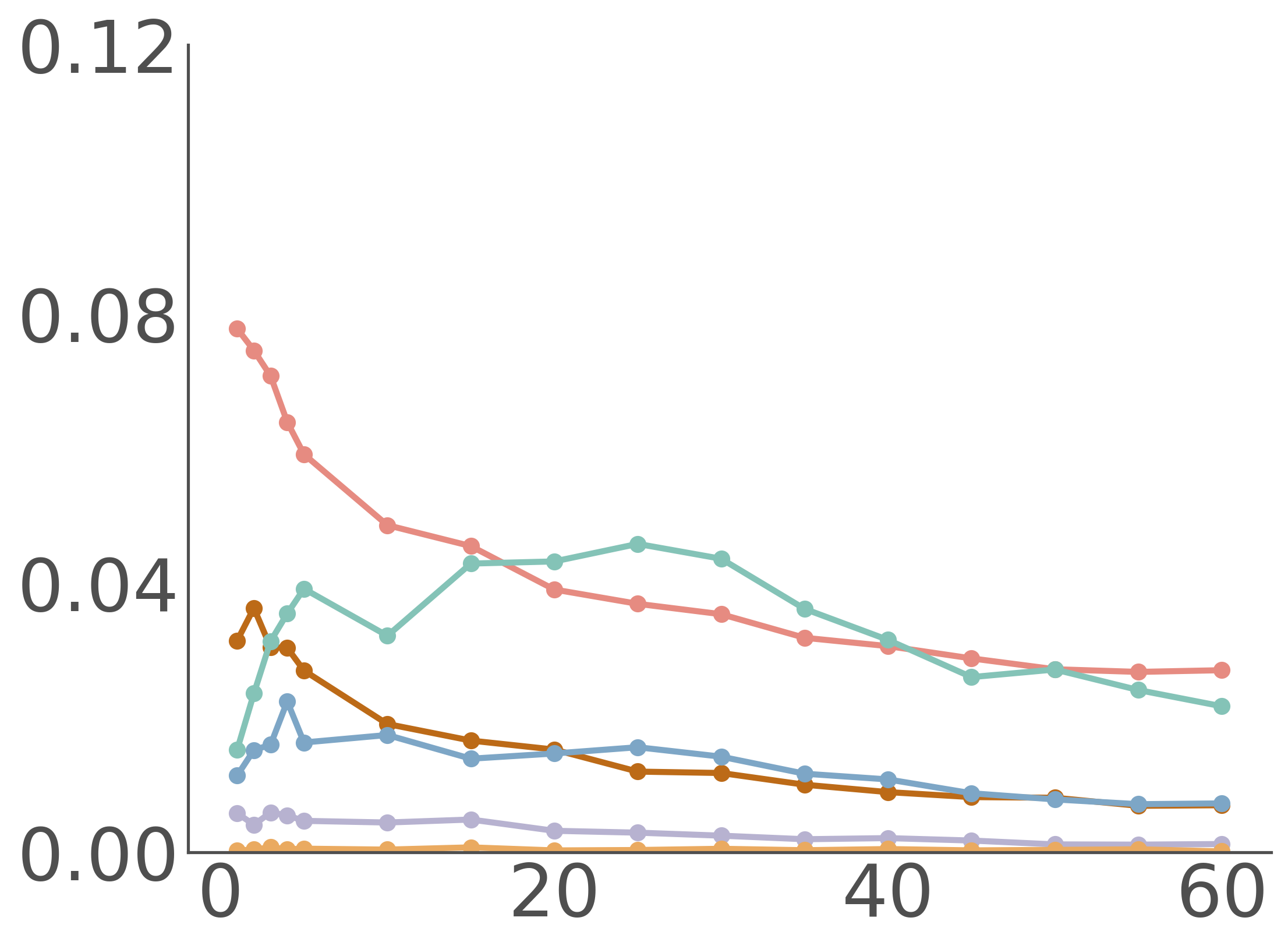}
        \caption{Initial, Short DL}
        \label{fig:os_ratios_256}
    \end{subfigure} &
    \begin{subfigure}[b]{0.22\textwidth}
        \centering
        \includegraphics[width=\textwidth]{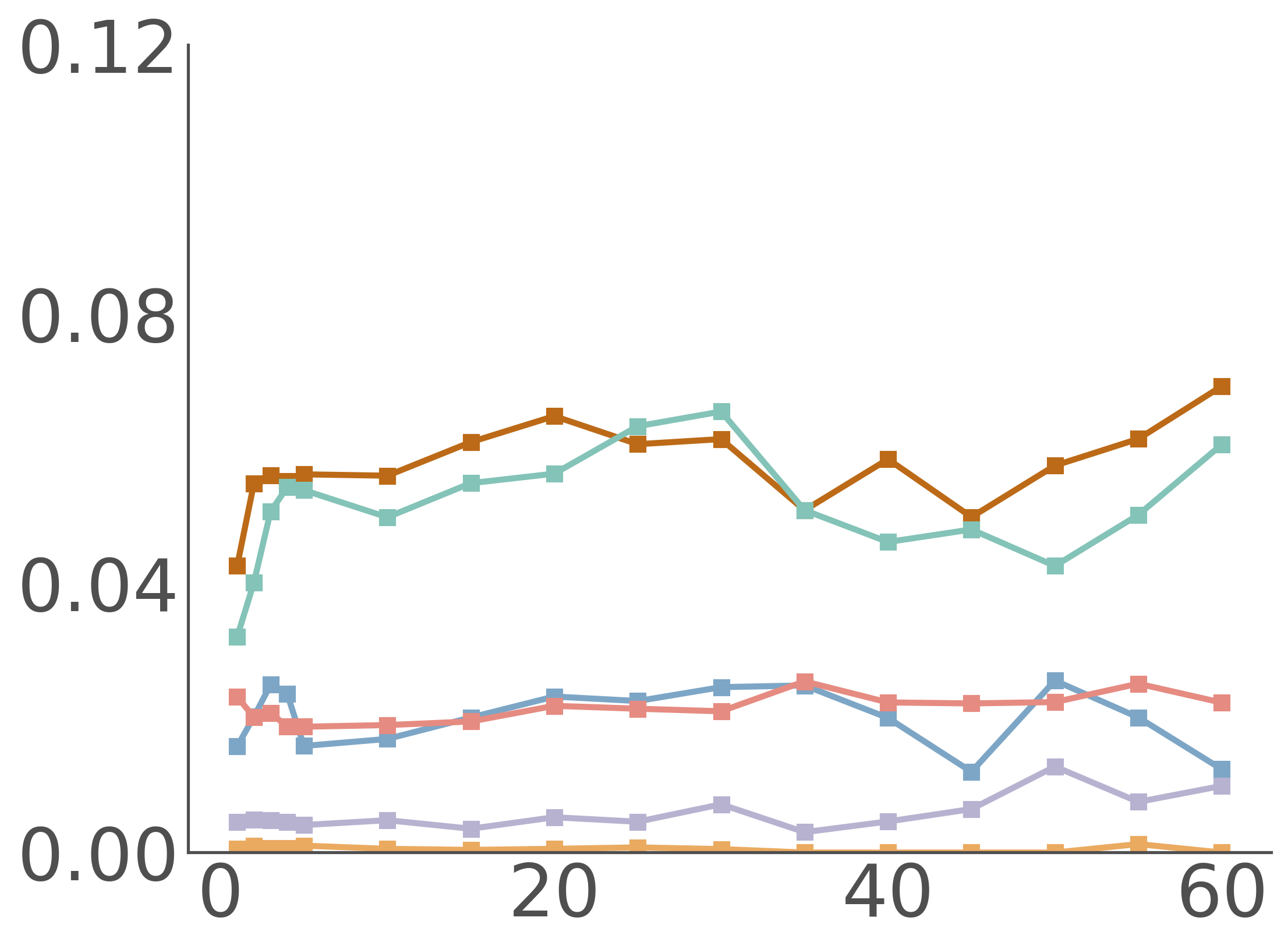}
        \caption{Initial, Long DL}
        \label{fig:so_ratios_256}
    \end{subfigure} &
    \begin{subfigure}[b]{0.22\textwidth}
        \centering
        \includegraphics[width=\textwidth]{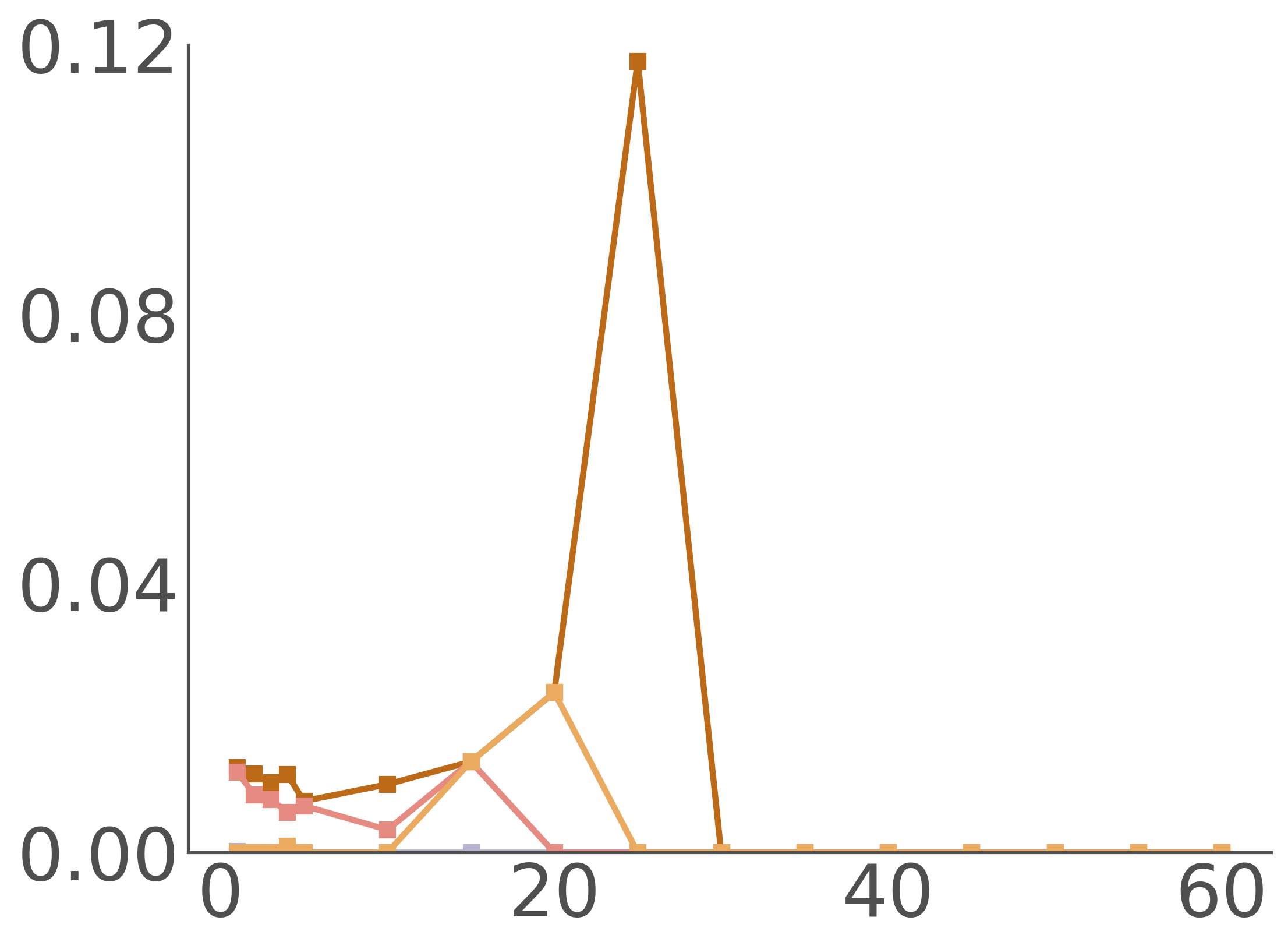}
        \caption{Final, Short DL}
        \label{fig:os_ratios_256_vf}
    \end{subfigure} &
    \begin{subfigure}[b]{0.22\textwidth}
        \centering
        \includegraphics[width=\textwidth]{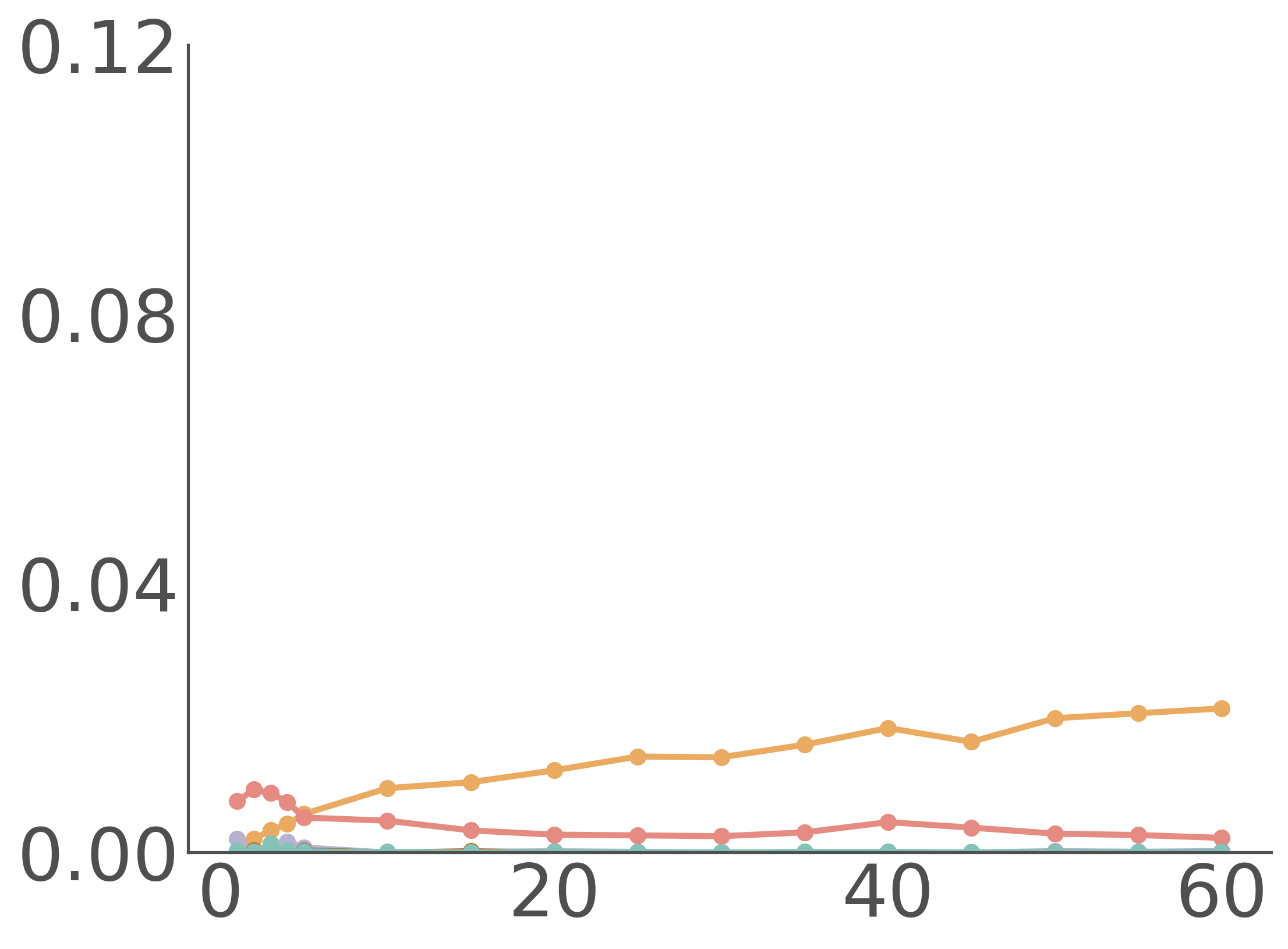}
        \caption{Final, Long DL}
        \label{fig:so_ratios_256_vf}
    \end{subfigure} \\[1em]
    
    % ---------- Third row: Noise 0.1 ----------
\raisebox{1.5cm}{\rotatebox{90}{Noise}} &
    \begin{subfigure}[b]{0.22\textwidth}
        \centering
        \includegraphics[width=\textwidth]{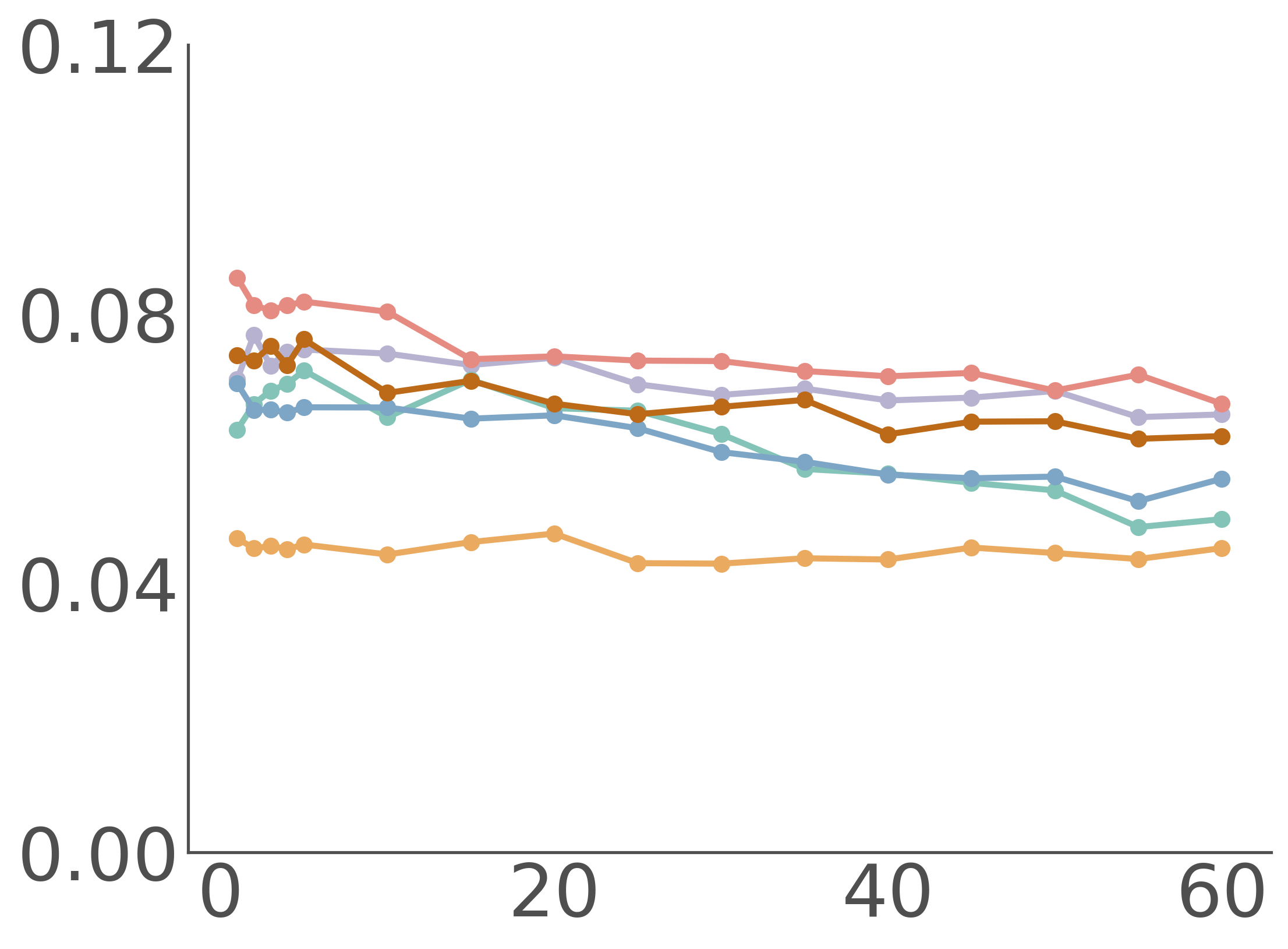}
        \caption{Initial, Short DL}
        \label{fig:os_ratios_noise}
    \end{subfigure} &
    \begin{subfigure}[b]{0.22\textwidth}
        \centering
        \includegraphics[width=\textwidth]{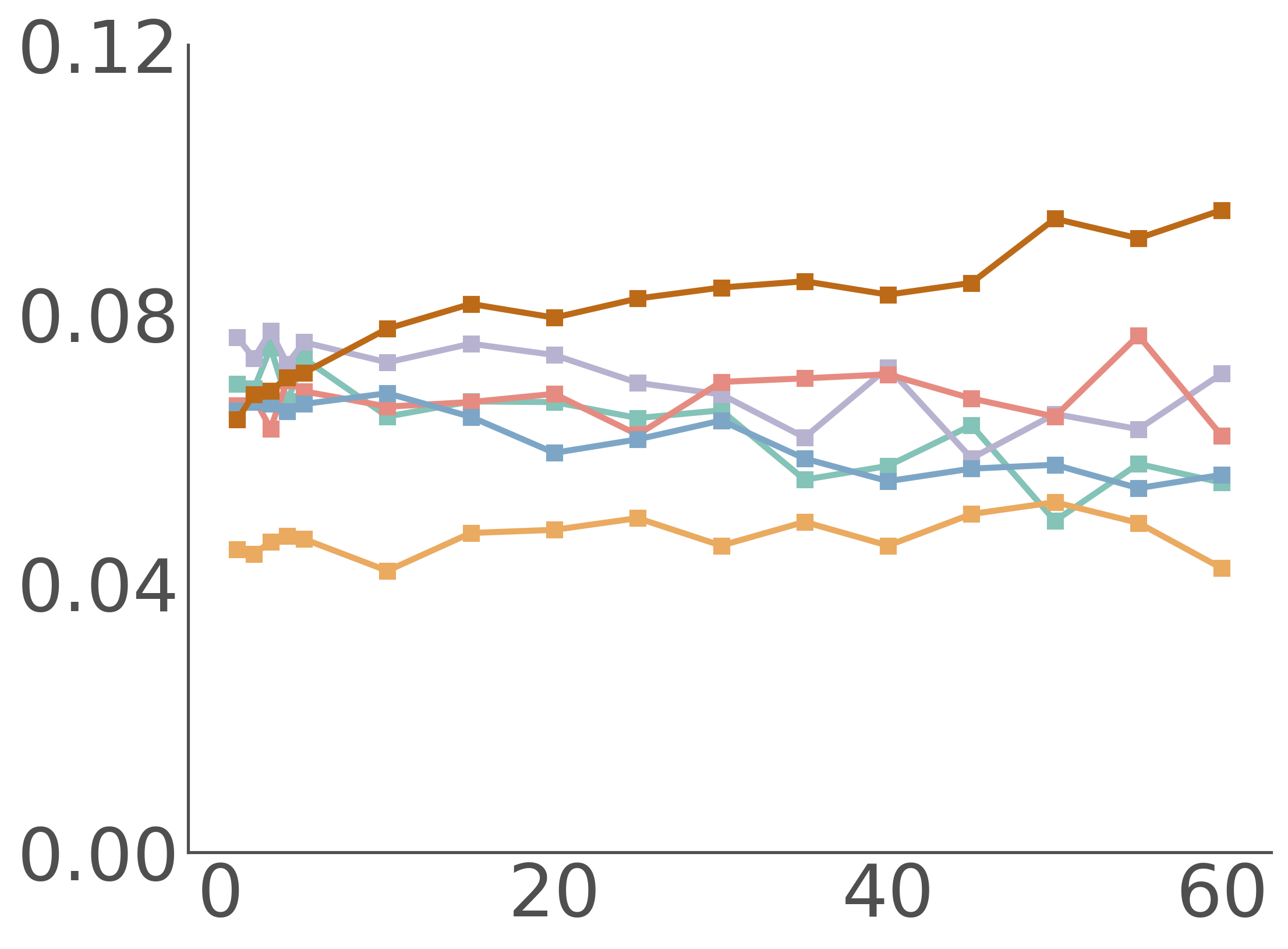}
        \caption{Initial, Long DL}
        \label{fig:so_ratios_noise}
    \end{subfigure} &
    \begin{subfigure}[b]{0.22\textwidth}
        \centering
        \includegraphics[width=\textwidth]{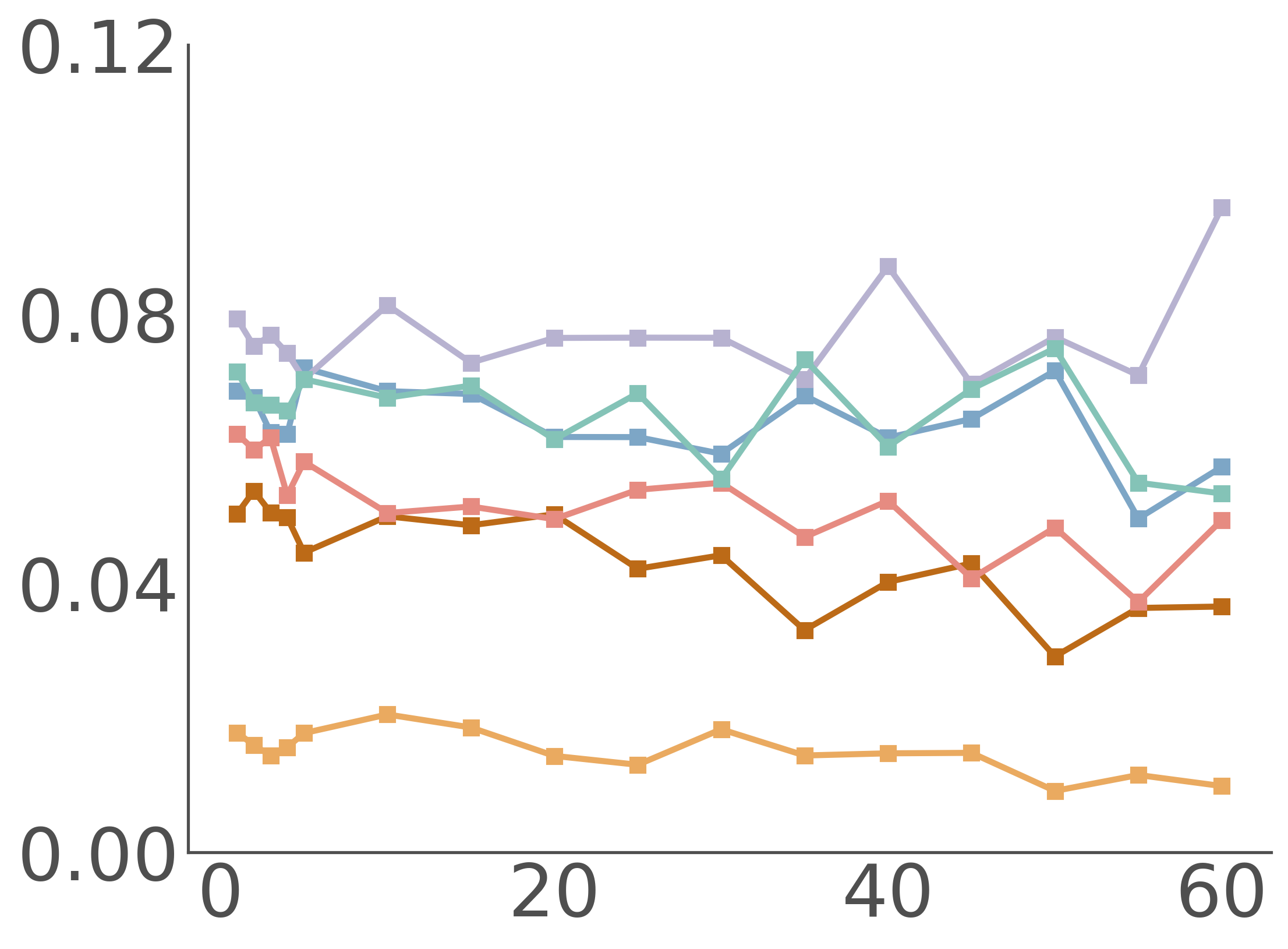}
        \caption{Final, Short DL}
        \label{fig:os_ratios_noise_vf}
    \end{subfigure} &
    \begin{subfigure}[b]{0.22\textwidth}
        \centering
        \includegraphics[width=\textwidth]{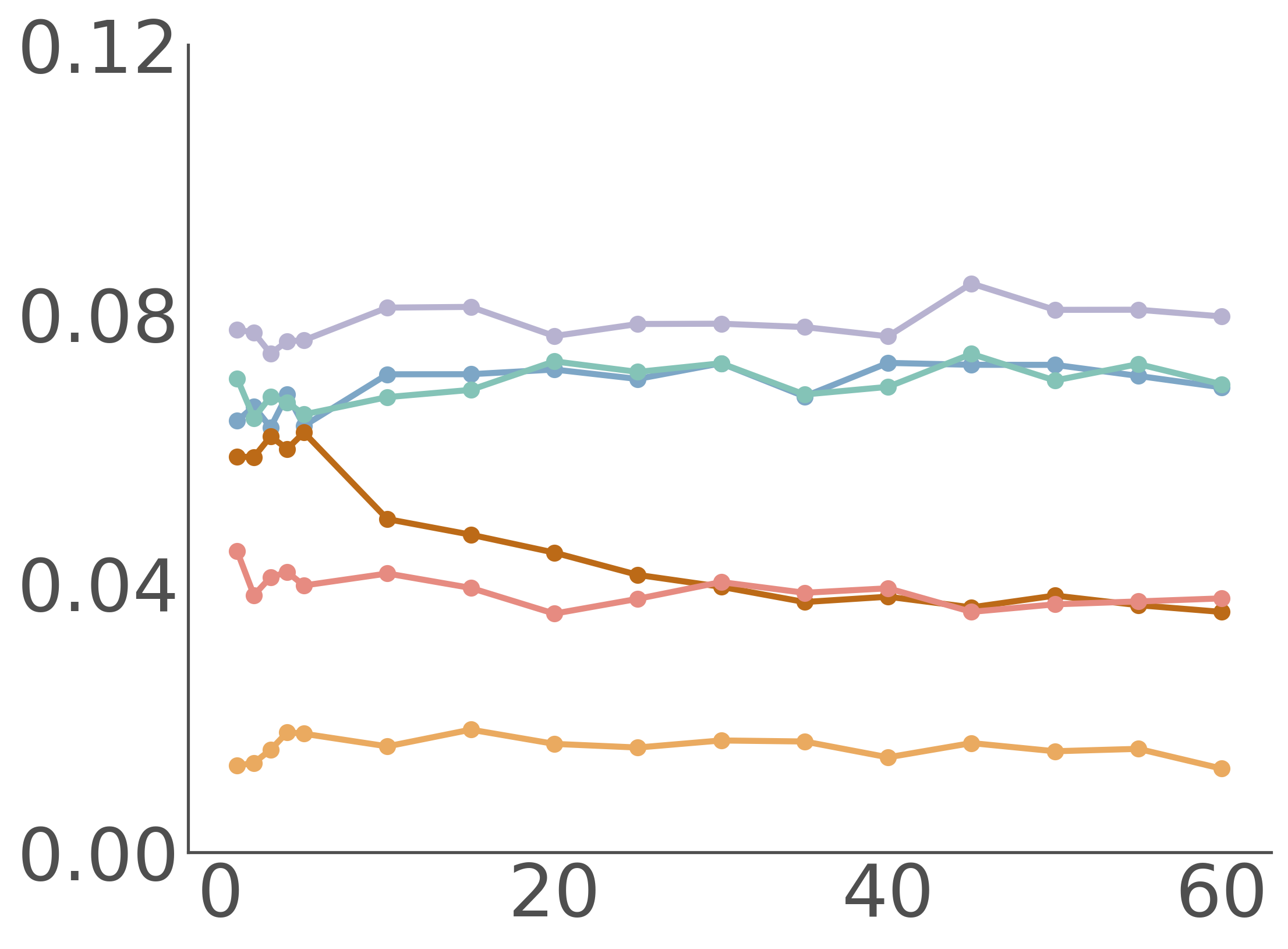}
        \caption{Final, Long DL}
        \label{fig:so_ratios_noise_vf}
    \end{subfigure} \\
\end{tabular}

\caption{Evolution of error percentages (y-axis) for all meaning items across epochs (x-axis). 
Columns 1 and 2 show results for \textbf{verb-initial} languages, while columns 3 and 4 show results for \textbf{verb-final} languages. 
Rows correspond to different experimental conditions: \textit{Baseline}, \textit{Lower speaker capacity}, and \textit{Noise}. 
Error rates are calculated separately for short- and long-dependency utterances. 
Adposition, Adjective, and Inanimate refer to adposition, adjective, and inanimate noun in the modifier phrase.}
\label{fig:os_so_ratios_all}
\end{figure}

To better understand agents' production behavior, 
we conducted error analyses of their production during communication (After 60 RL epochs). Figure~\ref{fig:os_so_ratios_all} illustrates the evolution of error percentages for all meaning items across epochs in the two languages. Error rates were calculated separately for utterances with short and long dependencies under three experimental conditions: \textit{Baseline}, \textit{Lower speaker capacity}, and \textit{Noise}. 
The error percentage for meaning item categories was calculated by dividing the number of prediction errors by the total number of generated utterances.

For the verb-initial language, in the \textit{Baseline} condition (top row of Figure~\ref{fig:os_so_ratios_all}), agents gradually achieved lower error rates for short dependencies (VOS) while maintaining higher rates for long dependencies (VSO). Combined with what we have seen in Figure~\ref{fig:comm_initial_rnn_skewed}, at the beginning of communication, agents produced an equal mix of SO and OS utterances, resulting in slightly higher error rates for meaning items, especially for agent and patient. Over time, the proportion of short DL utterances surpassed that of long DL utterances. In line with our expectations,  the overall error rates for agent and patient decreased. 
For the verb-final language, in the \textit{Baseline} condition (top row of Figure~\ref{fig:os_so_ratios_all}), agents gradually achieved lower error rates for long dependencies (OSV) while maintaining higher rates for short dependencies (SOV)\footnote{In the verb-final language, agents with lower speaker capacity only produced long-dependency utterances as of epoch 30, causing the error percentage in Figure~\ref{fig:os_ratios_256_vf} to drop to zero.}. 

The overall pattern suggested that a large portion of errors stemmed from role ambiguity, specifically misidentification of the agent or patient. For short-dependency utterances in the verb-initial language, while the initial error rate for agent or patient misidentification was relatively high, it declined noticeably over time, eventually reaching its lowest point compared to long-dependency utterances. The reversed pattern was found for the verb-final language.

In the lower speaker capacity condition (middle row) and the \textit{Noise} condition (bottom row), 
error rates were decreasing, but less strongly than in the \textit{Baseline} condition, with an overall higher error percentage. 
This pattern indicated that limited speaker capacity and noise hindered agents' ability to recover meanings. Despite this, the trends in agent and patient misidentification remained similar to the \textit{Baseline} condition.

% why agent and patient specifically are affected by the various conditions (more than other items), right? If so, the reason is that we're working with the skewed language, and in this language only agent and patient have conditional probs wrt to the verb... --- I agree

\end{document}